\documentclass[manuscript]{clv2025} 
\usepackage[T1]{fontenc}
\usepackage[utf8]{inputenc}

\usepackage{microtype}
\usepackage{soul}
\usepackage{xcolor}
\usepackage{graphicx}
\usepackage{adjustbox}
\usepackage{booktabs}
\usepackage{multirow}
\usepackage{multicol}
\usepackage{tikz}
\usetikzlibrary{shapes.geometric, arrows}
\usetikzlibrary{positioning, shapes, arrows.meta, calc, fit}

\usepackage{pgfplots}
\usepackage[edges]{forest}

\usepackage[normalem]{ulem}
\usepackage{subcaption}
\usepackage{pifont}

\usepackage{hyperref}

\usepackage{siunitx}
\makeatletter
\@ifclasswith{clv2025}{manuscript}{
    \usepackage{setspace}
    \usepackage{etoolbox}
    \usepackage[font=doublespacing]{caption}
    \AtBeginEnvironment{thebibliography}{\doublespacing}
    \AtBeginEnvironment{acknowledgments}{\doublespacing}
    
    \newcommand{\forceabstractspace}{\doublespacing} 
}{
    \newcommand{\forceabstractspace}{} 
}
\makeatother

\usepackage{amssymb}
\usepackage{amsmath}
\usepackage{comment}
\usepackage[capitalise]{cleveref}
\definecolor{safatodocolor}{RGB}{0,255,0}
\definecolor{paired-light-blue}{RGB}{198, 219, 239}
\definecolor{paired-dark-blue}{RGB}{49, 130, 188}
\definecolor{paired-light-orange}{RGB}{251, 208, 162}
\definecolor{paired-dark-orange}{RGB}{230, 85, 12}
\definecolor{paired-light-green}{RGB}{199, 233, 193}
\definecolor{paired-dark-green}{RGB}{49, 163, 83}
\definecolor{paired-light-purple}{RGB}{218, 218, 235}
\definecolor{paired-dark-purple}{RGB}{117, 107, 176}
\definecolor{paired-light-gray}{RGB}{217, 217, 217}
\definecolor{paired-dark-gray}{RGB}{99, 99, 99}
\definecolor{paired-light-pink}{RGB}{222, 158, 214}
\definecolor{paired-dark-pink}{RGB}{123, 65, 115}
\definecolor{paired-light-red}{RGB}{231, 150, 156}
\definecolor{paired-dark-red}{RGB}{131, 60, 56}
\definecolor{paired-light-yellow}{RGB}{231, 204, 149}
\definecolor{paired-dark-yellow}{RGB}{141, 109, 49}
\definecolor{light-green}{RGB}{118, 207, 180}
\definecolor{raspberry}{RGB}{228, 24, 99}
\newcommand{\cmark}{\ding{51}}%
\newcommand{\xmark}{\ding{55}}%
\newcommand{\clrev}[1]{\textcolor{black}{#1}}
\newcommand{\clfinrev}[1]{\textcolor{black}{#1}}
\Crefname{section}{Sec.}{Secs.}
\newcommand{\EmphBox}[1]{\vspace*{-3pt}\noindent\fcolorbox{black!20}{black!10}{\parbox{.99\textwidth}{#1}}}

\usepackage[nopostdot,toc,acronym,nomain,nonumberlist,nowarn]{glossaries}
\makeglossaries
\setacronymstyle{long-short}
\glsdisablehyper
\newacronym[type=\glsdefaulttype]{LLM}{LLM}{Large Language Model}
\newacronym[type=\glsdefaulttype]{NLP}{NLP}{Natural Language Processing}
\newacronym[type=\glsdefaulttype]{SOTA}{SOTA}{state-of-the-art}
\newacronym[type=\glsdefaulttype]{EPM}{EPM}{exact phrase match}
\newacronym[type=\glsdefaulttype]{EWM}{EWM}{exact word match}
\newacronym[type=\glsdefaulttype]{TOD}{TOD}{Task-Oriented Dialogue}
\newacronym[type=\glsdefaulttype]{BPMN}{BPMN}{Business Process and Model Notation}
\newacronym[type=\glsdefaulttype]{PDDL}{PDDL}{Planning Domain Definition Language}
\newacronym[type=\glsdefaulttype]{LTL}{LTL}{Linear Temporal Logic}
\newacronym[type=\glsdefaulttype]{FOL}{FOL}{First-Order-Language}
\newacronym[type=\glsdefaulttype]{ESD}{ESD}{Event Sequence Description}
\newacronym[type=\glsdefaulttype]{AP}{AP}{Affinity Propagation}
\newacronym[type=\glsdefaulttype]{HMM}{HMM}{Hidden Markov Model}
\newacronym[type=\glsdefaulttype]{SRL}{SRL}{Semantic Role Labeling}
\newacronym[type=\glsdefaulttype]{DRL}{DRL}{Deep Reinforcement Learning}
\newacronym[type=\glsdefaulttype]{TAP}{TAP}{Trigger Action Program}
\newacronym[type=\glsdefaulttype]{MSA}{MSA}{Multiple Sequence Alignment}
\newacronym[type=\glsdefaulttype]{ICL}{ICL}{In-Context Learning}
\newacronym[type=\glsdefaulttype]{RAG}{RAG}{Retrieval-Augmented Generation}
\newacronym[type=\glsdefaulttype]{EM}{EM}{Exact Match}
\newacronym[type=\glsdefaulttype]{SR}{SR}{Success Rate}
\newacronym[type=\glsdefaulttype]{GCS}{GCS}{Goal-Conditioned Success}
\newacronym[type=\glsdefaulttype]{ARA}{ARA}{Aligned Recipe Actions}
\newacronym[type=\glsdefaulttype]{PSE}{PSE}{Process Structure Extraction}
\newacronym[type=\glsdefaulttype]{PSI}{PSI}{Process Structure Induction}
\newacronym[type=\glsdefaulttype]{Open-IE}{Open-IE}{Open Information Extraction}
\newacronym[type=\glsdefaulttype]{STRIPS}{STRIPS}{Stanford Research Institute Problem Solver}
\newacronym[type=\glsdefaulttype]{RLHF}{RLHF}{Reinforcement Learning from Human Feedback}
\newacronym[type=\glsdefaulttype]{VerbNet}{VerbNet}{Verb Network}

\jvol{vv}
\jnum{nn}
\jyear{2026}

\dochead{Survey} 

\pageonefooter{Action editor: Yuki Arase. Submission received: 27 June 2025; revised version received: 23 December 2025; accepted for publication: 24 February 2026.}

\runningtitle{Survey on Instructional Text in NLP}
\runningauthor{Safa et al.}

\begin{document}

\title{Instructional Text Across Disciplines: A Survey of Representations, Downstream Tasks, and Open Challenges Toward Capable AI Agents}

\author{%
  Abdulfattah Safa\thanks{Corresponding author}\,$^{1,2}$, 
  Tamta Kapanadze\,$^{2}$, 
  Arda Uzunoğlu\,$^{3}$, 
  Gözde Gül Şahin\,$^{1,2,4}$
}

\affilblock{%
  \affil{KUIS AI Lab, Koç University, Istanbul, Türkiye\\\quad \email{https://gglab-ku.github.io/}}
  \affil{Koç University, Istanbul, Türkiye}
  \affil{Johns Hopkins University, Baltimore, USA}
  \affil{Friedrich-Alexander-Universität Erlangen-Nürnberg, Germany}
}

\maketitle

\begin{abstract}
\forceabstractspace
Recent advances in large language models have demonstrated promising capabilities in following simple instructions through instruction tuning. However, real-world tasks often involve complex, multi-step instructions that remain challenging for current NLP systems. \clrev{Robust understanding of such instructions is essential for deploying LLMs as general-purpose agents that can be programmed in natural language to perform complex, real-world tasks across domains like robotics, business automation, and interactive systems.} Despite growing interest in this area, there
is a lack of a comprehensive survey that systematically analyzes the landscape of complex instruction understanding and processing. Through a systematic review of the literature, we analyze available resources, representation schemes, and downstream tasks related to instructional text. Our study examines 181 papers, identifying trends, challenges, and opportunities in this emerging field. We provide AI/NLP researchers with essential background knowledge and a unified view of various approaches to complex instruction understanding, bridging gaps between different research directions and highlighting future research opportunities.
\end{abstract}

\section{Introduction}
The ability to program machines/computers with natural language, if it can be made successful, would fundamentally change the relationship between humans and computers. As of today, only a small percentage of humans (less than 1\%) have the necessary skill set to program their computers or phones to perform new tasks. Machines and computers, however, are mostly viewed as preprogrammed devices with a fixed-set of skills. To move towards that goal, \textit{programmable machines}\clrev{\textit{---more recently coined as a new term: AI agents---}}\footnote{Historically, programmable machines have been used to define systems that can be configured through natural language rather than code. Recently, it has evolved into today's AI agent paradigm, where users \textit{prompt} LLM models with natural language to pursue their goals in an environment using external tools such as Python interpreter.} should be equipped with (at least) excellent instruction understanding capabilities. 

With the recent advances in the \gls{NLP} field, \gls{LLM}s have demonstrated capacity on understanding instructions \citep{naveed2024comprehensiveoverviewlargelanguage}. This is mostly achieved via ``instruction tuning'' that performs supervised finetuning on base language models with a large amount of instruction-response pairs~\citep{zhang2023instruction}. These pairs are typically single sentences, describing a low-level task that can be performed in a single step. \clrev{More recently, instruction-tuned models are further trained with human preferences and reasoning traces~\cite{deepseekR12025} (e.g., chain-of-thought) via various reinforcement learning techniques~\cite{Rafailov2023DPO, ouyang2022training} that boost their performance on standard reasoning benchmarks
~\cite{MMMU,GPQA,hendrycksmath2021,codeforces}. This has ultimately led to another research direction: LLM-based AI agents that can interact with environments such as websites~\cite{Mind2Web2023, lai2024autowebglm, WebArena23, he2024webvoyagerbuildingendtoendweb} using tools such as python interpreters~\cite{CodeAct, InterCode, schick2023toolformerlanguagemodelsteach} to autonomously pursue user goals. Even though modern \gls{LLM}-based agents show promise in web navigation and tool use, they still struggle with complex, multi-step instructions\footnote{\clrev{SOTA \gls{LLM} agents achieve 8–30\% success rates~\cite{TheAgentCompany2025}, requiring an average of 27–40 interaction steps to completion depending on the model, with performance particularly degraded on long-horizon tasks requiring multi-step reasoning across changing contexts.}} which contain temporal~\cite{Robotouille2025}, conditional~\cite{Shridhar2020ALFWorldAT, ghazarian2025todprocbench}, hierarchical dependencies~\cite{WebArena23, PlanCraft2025}.
} 

Furthermore, the ability to ``follow instructions'' of such \clrev{models} is mostly evaluated on simple scenarios with simple instructions \citep{he2024largelanguagemodelsunderstand} (e.g., a single event with an explicit set of arguments), \clrev{on simulated environments that approximate real-world complexity}. However, real-world tasks require \clrev{resolving event and argument ambiguities in underspecified instructions}; as well as \clrev{understanding and following more complex instructions, which are mostly multi-step directives containing temporal, conditional, hierarchical dependencies}. Hence, the next frontier in NLP---in particular LLM-based agents---research will be to understand these complex instructions in a real-world-like setting.

Understanding such instructions requires---at minimum---understanding of events, the relations between events, their participants and the environment, and be able to perform multi-hop, common sense reasoning with such event knowledge. Various fields have investigated related subtopics, primarily as computational linguistics and \gls{NLP}~(e.g., event semantics, semantic parsing, script/scenario generation, common-sense reasoning, LLM agents), robotics~(e.g., manipulation via instructions), business intelligence~(e.g., process models), and computer vision~(e.g., recipe understanding). Different fields use different naming conventions for complex instructions (e.g., process, procedure, task), different techniques to represent them (e.g., graph, workflow, business model, programming language, etc.) and different venues to publish, which creates a high barrier to entry for new researchers. 
\tikzset{%
    parent/.style =          {align=center,text width=2cm,rounded corners=3pt, line width=0.3mm, fill=gray!10,draw=gray!80},
    child/.style =           {align=center,text width=2.3cm,rounded corners=3pt, fill=blue!10,draw=blue!80,line width=0.3mm},
    grandchild/.style =      {align=center,text width=2cm,rounded corners=3pt},
    greatgrandchild/.style = {align=center,text width=1.5cm,rounded corners=3pt},
    greatgrandchild2/.style = {align=center,text width=1.5cm,rounded corners=3pt},    
    referenceblock/.style =  {align=center,text width=1.5cm,rounded corners=2pt},
    data/.style =           {align=center,text width=1.2cm,rounded corners=3pt, fill=paired-light-blue!50,draw=paired-dark-blue!65,line width=0.3mm},   
    data_work/.style =      {align=center, text width=4.4cm,rounded corners=3pt, fill=paired-light-blue!50,draw=blue!0,line width=0.3mm},  
    model/.style =           {align=center,text width=1.2cm,rounded corners=3pt, fill=paired-light-orange!50,draw=paired-dark-orange!65,line width=0.3mm},   
    model_work/.style =      {align=center,text width=4.4cm,rounded corners=3pt, fill=paired-light-orange!50,draw=red!0,line width=0.3mm},    
    model_work2/.style =      {align=center,text width=2.8cm,rounded corners=3pt, fill=paired-light-orange!50,draw=red!0,line width=0.3mm},   
    pretraining/.style =        {align=center,text width=1.2cm,rounded corners=3pt, fill= paired-light-green!50,draw=paired-dark-green!75,line width=0.3mm},   
    pretraining_work/.style =   {align=center,text width=4.4cm,rounded corners=3pt, fill= paired-light-green!50,draw= cyan!0,line width=0.3mm},      
    finetuning/.style =           {align=center,text width=1.2cm,rounded corners=3pt, fill= paired-light-purple!50,draw=paired-dark-purple!75,line width=0.3mm},   
    finetuning_work/.style =      {align=center,text width=4.3cm,rounded corners=3pt, fill= paired-light-purple!50,draw= orange!0,line width=0.3mm},        
    inference/.style =           {align=center,text width=1.2cm,rounded corners=3pt, fill= paired-light-red!35,draw=paired-light-red!90,line width=0.3mm},   
    inference_work/.style =      {align=center,text width=4.3cm,rounded corners=3pt, fill= paired-light-red!35,draw= magenta!0,line width=0.3mm},
    hardware/.style =           {align=center,text width=1.2cm,rounded corners=3pt, fill= paired-light-yellow!35,draw=paired-light-yellow!90,line width=0.3mm},   
    hardware_work/.style =      {align=center,text width=4.3cm,rounded corners=3pt, fill= paired-light-yellow!35,draw= magenta!0,line width=0.3mm},
}

\begin{figure*}[!htb]
    \scriptsize
    \centering
    \resizebox{0.9\textwidth}{!}
    {
    \begin{minipage}[b]{0.54\linewidth}
    \centering
    \begin{forest}
        for tree={
            forked edges,
            grow'=0,
            draw,
            rounded corners,
            node options={align=center,},
            text width=2cm,
            s sep=4pt,
            calign=child edge, 
            calign child=(n_children()+1)/2,
            l sep=7.5pt,
        },
        [, phantom
            [Data \S\ref{sec:data}, for tree={data}
                [Unstructured, data
                    [\citet{wikihowMultilingualSummary,wikiHowStepInference,DeScript}, data_work]
                ]
                [Event-centric, data
                    [\citet{FengZK18,ParkM18,ZengTNDLX20}, data_work]
                ]
                [Entity-centric, data
                    [\citet{OpenPI,zhang-etal-2024-openpi2,ProPara1}, data_work]
                ]
                [Symbolic, data
                    [\citet{PayanMSNPBRCDN23,WebArena23,ALFRED20}, data_work]
                ]
            ]
            [Grounded Tasks \S\ref{ssec:grounded_tasks}, finetuning
                [Dialogue Agent, finetuning
                    [{ABCD~\citep{ABCD}; \\ TEACh~\citep{TEACh}}, finetuning_work]
                ]
                [Web Agent, finetuning
                    [{WebArena~\citep{WebArena23}; \\ Mind2Web~\citep{Mind2Web2023}}, finetuning_work]
                ]
                [Navigation Agent, finetuning
                    [{ALFRED~\citep{ALFRED20}; VirtualHome~\citep{VirtualHome}}, finetuning_work]
                ]
                [GUI Agent, finetuning
                    [{Mobile-Env~\citep{DanyangZhang2023_MobileEnv}; \citet{bombieri2023mapping}}, finetuning_work]
                ]
                [Robotic Agent, finetuning
                    [{CALVIN~\citep{Mees2021CALVINAB}; ~\citet{TellexKDWBTR11}}, finetuning_work]
                ]
                [Game Agent, finetuning
                    [{SmartPlay~\citep{wu2024smartplay}; MineCraft~\citep{jayannavar-etal-2020-learning}}, finetuning_work]
                ]
            ]            
        ]
    \end{forest}
    \end{minipage}
    \hspace*{0.4cm}
    \begin{minipage}[b]{0.51\linewidth}
    \begin{forest}
        for tree={
            forked edges,
            grow'=0,
            draw,
            rounded corners,
            node options={align=center,},
            text width=2cm,
            s sep=4pt,
            calign=edge midpoint,
            l sep=7.5pt,
        },
        [, phantom
            [Ungrounded Tasks \S\ref{ssec:ungr_tasks}, for tree={fill=red!45,model}
                [Summariz., model
                    [\citet{Boni2021HowSummAM,wikihowSummary}, model_work]
                ]
                [Event Alignment, model
                    [{DeScript~\citep{DeScript}; \\ \citet{DonatelliSBKZK21,wanzare-etal-2017-inducing}}, model_work]
                ]
                [Implicity Detect./ Correct., model
                    [{wikiHowToImprove~\citep{AnthonioBR20}; \\ ULN~\citep{FengFLW22}}, model_work]        
                ]
                [Entity Tracking, model
                    [\citet{KimS23,SconeLongPL16}, model_work]  
                ]
                [Parsing, model
                    [Process Structure Induction, model
                        [\citet{zhang-etal-2020-analogous,WuZHSP23}, model_work2]
                    ]
                    [Process Structure Extraction, model
                        [\citet{textMining23,MysoreJKHCSFMO19,jermsurawong-habash-2015-predicting}, model_work2]
                    ]
                ]
                [Process/Step Generation, model
                    [\citet{yuan-etal-2023-distilling,proScript}, model_work]  
                ]
                [Question Answering, model
                    [Reasoning, model
                        [\citet{tandon-etal-2019-wiqa,zellers-etal-2019-hellaswag}, model_work2]
                    ]
                    [Reading Comprehension, model
                        [\citet{Mcscript,Bolotova-Baranova23}, model_work2]
                    ]
                ]
                [Knowledge Acquisition/Mining, model
                    [\citet{JungRKM10,ChuTW17}, model_work] 
                ]
            ]
        ]
    \end{forest}
    \end{minipage}
    }
    \caption{\clrev{Taxonomy of instructional text research by means of data representation schemes and downstream tasks. \textbf{Data} representations include 
Unstructured, Event-centric, Entity-centric and Symbolic formats. 
\textbf{Tasks} are split into two main categories: Grounded (Dialogue/Web/Navigation/GUI/Robotic/Game Agents) 
and Ungrounded (Summarization, Event Alignment, Implicit Instruction Detection/Correction, 
Entity Tracking, Parsing, and Question Answering).}}
    \label{fig:typo-instruction}
    \vspace{-0.25cm}
\end{figure*}

The main goal of this survey is to equip the AI---in particular \gls{NLP}---researchers with the necessary background knowledge and provide guidance on the future challenges and opportunities for conducting research on complex instructions. In contrast to existing surveys on tangential areas~(see \S\ref{sec:relWork}), we provide a holistic presentation on \textbf{available resources} and \textbf{the range of tasks} related to instructional text, making connections to other fields such as robotics, business intelligence, and computer vision. With this survey, we aim to answer the following research questions:
\begin{itemize}
    \item \textbf{RQ1:} What are the most \textbf{common ways to represent} long-form, a.k.a., multistep, \textbf{instructional text} across different disciplines? What corpus/corpora (raw or structured) are available for various representation schemes?
    \item \textbf{RQ2:} Which \textbf{downstream tasks} are readily available on instructional text? How do they differ by means of domain, methods and evaluation metrics?
    \item \textbf{RQ3:} \clrev{What \textbf{recurring challenges} persist across tasks despite methodological advances, and what do these patterns reveal about fundamental gaps in current approaches?}
\end{itemize}
In Sec.~\ref{sec:relWork}, we identify the related areas and existing surveys that complement this work. Next in Sec.~\ref{sec:method}, we define our PRISMA-based survey methodology~\cite{page2021prisma} \clrev{and present the bibliographic characteristics of the reviewed papers}. Sections \ref{sec:data} and \ref{sec:tasks} aim to answer \textbf{RQ1} and \textbf{RQ2} respectively. We identify the common themes, and remaining challenges separately for each task in Sec.~\ref{sec:tasks} to answer \textbf{RQ3}. We present a taxonomy for the data representation types, and the range of tasks to guide the reader with \cref{fig:typo-instruction}.

\section{Related Work}
\label{sec:relWork}
\clrev{The goal of this survey is to provide an interdisciplinary---in particular NLP, robotics, business intelligence and computer vision---view on instructional text research. In other words, it focuses on i) how different fields represent instructions (e.g., unstructured, event-centric, entity-centric and symbolic); ii) how instructional text is used across fields (e.g., entity-centric data $\rightarrow$ entity state tracking, symbolic data $\rightarrow$ robotic navigation) and iii) identifying recurring challenges and outlining a roadmap for future research. Due to the large number of data representation formalisms and associated downstream tasks surveyed, there are many adjacent fields that might intersect with the surveyed content such as semantic parsing. This work does not serve as a comprehensive guide on each of the adjacent fields (e.g., another survey on semantic parsing), but rather a guide that talks about those fields \textbf{in the context of multi-step, complex instructional text} (e.g., instruction parsing).} We identify the two most related fields to procedural language understanding as\footnote{\clrev{We use the terms \textit{complex instructions}, \textit{procedure}, \textit{process}, and \textit{script} interchangeably to refer to sequences of natural language directives guiding task completion; terminology varies by field (NLP: instructions/procedures; cognitive science: scripts; business intelligence: processes).}}: i) event understanding, and ii) grounding.

\paragraph{\textbf{Event Understanding}} Event-centric \gls{NLP} field primarily deals with extracting event information from textual documents. \citet{xiang2019survey} compile various approaches and challenges for event extraction from textual data, providing an overview of tasks, methods, and performance metrics. The survey categorizes techniques ranging from pattern matching to advanced machine learning models. \citet{ACL21-event-centric-NLP-tutorial} offer a comprehensive tutorial\footnote{\href{https://cogcomp.seas.upenn.edu/page/tutorial.202102/}{Cognitive Computation Group}} on event-centric information extraction, prediction, and knowledge acquisition, focusing on event-centric tasks and methodologies. In addition, \citet{li2022survey} survey the deep learning techniques for event extraction, exploring sophisticated models that detect, categorize, and analyze events across various domains. \clrev{\textbf{Semantic parsing}, a subfield of event understanding, aims to map natural language to logical forms or database queries. The subfield has a rich history and a large number of dedicated surveys~\cite{Kamath2018SemanticParsSurvey, li-etal-2020-context, chen2025semantic} focusing on semantic formalisms (e.g., abstract meaning representation) and parsing methodology. Unlike these surveys, we focus on the higher-level relations between events, participants and their environments in long procedural text rather than extracting atomic event information.}
\paragraph{\textbf{Grounding Instructions}} In broad terms, grounding generally refers to a type of task that involves connecting language to some form of external knowledge or real-world context such as images, knowledge bases, robot arms and even operating systems. \citet{ChanduBB21} discuss the evolution of the term ``grounding'' and draw connections to cognitive science. More recently, several surveys have been published on specific environments and grounding types. For instance, \citet{roboticGrounding24} survey different meaning representations for grounding robotic language for navigation/manipulation tasks, while, \citet{WangMFZYZCTCLZWW24} discuss various aspects of \gls{LLM} agents on grounded tasks. In contrast, our objective is to survey the wide range of applications and environments to ground complex instructions rather than focusing on one environment or approach. \clrev{Finally, one relevant literature to grounded tasks is code generation, i.e., program synthesis. Here, similar to Web or mobile environments, the goal is to generate code that can be executed by a symbolic interpreter, e.g., Python, SQL, etc., on a certain environment. However, the focus is learning the conversion process in an offline manner through large annotated sets, rather than through environment interaction~\cite{PayanMSNPBRCDN23}. We refer readers to comprehensive surveys on program synthesis~\cite{CodeSurvey241, ZanCZLWGWL23}.} 

To the best of our knowledge, there is no survey focusing on procedural text. The only resource is the tutorial by \citet{ProcedureTutorial} that compiles a set of selected resources containing procedures, along with a set of selected applications. In contrast, we provide a \textbf{systematic} methodology and taxonomy covering a considerably wider range of representation types, resources and tasks for procedural text. We also extend the scope to other fields such as robotics, business intelligence, and computer vision, aiming to provide a unified perspective across disciplines. In addition, our survey is related to \textbf{scripts}---a sequence of events with multiple actors---and \textbf{planning}, i.e., generating a feasible (and hopefully minimal) sequence of steps to achieve a specific goal. Although we mention them where related (e.g., script generation, Web agents), we consider them to be outside the scope of this survey, and refer the readers to the classical book by \citet{schank2013scripts}. \clrev{Finally, the boundary between the discussed paradigms is not absolute. For instance, work on semantic parsing for robot commands~\cite{Tellex2011ApproachingTS, Matuszek2012LearningTP} bridges both generating executable symbolic representations while learning from environmental feedback.} 
  
\section{Methodology}
\label{sec:method}
\begin{figure}[h]
\centering
\includegraphics[width=0.7\textwidth]{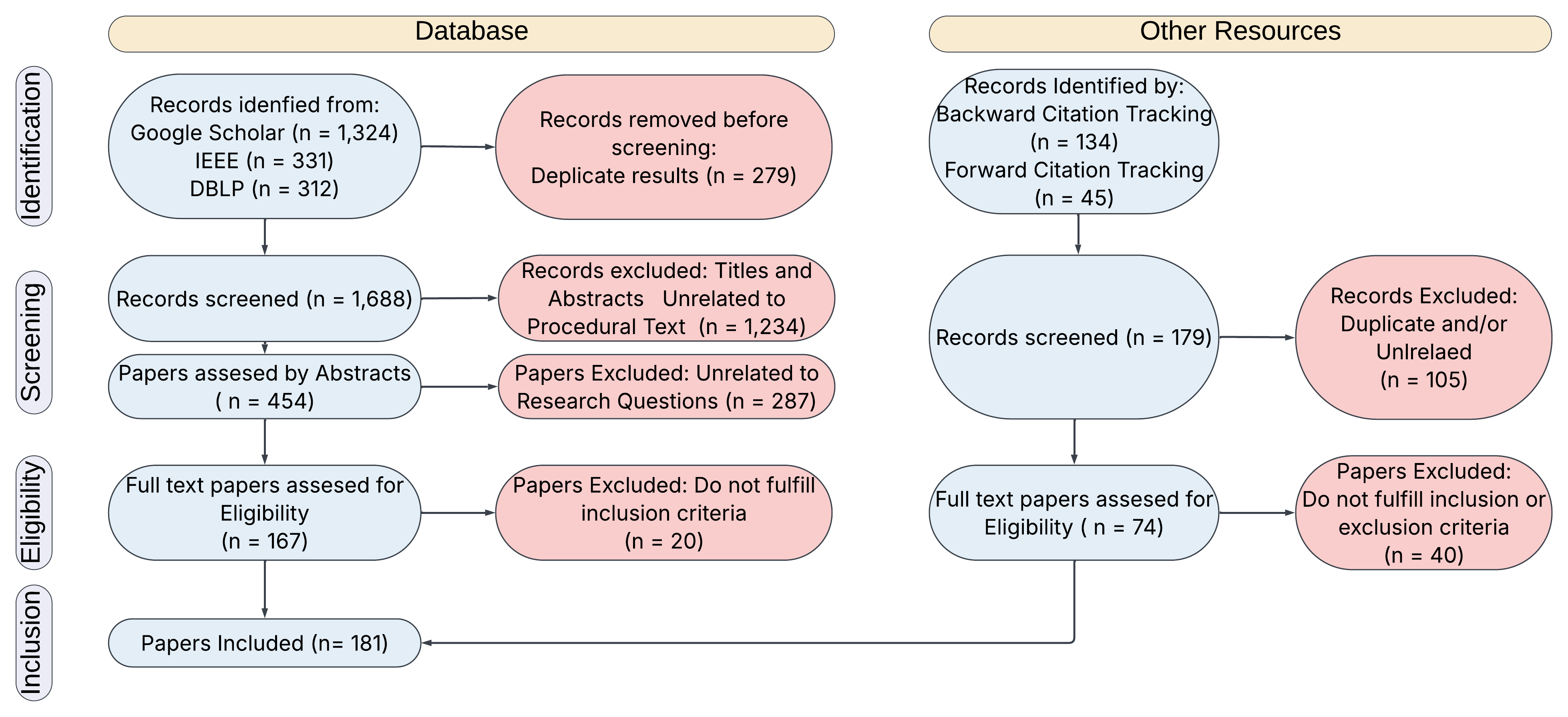}
\caption{Study protocol}
\label{fig:study_protocol}
\end{figure}

We have followed the guidelines given in the PRISMA statement~\cite{page2021prisma,page2021prisma2} to conduct this systematic review. The overview of our methodology is given in Fig.~\ref{fig:study_protocol}. It contains four steps: i) identification ii) screening iii) eligibility check and iv) inclusion, which are explained below.

\subsection{Identification}
   
     We selected various libraries to ensure broad coverage of publications in fields relevant to this survey, such as AI and its subfields—particularly NLP, robotics, machine learning, and industrial engineering. \href{https://dblp.org/faq/What+is+dblp.html}{DBLP} indexes key computer science works, while \href{https://ieeexplore.ieee.org/Xplorehelp/overview-of-ieee-xplore/about-content}{IEEE Xplore} provides a wider range of scientific content. \href{https://scholar.google.com/intl/en/scholar/publishers.html\#overview}{Scholar} primarily indexes peer-reviewed papers without field distinction. Given the rapid publication rate in NLP, we also considered notable non-peer-reviewed ArXiv papers. The review period spans 2010 to the present to capture relevant publications. We used the Google Scholar API from \href{https://serpapi.com/}{Serp API} for our search, manually extracting other digital library entries. Since Google Scholar API lacks abstract retrieval, we sourced missing abstracts from Semantic Scholar. Full list of queried keywords that are based on research questions to capture relevant papers while maintaining a manageable volume are given in App.~\ref{app:db-query}. 

      \begin{figure}[h]
        \centering
            \includegraphics[width=0.65\textwidth]{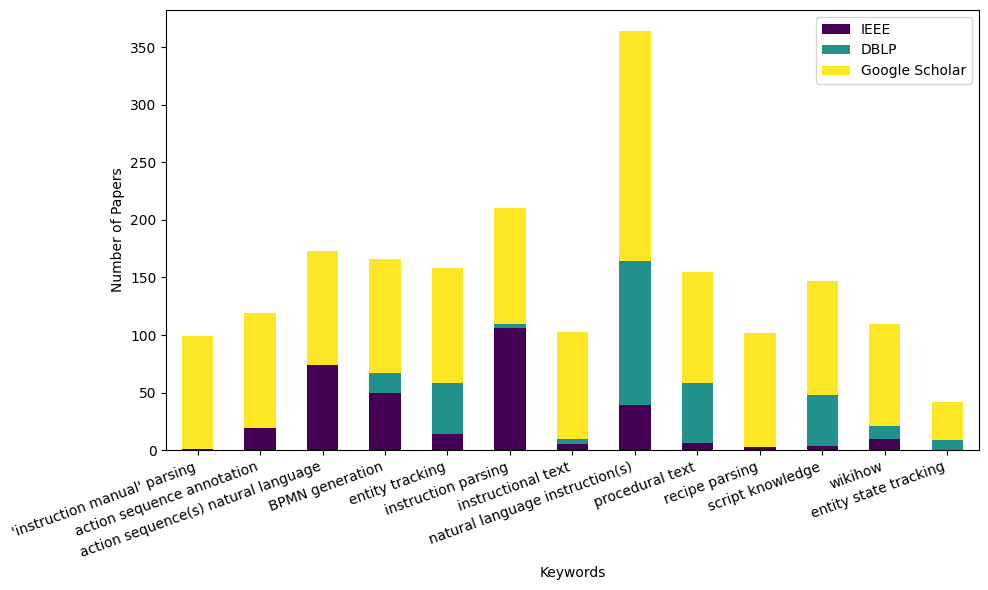}
        \caption{Publications Per Keyword Per Database}
        \label{fig:pub_per_keyword_db}
     \end{figure}
     
     Fig.~\ref{fig:pub_per_keyword_db} shows the distribution of retrieved publications across different keywords and databases after deduplication. An \gls{EPM} searches for the exact phrase in papers, while \gls{EWM} matches individual words. \gls{EPM} is supported by Google Scholar and IEEE Xplore, whereas DBLP supports only \gls{EWM}. To account for variations in terms (e.g.,``instruction'' vs. ``instructions''), we searched for both forms and combined the results. The AND operator was used to combine search terms for Google Scholar and IEEE Xplore, while DBLP was searched for each word individually with the AND operator. From all digital libraries, we retrieved up to a hundred search results per keyword returned by the database, as papers beyond were highly irrelevant. The execution of search queries in the four digital libraries returned 1,967 papers. After deduplicating the merged results from the databases, 1,688 papers remained. As expected, Google Scholar and IEEE Xplore return more results since they index a broader range of venues and domains, while DBLP only indexes computer science papers. We observe that the most number of papers are retrieved via the keyword ``natural language instructions'' since it covers various applications and methodologies. The least number of papers are retrieved for more specialized, narrow fields such as ``entity state tracking'' and ``recipe parsing''. 

\subsection{Screening}
    
    Our screening process involved two phases. First, we manually reviewed titles and abstracts to remove irrelevant papers, starting with an initial pool of 1,688 papers. During this inspection, we noticed that papers relevant to our \textbf{RQ1} and \textbf{RQ2} often included keywords like ``dataset,'' ``evaluation set,'' ``benchmark,'' or ``test set.'' This makes sense, as the studies of interest typically introduce new datasets on instructional text or novel downstream tasks using existing resources, or sometimes both. These papers were then further filtered based on these keywords, reducing the count to 454.

\subsection{Eligibility}
    Assessing the eligibility entailed manual inspection. Assigned researchers carefully examined the abstracts of the filtered papers to determine their alignment with our research inquiries. This manual review significantly narrowed down the selection to 167 papers.

\subsection{Inclusion}
    We curated a shared library of relevant papers for comprehensive full-text review, refining the selection to 147 papers. After retrieving the papers from digital libraries, we performed forward and backward citation tracking, identifying 34 additional papers, resulting in a final count of 181 papers. Once the list was finalized, we tagged the papers based on tasks, data representations, and research locations. This tagging helped us develop a taxonomy of representation types and downstream tasks (related to \textbf{RQ1} and \textbf{RQ2}) and \clrev{visualize the bibliographic properties of the publications}.

\subsection{Bibliographic Properties}
\begin{figure}[h]
    \centering
        \includegraphics[width=0.6\textwidth]{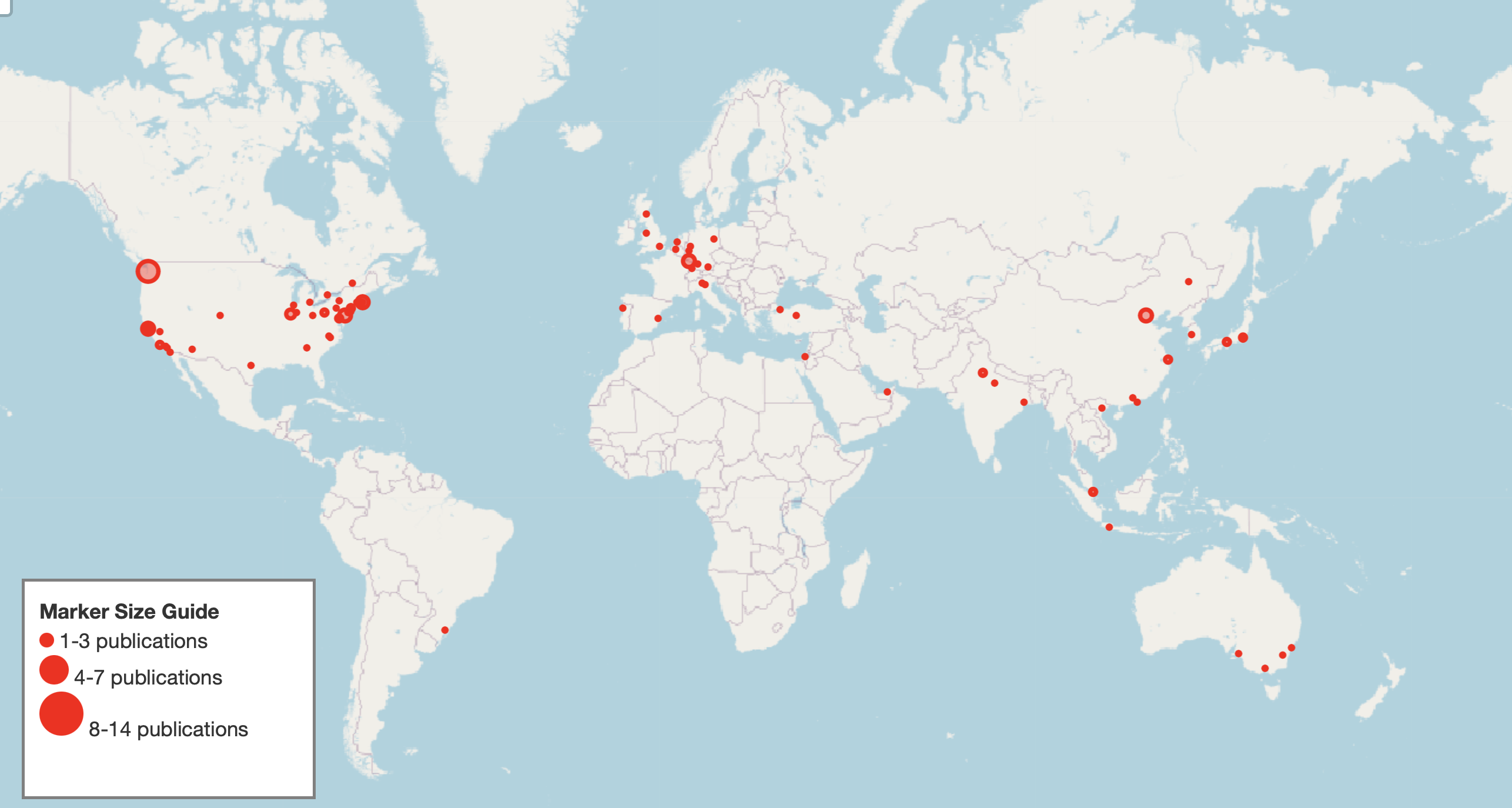}
    \caption{Geographical distribution of publications}
    \label{fig:pub_origin}
    \end{figure}

    To gain a comprehensive view of the instructional text processing research landscape, we visualized various data aspects. Fig.~\ref{fig:pub_origin} shows the geographical distribution of publications, with a large number coming from renowned research hubs like Stanford, Seattle, New York City, and Beijing. This highlights the concentration of research efforts in urban centers known for their academic institutions. Notably, Stanford and Seattle lead in publication numbers, reflecting strong research output in \gls{NLP}, especially in procedural text fields. This dominance, also seen in other AI subfields\cite{Breit2023CombiningML}, underscores the lack of diversity in languages and research groups.

    \begin{figure}[h]
    \centering
    \begin{subfigure}[b]{0.45\textwidth}
        \includegraphics[width=\linewidth]{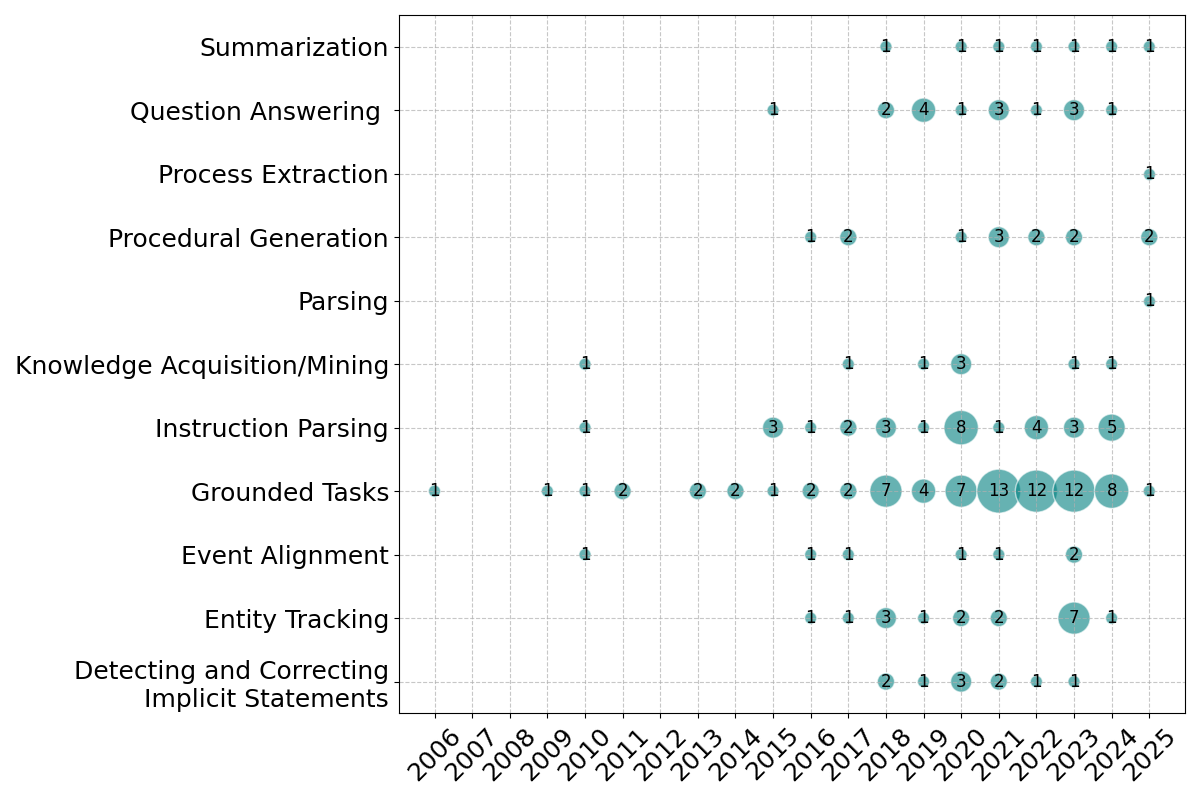}
        \caption{Papers Per Task Distribution over Time}
        \label{fig:task-year}
    \end{subfigure}
    \hfill 
    \begin{subfigure}[b]{0.45\textwidth}
        \includegraphics[width=\linewidth]{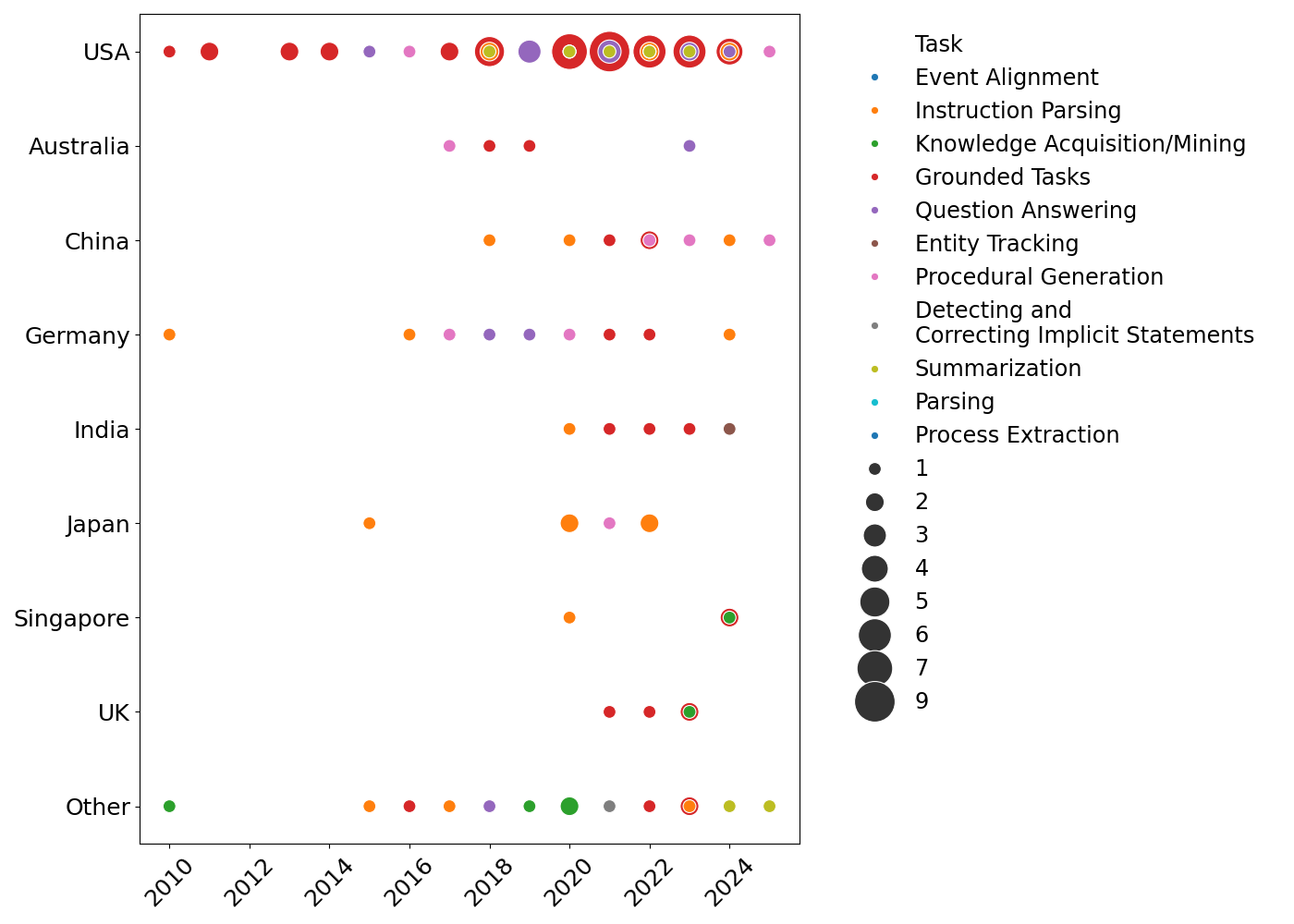}
        \caption{Papers from each country Per Task Distribution over Time}
        \label{fig:task-year-country}
    \end{subfigure}
    \caption{Comparison of Task Distribution and Country Distribution over Time}
    \label{fig:task-country-comparison}
\end{figure}

    Fig.~\ref{fig:task-year} shows the temporal distribution of tasks, illustrating a steady growth in \textit{Summarization} and \textit{Question Answering}, with a notable peak in the latter starting from 2018. Newer fields such as \textit{Grounded Tasks} have seen a significant rise in interest since 2017, reflecting a shift towards more contextual and task-driven AI applications. The increasing trend in \textit{Instruction Parsing} from 2018 onward indicates renewed interest in more structured task interpretation. Meanwhile, \textit{Procedural Generation} has seen a consistent, albeit slower, development. Traditional areas such as \textit{Entity Tracking} and \textit{Event Alignment} have remained relatively steady, while areas like \textit{Knowledge Acquisition/Mining} have gained more prominence in recent years. The growth from 2018 onwards, particularly in these emerging areas, suggests a trend toward more dynamic and interactive AI systems.

    Fig.~\ref{fig:pub_origin} and \ref{fig:task-year-country} illustrate the global scope of instructional text research, with certain countries excelling in specific areas. The USA pioneered Step Inference in 2006 and leads in \textit{Grounded Tasks} and Instruction Parsing. Germany has focused on \textit{Instruction Parsing} since 2016. China, active since 2018, has expanded into \textit{Instruction Parsing} and \textit{Entity Tracking}. Japan and India contribute to \textit{Summarization} and \textit{Procedural Generation}. Key research hubs include San Francisco, Berlin, and Beijing. These findings suggest potential collaborations, such as the USA and Germany on Parsing tasks or China and India on domain-specific applications, to advance global instructional text research. \clrev{It should be noted that the geographic and temporal distributions are influenced by our choice of databases (Google Scholar, IEEE Xplore, DBLP) and the dominance of English as the primary language in AI research. Despite our best efforts, our selection criteria may not comprehensively capture work from under-represented communities who might publish elsewhere (e.g., local venues for cost efficiency) in their local language; and under-represented fields such as process extraction in business intelligence, where research might appear in smaller venues using varying terminologies.}

\section{Data}
\label{sec:data}
Even though a substantial amount of work for procedural text relies on \textbf{unstructured} corpora derived from web sources like WikiHow\footnote{\url{https://www.wikihow.com/}}, the field is not short of structured, labeled datasets. We categorize structured representations into three main categories: i) \textbf{event-centric} (see \S\ref{ssec:task_repr}) ii) \textbf{entity-centric} (see \S\ref{ssec:ent_repr}) and iii) \textbf{symbolic} (see \S\ref{ssec:symbolic}). Event-centric representations mostly ignore the properties of entities, and how certain events affect those. The entities are mostly referred to as ``the argument'' of the event---a.k.a., action---and are represented as free text. Instead, they put the events/tasks into focus and define a rich set of relations between them (e.g., subsequent, conditional, concurrent, parent/child, etc.). On the other hand, entity-centric representations predefine intrinsic properties of entities (e.g., color, location) and track the state changes caused by extrinsic and intrinsic events. The events are mostly not tied to any static knowledge base or lexicon and are simply represented as free text. Finally, symbolic representations aim to transform the natural language instructions into a machine-readable format, such as first-order logic, robot actions, and UI actions, so that they can be executed/performed on a specific \clrev{simulated} environment. They are mostly used for grounding tasks~(see \S\ref{sec:tasks}). 

\clrev{Most \textbf{event-centric} representation schemas are linguistically motivated. For instance, SIMMR~\cite{jermsurawong-habash-2015-predicting} define a dependency grammar to represent action-entity relations in recipes. Similarly, a large number of studies (e.g., \citet{FengZK18,MysoreJKHCSFMO19}) employ semantic representations inspired from common semantic formalisms (e.g., PropBank, FrameNet) with varying levels of  granularity. Although most entity-centric representations avoid linguistic formalisms, one exception~\cite{clark2018leveraging} leverages \gls{VerbNet} as an additional background knowledge for effect (i.e., postcondition) prediction. While linguistically motivated frameworks provide foundational tools and inspiration for researchers in the field, procedural text often requires domain-specific adaptations or extensions---such as fine-grained domain-specific arguments~\cite{MysoreJKHCSFMO19}, exclusive arguments~\cite{FengZK18}, concurrent event relations~\cite{proScript,jermsurawong-habash-2015-predicting}, and implicit entity handling~\citet{kiddon-etal-2015-mise}. Notions such as exclusive arguments, richer event-event relations, and implicit entities are not fully handled by traditional semantic formalisms; and are active area of research.}

    \subsection{Unstructured}
    \label{ssec:unstr_data}
      The field has mainly used three techniques to create unstructured corpora to study procedural text: a) scraping/filtering available web resources, b) crowdsourcing, and c) synthetic data generation.

      \subsubsection{Web Corpora} 
      
        The most popular way to construct large-scale corpora that can be repurposed (or annotated) for many downstream tasks has been scraping websites that contain how-to information. WikiHow has been exploited tremendously to study procedural, a.k.a., instructional text. Even though it is referred to as ``unstructured'', WikiHow articles follow strict writing style, possessing a certain file structure. For instance, each article contains a goal \textit{(e.g., How to destroy an old computer?)}, several methods to reach the goal \textit{(e.g., destroying a computer entirely, destroying a computer for recycling)} and steps to execute the method \textit{(e.g., 1. Wipe your hard drive, 2. Remove battery, etc.)}. Furthermore, each article has a community Q\&A and warning/tips sections for people following the article to seek help. WikiHow has a considerably high domain coverage---from make-up tips to writing indie songs. Articles are curated and edited by experts from all around the world in different languages. Nonetheless, similar to other web corpora, it is \textbf{predominantly in English}. Furthermore, most articles contain pictures that accompany the step description, which makes WikiHow suitable for multimodal studies.
        
        In summary, WikiHow has been the most popular resource due to i) the strict style followed by the authors that enabled easy scraping, ii) high domain coverage (19 domains, and subdomains), iii) large-scale (e.g., more than 235,000 articles), iv) multilinguality, v) multimodality and vi) high quality (i.e., expert curated articles). In addition to WikiHow, \href{https://instructables.com}{Instructables}, \href{https://www.ifixit.com}{ifixit} and  \href{https://www.ehow.com/}{ehow}, several recipe websites such as \href{https://www.allrecipes.com/}{Allrecipes}, \href{http://cookpad.com/}{Cookpad}, \href{http://www.haodou.com/recipe/}{HaoDou}, \href{https://www.food.com/}{Food} and \href{https://m.xiachufang.com/}{Xiachufang} have been used~\cite{RecipeQA,nabizadeh-etal-2020-myfixit, LiuGPSL18, MIAIS, diwan2020named, JungRKM10, RecipeNER}.

        Web corpora are used in many shapes and forms, i.e., for pretraining or finetuning language models to enhance reasoning or planning capacities. It is also repurposed for a variety of downstream tasks, such as summarization~\cite{wikihowMultilingualSummary}, question answering~\cite{wikiHowStepInference,zellers-etal-2019-hellaswag,yang-etal-2021-visual}, classification~\cite{zhou-etal-2019-learning-household,ParkM18}, detecting implicit information in instructions~\cite{AnthonioBR20}, learning task hierarchy~\cite{zhou-etal-2022-show}, generation (e.g., next step, all steps)~\cite{lyu-etal-2021-goal,NguyenNCTP17} and benchmarking procedural language understanding and planning abilities~\cite{UzunogluS23,uzunoglu-etal-2024-paradise} without any additional annotation layers. Even though recipes and WikiHow are mostly exploited, we find that \textbf{many resources such as troubleshooting websites from tech companies} (e.g., \href{https://oip.manual.canon/FAQ01-0111-zz-DR-enUV/?type=}{Canon} \textbf{are overlooked} in the literature with some exceptions~\cite{TechTrack}. The reasons might be i) their small size ii) the lack of a consistent structure that makes them hard to parse. Another \textbf{overlooked resource is local websites} (e.g., \href{https://www.nefisyemektarifleri.com/}{Turkish food recipes}), which might be due to the lack of enough NLP experts or interest in local communities.    
      
      \subsubsection{Crowdsourcing} 
        Researchers have used \textbf{crowdsourcing to create smaller, but more targeted datasets}. For instance, DeScript~\cite{DeScript} is a small corpus of scenarios on everyday tasks such as going shopping, riding a bus, getting a haircut and taking a bath. It contains a rather limited number of tasks (only 40), however, each one has been described step-by-step by 100 different crowdtaskers. Steps, a.k.a., event sequence descriptions, written by different taskers are later aligned with each other. Several other studies such as Task2Dial~\cite{strathearn-gkatzia-2022-task2dial} and ABCD~\cite{chen-etal-2021-action}, employ crowdworkers to generate dialogue datasets grounded on instructional documents such as recipes and call center guidelines by assigning different roles to crowdworkers (e.g., call center employee, information giver on a certain recipe). It is also commonly used to annotate existing small corpora for a specific task. Such tasks are mostly related to extracting some form of information from instructional text, e.g., tools from fixing manuals~\cite{nabizadeh-etal-2020-myfixit}, ingredients from recipes~\cite{MIAIS}. Due to the costs associated with crowdsourcing, this technique has been mostly used for generating scripts, grounded dialogues or to add small annotation layers to existing corpora. We find that, even though, for example, the corpus \href{https://fedora.clarin-d.uni-saarland.de/sfb1102/}{DeScript~\cite{DeScript}} and the alignments are publicly available, \textbf{such resources are overlooked and not used for related downstream tasks e.g., event paraphrasing or alignment} to the best of our knowledge, with some exceptions~\cite{wanzare-etal-2017-inducing}.
      
      \subsubsection{Synthetic data generation} 
        Synthetic datasets and environments are also common for creating toy setups under simplifying assumptions. For example, \citet{bAbI} formulate 20 question answering tasks to evaluate different linguistic and reasoning abilities such as co-reference resolution, temporal and spatial reasoning. The stories are generated in a simulated world which is defined by manually written rules \textit{(e.g., one should find food if hungry)} on a predefined set of entities with predefined attributes such as location and size. Similarly, \citet{KimS23} define a set of entities (e.g., book, hat, etc.), properties (e.g., location) and a small set of actions (e.g., MOVE) to generate synthetic procedures about moving entities between different boxes via a simple Python script. It is then used to evaluate the entity state tracking abilities of large language models. Another research line develops instruction following game environments such as TextWorld~\cite{TextWorld} as a testbed for reinforcement learning agents to measure their planning and exploration abilities. Similar to others, the environment is defined via a fixed set of rooms, entities, actions, and entity properties; however, the game is interactive and played step-by-step. 
        
        Synthetic generation gives researchers the ability to \textbf{control} the complexity of the task and the environment. However, synthetic datasets mostly \textbf{lack the linguistic diversity, and long-tail, rare events that cannot be programmed}. In other research areas of \gls{NLP}, it is common to post process the synthetic data via asking crowdsourcers to \textbf{paraphrase the synthetic text~\cite{KimS23} to increase linguistic diversity. However, we don't observe the same trend for procedural text literature, which could be a promising future direction}. Similar to web corpora, synthetic data is also commonly used for pretraining models. For instance, \citet{Lemon22} create a large-scale synthetic dataset by randomly sampling the environment states and associated programs to pretrain a BART-large model, then evaluate on entity state tracking test splits.   
        
      \subsubsection{Generation with \gls{LLM}s}
        One of the latest trends in the field is to \textbf{exploit large language models to generate silver, large-scale datasets}. Instructional text literature also \textbf{follows this trend}. For instance, \citet{yuan-etal-2023-distilling} create a constrained planning dataset~(CoScript) of 55,000 procedures, e.g., procedures to make a cake for \textbf{diabetics}, by first generating with InstructGPT and then filtering with constraints.
 
    \subsection{Event-centric representation}
    \label{ssec:task_repr}
        Despite the differences in naming conventions, we use event, a.k.a., instruction, to refer to a step in a procedure that contains an action phrase to accomplish a subgoal \textit{(e.g., Bake in the oven)}. We define action verb as the predicate (e.g., Bake), and the arguments of the predicate (e.g., it=the thing that is baked, oven=Location) as the entities. In Table~\ref{tab:rep_format}, we demonstrate the diversity of representations from different angles, on a set of work that is manually selected to give the best overview. We determine the differences to be along the formats used for the \textbf{actions} and \textbf{entities}, and the \textbf{entity roles w.r.t the action}, and \textbf{event-event relations}. 
        
        We note that, \textbf{business information literature} that employs \gls{BPMN} \textbf{uses a different terminology}. Here, an event is represented as an empty circle to mark the start, middle and end of the process; while the event is called an action or an activity. Examples of event-centric representations, such as `baking a cake' and the \gls{BPMN}, can be found in App.~\ref{app:event-centric-represent}.
        
        \begin{table*}[!htp]
            \scalebox{0.65}{
            \begin{tabular}{lp{0.15\linewidth}p{0.15\linewidth}p{0.25\linewidth}p{0.15\linewidth}c} 
                \toprule
                 & Action & Entity & Action-Entity & Event-Event & OD \\ 
                \midrule
                Wiki HG~\cite{FengZK18} & Text & Text & Ex, Es, Op & Suc, Ex & \cmark \\
                MS~\cite{MysoreJKHCSFMO19} & Text & Tagged & Tagged & Suc & \xmark \\
                SIMMR~\cite{jermsurawong-habash-2015-predicting} & Tagged & Tagged & Input, Output & Suc, Con & \xmark \\
                Action Graph~\cite{kiddon-etal-2015-mise} & Text+Implicit & Text, Food, Location & \xmark & \xmark & \xmark \\
                \midrule
                APSI~\cite{zhang-etal-2020-analogous} & Text & \xmark & \xmark & Suc & \cmark \\
                proScript~\cite{proScript} & Text & \xmark & \xmark & Suc, Con & \cmark \\
                InScript~\cite{InScript} & Tagged & Tagged & \xmark & Suc & \xmark \\
                \midrule
                Process Graph~\cite{ParkM18} & Text & Text & Actor, Time, Location & Suc, Opt, Con, PC, If-Else & \cmark \\
                Process Model~\cite{ProcessModelExtraction} & Text & Text & Object & Suc, Opt, Con & \cmark \\
                BPMN~\cite{Zeng2020Missing} & Text & \xmark & Object & Suc, Ex, Con & \cmark \\
                \bottomrule
            \end{tabular}
            }
            \caption{Common task-centric representation schemes for procedural text. OD: Open domain, i.e., no domain-specific tags. Ex: Exclusive, Es: Essential, Opt: Optional, Suc: Successive, Con: Concurrent, PC: Parent-Child, Pre: Precondition, Eff: Effect}
            \label{tab:rep_format}
        \end{table*}
        
        \subsubsection{Actions and Entities} 
            We use event, action and task interchangeably following the previous work since the \textbf{granularity of the task representations varies greatly}. They refer to \textbf{a single step in the instruction} that needs to be performed to accomplish a goal. Furthermore, the majority of the work makes a simplifying assumption that a step/task contains only one action.
            For instance, \citet{proScript} take the \textbf{full step} \textit{(e.g., bake for the right amount of time)} as the task representation, while \citet{zhang-etal-2020-analogous} represent them as an \textbf{action phrase} that contains only one predicate and argument \textit{(e.g., search car, apply loan)}. Another line of work~\cite{ParkM18,ProcessModelExtraction,Zeng2020Missing} use the \gls{BPMN}~\cite{KocbekJHP15}---an industry standard---and similarly \textbf{represent a task as a predicate-argument pair}, where the argument is always an object. All the studies mentioned above use a free text representation, (denoted as ``Text'' in the Action and Entity columns of Table~\ref{tab:rep_format}), sometimes limiting the number of actions to the most frequent ones. \textbf{This assumption might be problematic, especially when the articles contain different author styles} as in WikiHow, or DeScript. To address that, \textbf{several studies limit the domain and use domain-specific tags}. For instance \citet{InScript} define \textbf{scenario-specific events} (e.g., apply-soap, undress, turn water on) and tag all the action phrases. Similarly, SIMMR~\cite{jermsurawong-habash-2015-predicting} use a \textbf{predefined recipe-specific language} named MILK to represent the actions and ingredients. These are noted with ``Tagged'' in the Action and Entity columns of Table~\ref{tab:rep_format}.
            
            The studies that use free text representation mostly annotate the indices of the objects of interest (e.g., action verb, action phrase, argument etc...)~\cite{FengZK18,MysoreJKHCSFMO19}, 
            or simply extract them with existing tools like semantic role labelers~\cite{zhang-etal-2020-analogous,ParkM18,ProcessModelExtraction}. \textbf{Majority presume that all actions are explicitly stated in the text}.
            
            Although there is a growing body of work on implicit instructions~(see \S \ref{sssec:implicit}), \textbf{event-centric representations mostly ignore implicit entities}. For instance, imagine the phrase \textit{``Mix a, b, c and then bake''}. The thing that is baked is the implicit, unmentioned mixture. \clrev{\citet{MysoreJKHCSFMO19} explicitly acknowledge such cases---``a material entirely absent from the text being the true argument''---but exclude them from their annotation schema, as their framework does not allow arguments to have multiple parents when entity states change. This design choice, while pragmatic, results in systematic exclusion of semantically crucial information necessary for tracking ingredient transformations through procedural chains}.

            \textbf{Entities} are marked (e.g., for the position) in the text~\cite{FengZK18,ParkM18,ProcessModelExtraction}, tagged with predefined entity types~\cite{jermsurawong-habash-2015-predicting,InScript,MysoreJKHCSFMO19}, or simply ignored~\cite{proScript,ProcessModelExtraction}. \citet{kiddon-etal-2015-mise} extract entities, however, only label a few important entity types, namely as food and location. \citet{jermsurawong-habash-2015-predicting} represent the ingredients/entities and actions of the recipe as terminal and internal nodes respectively. 
            
        \subsubsection{Action-Entity Relation} 
            The role the entity plays in the action is crucial for task representation. Several studies~\cite{MysoreJKHCSFMO19,InScript} define highly fine-grained entity types that also include information about the relation (e.g., Material\_of). \textbf{Studies on scripts~\cite{Zeng2020Missing,proScript,InScript} mostly ignore such relations, while others do not have any consensus}. For instance \citet{FengZK18} defines the arguments as exclusive, essential and optional. Imagine the phrase ``Take a pen or a pencil''. Here, pen and pencil are defined as exclusive arguments of action ``take''. Essential and optional arguments are also common in semantic schemas (e.g., Arg-0, Arg-TMP in PropBank). However the ``actor'' role rarely exists in procedural text, while the text is mostly imperative. Furthermore, \textbf{exclusive arguments are not part of the standard semantic schemes (e.g., PropBank, FrameNet, \gls{VerbNet}), however, is important for procedural text}. 
           
            Despite the representation power and ample of choices in \gls{BPMN}, the \textbf{surveyed studies mostly focus on the ``object'', ignoring other essential information such as duration and location} with some exceptions~\cite{ParkM18}. 
            Another approach is to represent such relations with unlabeled dependency links~\cite{jermsurawong-habash-2015-predicting}. Here, the entities are given as input to the action. Unlike others, this enables linking entities with multiple actions, representing \textbf{implicit} entities such as mixtures with a link towards the root and representing entities and actions on the same level. 

        \subsubsection{Event-Event Relation} 

            One important feature that \textbf{differentiates the field of procedural language understanding from event extraction is the rich set of higher-level relations between tasks/events}. \textbf{Defining procedures with only successive events has the minimal representation power, but has been the default way}. This simplistic approach is problematic and is mostly addressed by business information management, and software engineering literature via defining and standardizing richer relations such as ``exclusive'', ``successive'', ``concurrent'' and ``optional'' (see Table~\ref{tab:task_relation} for examples). Some exceptions for \gls{NLP} field are \citet{FengZK18} and \cite{proScript} including exclusive, and concurrent relations respectively. Another exception is \citet{jermsurawong-habash-2015-predicting} that can also handle concurrent events implicitly in a dependency tree structure.   
            \begin{table}[!htp]
            \centering
             \scalebox{0.65}{
                \begin{tabular}{lp{0.2\linewidth}p{0.25\linewidth}p{0.30\linewidth}p{0.25\linewidth}} 
                    \toprule
                    Relation & Event A & Event B & Explanation & \clrev{Example Paper} \\  
                    \midrule
                    Exclusive & Pay by cash & Pay by credit & Either/Or relation & Wiki-HG~\cite{FengZK18} \\
                    \midrule
                    Successive & Receive feedback & Process feedback & Event B happens after Event A & MS~\cite{MysoreJKHCSFMO19} \\
                    \midrule
                    Concurrent & Book a flight & Book a hotel & Event A and Event B occur simultaneously & SIMMR~\cite{jermsurawong-habash-2015-predicting} \\
                    \midrule
                    Optional & Book a flight & Add travel insurance & Event B is optional & Process Model~\cite{ProcessModelExtraction} \\
                    \midrule
                    Parent Child & Write a paper & Do literature search & Event B is the child process of Event A & Process Graph~\cite{ParkM18} \\
                    \midrule
                    Conditional & Application is received & Application is confirmed & There is a condition for moving from Event A to Event B, e.g., GRE>80 & proScript~\cite{proScript} \\
                    \bottomrule
                \end{tabular}
                }
            \caption{Examples of event relations from the surveyed literature}
            \label{tab:task_relation}
            \end{table}

            \textbf{Parent-child relation for procedures is also understudied}. The reasons are that procedures are mostly studied in isolation due to their challenging structure, and links to subtasks are usually nonexistent. Similarly, \textbf{conditional tasks are vastly ignored}. To study those, one needs alternative methods and the conditions to switch. Even though WikiHow contains different methods to accomplish a goal, which one to choose is not explicitly stated, and inferred by the readers.
            \textbf{Although there are exceptions, hierarchy and conditional processes are severely understudied according to our survey results}. 
        
    \subsection{Entity-centric representation}
    \label{ssec:ent_repr}
        
        \begin{table*}[!htp]
            \centering
            \scalebox{0.60}{
            \begin{tabular}{lp{0.1\linewidth}p{0.1\linewidth}p{0.1\linewidth}p{0.15\linewidth}p{0.1\linewidth}p{0.25\linewidth}p{0.1\linewidth}} 
                \toprule
                 & Entity Types & \#States & \#Actions & \#Entities & Avg \#Steps & \#Examples & OD \\ 
                \midrule
                ProPara~v1.0~\cite{ProPara1} & Multiple & 2 & 3 & <10 & >4 & 488 paragraphs & \xmark \\
                OpenPI~\cite{OpenPI} & Unlimited & Unlimited & Unlimited & Unlimited & n.a & 810 articles, 30K state changes & \cmark \\    
                NPN~\cite{BosselutLHEFC18} & Multiple & 6 & 384 & Varying & n.a & 65K recipes & \xmark \\            
                \midrule
                Boxes~\cite{KimS23} & Multiple & 2 & 3 & Varying & 12 & 2200 scenarios & \xmark \\ 
                SCONE~\cite{SconeLongPL16} & 1 & <3 & <6 & <6 & 5 & 4K scenarios & \xmark \\
                \bottomrule
            \end{tabular}
        }
        \caption{Common entity tracking datasets. OD: Open domain}
        \label{tab:rep_format_entity}
        \end{table*}
 
        In Table~\ref{tab:rep_format_entity}, we compile the common entity tracking studies and analyze them in several dimensions, such as domain coverage and the complexity (e.g., number of steps, number of actions). 
        
\begin{figure}[htbp]
            \centering
                \begin{subfigure}[b]{0.40\textwidth}
                   \includegraphics[width=\linewidth]{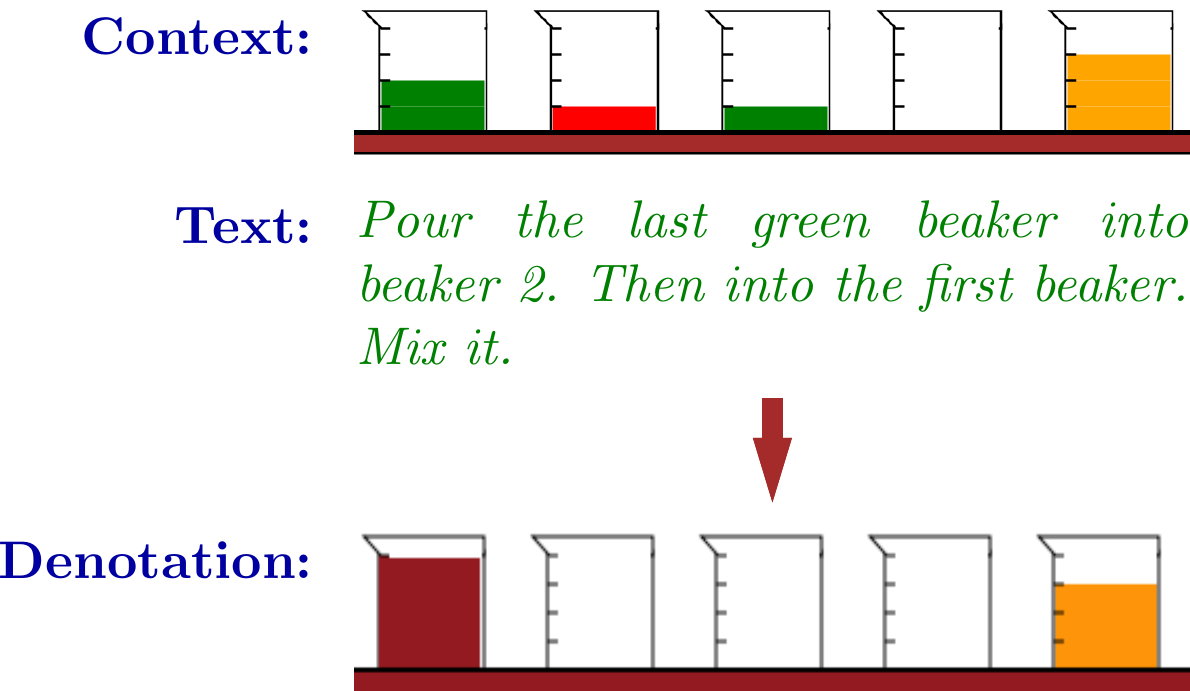}
                   \caption{An example from the ALCHEMY dataset. The content of each beaker at each step is visualized showing the results of the corresponding action.}
                   \label{fig:alchemy}
                \end{subfigure}
                \hfill 
                \begin{subfigure}[b]{0.40\textwidth}
                   \includegraphics[width=\linewidth]{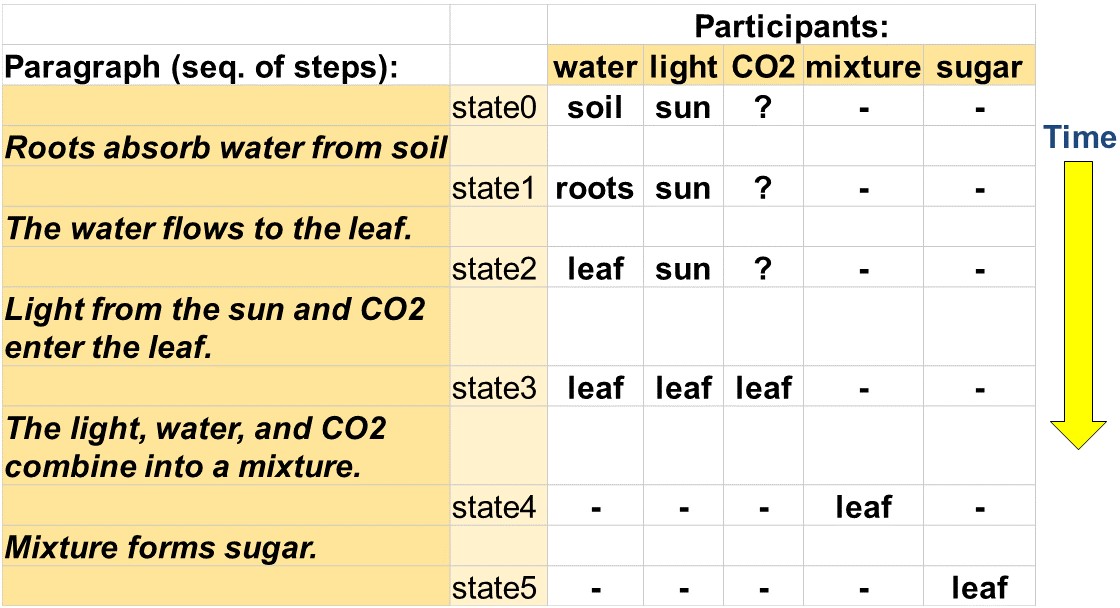}
                   \caption{An example from ProPara dataset. Each filled row shows the existence and location of participants between each step.}
                   \label{fig:tangrams}
                \end{subfigure}
            \caption{Examples of Entity Tracking Scenarios from the Alchemy~\cite{SconeLongPL16} (simulated) and ProPara~\cite{ProPara1} (crowdsourced) datasets.}
            \label{fig:entity-tracking-datasets}
        \end{figure}

        The datasets fall into two main categories: \textbf{simulated} and \textbf{crowdsourced}. Boxes~\cite{KimS23}, TextWorld~\cite{li-etal-2021-implicit} and SCONE~\cite{SconeLongPL16} use simulation to easily keep track of object states and can be easily extended. For the simulated ones, the vocabulary is generally small—approximately 1,100 to 1,200 words, the text is artificially generated, hence usually uses a simple generative grammar and might contain spurious correlations. \citet{KimS23} attempt to increase the language diversity by including paraphrases. The second category of datasets include ProPara~v1.0~\cite{ProPara1}, OpenPI~\cite{OpenPI} and NPN~\cite{BosselutLHEFC18}; and use natural language text written by experts where the entities and states are annotated by crowdworkers. Opposite to simulated text, natural procedures are linguistically rich. Thus, the same entity types or states can be referred to using different names. Recently released OpenPI-v2.0~\cite{zhang-etal-2024-openpi2} attempts to address this by clustering similar objects (e.g., spice and seasoning) together for a more uniform dataset. Fig.~\ref{fig:entity-tracking-datasets} shows two examples from simulated (left) and crowdsourced (right) datasets.
        
        \textbf{One important difference between the two approaches is the initial environment/world state}. \textbf{Simulated} datasets explicitly include \textbf{statements about the initial state} of all entities (e.g., Box 1 is empty) \textbf{while real-world procedures/processes do not} mention the initial states but are required to be inferred. For instance, for the photosynthesis process, we assume (but not explicitly stated) that, the plants are buried in the ground, they have leaves, and there is sun above etc... There are several studies~\cite{clark2018leveraging,ZhangGQWJ21} that aim to \textit{inject} such common sense knowledge through external knowledge bases. \clrev{For instance, \cite{clark2018leveraging} use \gls{STRIPS}-style Semantic Lexicons to define preconditions and effects of actions. They use commonsense persistence  algorithms to propagate states when information is missing.} For some of the domains such as cooking (e.g., Onion: NotCooked, Pan: Empty), default initial states can be easier to identify \clrev{when the task is restricted to a closed vocabulary of fixed attributes, such as existence and location (e.g., ProPara~\cite{ProPara1})}, while not so easy for other expert domains e.g., technical manuals~\cite{TechTrack}. \clrev{However, the difficulty increases significantly open-vocabulary datasets (e.g., OPENPI~\cite{OpenPI}) where entities, attributes, and state values are unrestricted}. On the other hand, \textbf{the goal is known beforehand in crowdsourced work} (e.g., how to fix Wifi adapter, produce energy by photosynthesis), although the \textbf{end states are again not explicitly stated but can be inferred} (e.g., Adapter: fixed). Meanwhile, \textbf{simulated work is mostly random simulations without any clear goal}, i.e., things moving from box to box for no reason. But the \textbf{final states are again well-defined} 
        \clrev{because they are deterministically computed based on the explicit actions and initial state definitions in the closed world model}. Another important difference is the ``validity'' of the scenarios. \textbf{Crowdsourced} procedures written by humans are \textbf{valid} by nature; however the simulated environments need to have certain measures for the validity of the generated scenario. For instance \citet{li-etal-2021-implicit} define the game as valid, if it is possible to reach the end goal from the initial state.
        
        Since these studies focus on tracking the state changes of the entities, the number of entity types and possible states are domain-specific and limited to a small number. One exception is OpenPI~\cite{OpenPI} and its derivatives~\cite{zhang-etal-2024-openpi2,OpenPIC} that consider all entity and state configurations in hundreds of WikiHow articles from various domains. The earliest work by \citet{SconeLongPL16} define only a single entity type such as a beaker, bottle or a person. Later studies~\cite{BosselutLHEFC18,ProPara1} increase the number of entity types, i.e., \citet{BosselutLHEFC18} keep a short list of ingredients, where \citet{ProPara1} focus on a small set of entities that are part of physical processes (e.g., photosynthesis) and \citet{li-etal-2021-implicit} incorporate text adventure game objects (e.g., player, chest). 

        For the sake of tractability, number of possible actions that can be applied to entities are limited to a small number (from 2 to 384). Most actions are generic---not domain dependent---such as create, remove, move; while NPN~\cite{BosselutLHEFC18} also includes several domain-specific actions (e.g., bake, boil, cut). Entity properties, a.k.a., states, are again defined as a small set of fixed states (between 1 and 6). Although they are mostly domain independent (e.g., Location, Existence), some are domain-specific (e.g., Cookedness). States can be binary (edible, isOpen, exists etc...), categorical (shape, color) or free text (location). Note that location for simulated datasets are dominantly categorical (Box1, Beaker1 etc...). 

        Another factor that affects the complexity of the task is the number of steps that cause at least one state change in the set of entities. Both for the simulated and crowdsourced datasets, the researchers use a threshold for the minimum and maximum number of steps. For instance \citet{ProPara1} sets the minimum to 4, while the simulated ones~\cite{KimS23,SconeLongPL16} explicitly set the number of steps to 12 and 5 accordingly. In theory, the simulated datasets can be easily extended, however, currently they are restricted to a few thousand scenarios (2K-4K). On the other hand, crowdsourced datasets are generally smaller, mostly containing less than 1,000 articles. 
        Finally, several datasets provide additional annotations for supporting tasks. For instance \citet{ProPara1} provides an additional dependency explanation graph to include a cause-effect relation between the sentences; and OpenPI-v2.0~\cite{zhang-etal-2024-openpi2} introduces an additional ``saliency score'', i.e., importance in accomplishing the task for entities in the range of 1-5.

    \subsection{Symbolic representation}
    \label{ssec:symbolic}
    \clrev{Entity- and event-centric representations capture procedural structure but lack executability---the ability to directly drive automated agents in interactive environments. Executable symbolic representations address this limitation through formal languages (e.g., PDDL, LTL, domain-specific action languages) that can be automatically interpreted and executed in simulation environments. These representations enable grounding natural language instructions to concrete environment actions, though this executability introduces distinct design constraints absent in purely descriptive formalisms.}
    
    This final category of work generally define---or use a predefined---formalism/language that can be automatically interpreted by an external symbolic interpreter and executed/performed on a simulation environment interactively. The type of the symbolic language depends on the environment it is interpreted. To name a few, \gls{PDDL}~\cite{ManualToPDDL}, \gls{LTL}~\cite{bombieri2023mapping}, \gls{FOL}~\cite{SconeLongPL16} are commonly used in robotic environments for navigation and manipulation tasks. UI actions related to mobile and operating systems are used for environments such as AndroidHowTo~\cite{LiHZZB20}, PixelHelp~\cite{LiHZZB20}, WinHelp~\cite{branavan-etal-2009-reinforcement}. Custom-made, simplistic languages are designed mostly for 3D simulation and game environments like ALFRED~\cite{ALFRED20}, VirtualHome~\cite{VirtualHome}, JerichoWorld~\cite{AmmanabroluR21a}, FAVOR~\cite{FAVOR}, TextWorld~\cite{TextWorld}, SmartPlay~\cite{wu2024smartplay}. In addition to environments focused on UI actions and operating systems, the procedural text is also used in the realm of code generation~\citet{PayanMSNPBRCDN23}. We discuss the grounded tasks and surveyed environments in more details in \S\ref{ssec:grounded_tasks}.

    \clrev{Symbolic formalisms reflect a \textbf{recurring tension}: the more you can express, the harder the system becomes to learn and use. General-purpose planning languages like PDDL maximize expressiveness~\cite{ManualToPDDL}. They support first-order logic, conditional effects, and complex precondition–effect specifications, typically represented as $\langle \text{Name}, \text{Parameters}, \text{Preconditions}, \text{Effects} \rangle$ tuples. \textbf{Temporal logics} like LTL take a different approach, emphasizing temporal structure through operators such as next and until—constructs that map naturally onto how we describe procedures in language~\cite{bombieri2023mapping}.}

    \clrev{On the other hand, \textbf{domain-specific} languages sacrifice this expressiveness for practical gains. By restricting the action vocabulary to small primitive sets, i.e., 12 household actions, or even just 4 UI commands, these formalisms become easier to learn and more robust~\cite{VirtualHome,branavan-etal-2009-reinforcement,LiHZZB20}. Yet even with a limited scope, they maintain compositional structure, separating actions from arguments (\texttt{pour(beaker1, beaker2)}, \texttt{click(OK\_button)}). This compositionality enables systematic grounding between linguistic expressions and environment entities~\cite{ManualToPDDL,bombieri2023mapping,VirtualHome,LiHZZB20}.}
    
    \clrev{\textbf{Design Constraints.} Moving from description to execution introduces design constraints. To translate natural language like utterance ``make coffee'' into executable steps, first the system needs to infer that one must walk to the coffee maker before pressing its button---\textbf{commonsense prerequisites} that natural language leaves implicit~\cite{VirtualHome}. Second, \textbf{completeness} proves even trickier. When researchers analyzed human-authored programs, they found only 64\% were actually complete~\cite{VirtualHome}. Another 28\% skipped minor steps (sitting down before standing up), while 8\% omitted crucial actions entirely. Two approaches have emerged for handling this: \textbf{Prescriptive} approaches simply restrict what people can say, allowing only predetermined verb forms and connectors~\cite{bombieri2023mapping}. \textbf{Accommodative approaches} take the opposite tack: accept linguistic variation but normalize it aggressively through lemmatization and co-reference resolution~\cite{ManualToPDDL}. Neither approach is perfect. Symbolic structures demand token-level precision, yet procedural knowledge is inherently flexible, e.g., \textit{walk to the kitchen, go to the kitchen, and head to the kitchen} all mean the same thing~\cite{VirtualHome,ManualToPDDL}.}

    \clrev{\textbf{Data Creation.} Creating symbolic training data introduces practical challenges.
    First, \textbf{semantic equivalence}: multiple symbolic representations can express identical behavior. While parsers with lemmatization can normalize synonyms (mapping "walk" and "go" to the same action~\cite{ManualToPDDL}), structurally different programs can achieve identical goals~\cite{VirtualHome,PayanMSNPBRCDN23}. VirtualHome found that the activity ``Make coffee'' had 69 functionally correct programs with only 26\% structural overlap~\cite{VirtualHome}. This means annotations must validate functional correctness rather than exact symbolic matches~\cite{ManualToPDDL,PayanMSNPBRCDN23}.
    Validating functional correctness requires reference models or simulator execution, which is resource-intensive~\cite{PayanMSNPBRCDN23}. Second, \textbf{linguistic naturalness}: existing datasets have largely circumvented the functional correctness validation problem through synthetic generation, producing 295K templated commands in one study and 5,193 procedural programs in another~\cite{LiHZZB20,VirtualHome}. While this reduces annotation costs, it produces linguistically constrained data: VirtualHome’s synthetic programs exhibit vocabulary bias toward code-like expressions~\cite{VirtualHome}, and broader studies of synthetic text report decreased lexical and syntactic diversity compared to human-written data~\cite{guo-etal-2024-curious,chen2024diversitys}}

    \clrev{\textbf{Cross-Domain Transfer.} Symbolic formalisms require domain-specific infrastructure that hinders transfer. PDDL domains specify an ontology (predicates and objects) and action definitions with parameters, preconditions, and effects~\cite{ManualToPDDL}. For instance, a PDDL action definition is a four-tuple \texttt{(Name, Parameters, Preconditions, Effects)} where preconditions are predicate values that must be met before execution~\cite{ManualToPDDL}. LTL templates encode temporal-causal flows using logical operators, mapping natural language connectors (if/otherwise, then/once, until/repeat) to formal logic~\cite{bombieri2023mapping}. Each formalism demands distinct annotation infrastructure. PDDL requires defining domain predicates and grounding them to commonsense ontologies~\cite{ManualToPDDL}. LTL needs language constraint definitions specifying how temporal sequences and conditions are expressed in domain texts~\cite{bombieri2023mapping}. Custom action primitives for simulators require separate predefined action lists~\cite{VirtualHome,wu2024smartplay}. Consequently, annotation efforts developed for one formalism rarely transfer to another~\cite{ManualToPDDL,bombieri2023mapping,LiHZZB20}. This fragmentation substantially increases the cost of scaling symbolic approaches across domains.}
        
    \clrev{These challenges point toward two research directions we explore later. The expressiveness-learnability trade-off and difficulty ensuring functional correctness motivate improved data quality approaches (\S\ref{sec:future_data}). The fragmentation of annotation infrastructure across formalisms motivates better neural-symbolic integration (\S\ref{sec:future_neurosym}).}

\clrev{\subsection{From Representation to Task}}
    \label{sec:mapping}
    
    \clrev{The choice of data representation fundamentally shapes which downstream tasks can be addressed. In short, \textbf{symbolic representations} concentrate on grounded tasks because execution requires translating language into interpretable action primitives—a capability only symbolic formalisms can provide. \textbf{Entity-centric representations} specialize in entity state tracking, as they are mostly designed around predefined entity types, properties, and valid state values that enable precise evaluation while constraining the cross-domain transfer.} 

    \clrev{\textbf{Event-centric representations.} dominate process structure extraction and induction tasks since they explicitly encode the relational structure (sequential, concurrent, conditional) that extraction and induction tasks aim to recover. \textbf{Unstructured corpora} serve as a generalist foundation, enabling massive scale without annotation burden but providing no explicit supervision for structured predictions. Fig.~\ref{fig:repr-task-bipartite} summarizes these representation–task associations and highlights underexplored pairings.}

\begin{figure*}[t]
\centering
\scalebox{0.75}{
\begin{tikzpicture}[
    repr/.style={rectangle, draw=black, fill=blue!15, minimum width=2.8cm, minimum height=0.8cm, font=\small\bfseries},
    task/.style={rectangle, draw=black, fill=green!15, minimum width=2.4cm, minimum height=0.7cm, font=\small},
    taskgrp/.style={rectangle, draw=gray, dashed, rounded corners, inner sep=8pt},
    strong/.style={->, >=Stealth, thick, black},
    moderate/.style={->, >=Stealth, gray},
    gap/.style={->, >=Stealth, dashed, red!70, thick},
    lbl/.style={font=\scriptsize, text=gray}
]

\def\xRep{0}      
\def\xUng{7}      
\def\xGro{-7}     

\node[repr] (unstr)  at (\xRep, 6) {Unstructured};
\node[repr] (event)  at (\xRep, 4) {Event-centric};
\node[repr] (entity) at (\xRep, 2) {Entity-centric};
\node[repr] (symbol) at (\xRep, 0) {Symbolic};

\node[task] (web)   at (\xGro, 6.5) {Web};
\node[task] (nav)   at (\xGro, 5.5) {Navigation};
\node[task] (robot) at (\xGro, 4.5) {Robotics};
\node[task] (game)  at (\xGro, 3.5) {Game};
\node[task] (gui)   at (\xGro, 2.5) {GUI};
\node[task] (dial)  at (\xGro, 1.5) {Dialogue};

\node[task] (summ)  at (\xUng, 8.5) {Summarization};
\node[task] (qa)    at (\xUng, 7.5) {Question Answering};
\node[task] (gen)   at (\xUng, 6.5) {Script Generation};
\node[task] (align) at (\xUng, 5.5) {Event Alignment};
\node[task] (extr)  at (\xUng, 4.5) {Process Extraction};
\node[task] (ind)   at (\xUng, 3.5) {Process Induction};
\node[task] (impl)  at (\xUng, 2.5) {Implicit Detection};
\node[task] (track) at (\xUng, 1.5) {Entity Tracking};
\node[task] (know)  at (\xUng, 0.5) {Knowledge Acquisition};

\node[taskgrp, fit=(summ)(qa)(gen)(align)(extr)(ind)(impl)(track)(know),
      label={[font=\small\bfseries]above:Ungrounded Tasks}] {};
\node[taskgrp, fit=(web)(nav)(robot)(game)(gui)(dial),
      label={[font=\small\bfseries]above:Grounded Tasks}] {};

\draw[strong]   (unstr.east) -- (summ.west);
\draw[strong]   (unstr.east) -- (qa.west);
\draw[strong]   (unstr.east) -- (gen.west);
\draw[moderate] (unstr.east) -- (align.west);
\draw[moderate] (unstr.east) -- (impl.west);

\draw[strong]   (event.east) -- (extr.west);
\draw[strong]   (event.east) -- (ind.west);
\draw[moderate] (event.east) -- (align.west);
\draw[moderate] (event.east) -- (know.west);

\draw[strong]   (entity.east) -- (track.west);
\draw[moderate] (entity.east) -- (impl.west);
\draw[moderate] (entity.east) -- (qa.west);

\draw[strong] (symbol.west) -- (web.east);
\draw[strong] (symbol.west) -- (nav.east);
\draw[strong] (symbol.west) -- (robot.east);
\draw[strong] (symbol.west) -- (game.east);
\draw[strong] (symbol.west) -- (gui.east);
\draw[strong] (symbol.west) -- (dial.east);

\draw[gap] (entity.east) -- node[lbl, above, sloped] {gap} (gen.west);

\draw[gap] (symbol.east) to[out=30,in=210]
    node[lbl, above, sloped, pos=0.3] {gap} (align.west);

\node[anchor=north west] at (-1, -1.2) {
    \begin{tikzpicture}[font=\scriptsize]
        \draw[strong] (0,0) -- (0.8,0) node[right] {Strong association};
        \draw[moderate] (0,-0.4) -- (0.8,-0.4) node[right] {Moderate association};
        \draw[gap] (0,-0.8) -- (0.8,-0.8) node[right] {Underexplored (gap)};
    \end{tikzpicture}
};

\end{tikzpicture}
}
\caption{\clrev{Representation-task associations in procedural language understanding. Strong edges indicate dominant pairings in the literature; moderate edges show secondary associations. Dashed red edges highlight underexplored combinations representing research opportunities: entity-centric representations for script generation tasks, and symbolic methods for alignment.}}
\label{fig:repr-task-bipartite}
\end{figure*}

    \clrev{Two notable gaps emerge from this mapping. First, event-centric and entity-centric representations capture complementary aspects—action sequences versus state changes—yet remain largely separate. Real-world procedural understanding requires both: an agent following cooking instructions must know the action sequence \textit{and} track ingredient states. Second, certain pairings remain unexplored, such as entity-centric approaches for generation or symbolic methods for alignment.}
    
    \clrev{Beyond these theoretical gaps, practical constraints also shape the landscape. Representation-task associations manifest differently across application domains, as domain characteristics often dictate which representations are practical and which tasks are feasible. We provide a more detailed analysis below:}

    \clrev{\textbf{Recipes and cooking instructions domains} dominate the literature due to generally well-formed web text mostly created by experts, abundant web data (e.g., Recipe1M~\cite{marin2019learning}, WikiHow-sourced datasets and corpora, AllRecipes), and accessible semantics requiring no specialized expertise. These domains favor unstructured, event-centric and later, entity-centric data representations and support the broadest task diversity—from summarization and question answering to grounded execution—making cooking the \textit{defacto} testbed for procedural understanding.}

    \clrev{\textbf{Scientific and technical domains.} (e.g., materials synthesis~\cite{MysoreJKHCSFMO19}, technical troubleshooting~\cite{TechTrack}) present a contrasting profile. These domains involve specialized lexicons—domain-specific action vocabularies and entity types such as chemical compounds and hardware components. Because of this 
    specialization, annotation requires domain expertise, leading to high costs that constrain dataset scale. This pushes research toward event-centric representations with controlled vocabularies and extraction tasks.}
    
    \clrev{\textbf{Game and simulation domains.} (e.g., TextWorld~\cite{TextWorld}, ALFRED~\cite{ALFRED20}, Minecraft~\cite{Narayan-ChenJH19}) use symbolic representations by design, as environments require executable action primitives. Event granularity varies from atomic UI actions to high-level goals, but linguistic diversity is typically sacrificed for environment compatibility. The tight coupling between representation and execution means these domains focus almost exclusively on grounded tasks.}

    \clrev{\textbf{Business process domains.} occupy a unique niche, using standardized formalisms like \gls{BPMN} that encode rich event-event relations (successive, concurrent, 
conditional, exclusive, optional, parent-child) rarely captured in other domains~\cite{ParkM18, ProcessModelExtraction}. This representational expressiveness enables sophisticated process extraction and induction but limits cross-domain transfer due to formalism-specific assumptions.}

    \clrev{This domain concentration has methodological implications: the field's heavy reliance on cooking data means that models are optimized for and evaluated on procedures with accessible vocabulary, explicit action-object structure, and entity-centric state changes. Whether these models generalize to technical procedures, where implicit domain knowledge, specialized terminology, and different event granularities are the norm, remains largely untested. We return to these gaps and their implications for evaluation in \S\ref{sec:future}. With these representations, task, and domain relationships established, we now examine each task category in detail.}

\section{Tasks}
\label{sec:tasks}
Plenty of downstream tasks have been proposed to evaluate procedural language understanding in general. These tasks are naturally constrained by the data representation. We broadly categorize the procedural tasks into two categories: (1) grounded and (2) ungrounded. \textit{Grounded} tasks refer to the tasks where the instruction is grounded on an external database, document, or environment, whereas \textit{ungrounded} tasks are solely defined on textual data. Please refer to \cref{fig:typo-instruction} for the taxonomy and to App.~\ref{app:tasks-papers} for the full list of tasks and papers.

\subsection{Ungrounded Tasks}
\label{ssec:ungr_tasks}
In this section, we discuss various ungrounded tasks related to instructional text. \clrev{Table~\ref{tab:ungrounded-comparison} summarizes the landscape of ungrounded tasks, highlighting the datasets, models and evaluation metrics that characterize each area.}

\begin{table}[!htp]
\centering
\small
\scalebox{0.90}{
\begin{tabular}{m{2.5cm}|m{5cm}|m{2.5cm}|m{2.5cm}}
\hline
\textbf{Task} & \textbf{Common Datasets} & \textbf{Common Models} & \textbf{Metrics} \\
\hline
Summarization & WikiHow~\cite{wikihowSummary} \newline HowSumm~\cite{Boni2021HowSummAM}& T5 \newline mBART \newline BART \newline Pegasus & ROUGE \newline BLEU \newline BERTScore \newline METEOR \\
\hline
Event Alignment & Recipe1M~\cite{marin2019learning} \newline ARA Corupus~\cite{DonatelliSBKZK21} \newline DeScript~\cite{DeScript} & BERT \newline LSTM \newline HMM & Accuracy \newline P/R/F1 \\
\hline
Implicit Detection & R2R-ULN~\cite{FengFLW22} \newline KIDSCOOK~\cite{bisk-etal-2019-benchmarking} & Pretrained LMs \newline Vision-LMs & P/R/F1 \newline Accuracy \\
\hline
Entity Tracking & ProPara~\cite{ProPara1} \newline OpenPI~\cite{OpenPI} & Neural Process Networks \newline BERT variants & Accuracy \newline P/R/F1 \newline BLEU \newline ROUGE \\
\hline
Process Structure \newline Extraction & RecipeDB~\cite{10.1093/database/baaa077} \newline Recipe1M~\cite{8099810} \newline MSComplexTasks~\cite{zhang-etal-2021-learning} \newline MIAIS~\cite{MIAIS} & Bi-LSTM \newline CNN \newline BERT \newline GPT-4 & P/R/F1 \newline Success Rate \newline BERTScore \\
\hline
Process Structure \newline Induction & MiniWoB~\cite{LiuGPSL18} \newline DeScript~\cite{DeScript} & MSA \newline Clustering \newline Seq2Seq \newline GPT-4 & E-ROUGE \newline P/R \\
\hline
Procedural\newline Generation & proScript~\cite{proScript} \newline CoScript~\cite{yuan-etal-2023-distilling} \newline InScript~\cite{InScript} \newline WIKIPLAN~\cite{lu2023multimodal} & T5 \newline GPT-3 \newline LSTM/GRU & BLEU \newline ROUGE \newline BERTScore \newline METEOR \\
\hline
Question\newline Answering & WikiHowQA~\cite{Bolotova-Baranova23} \newline MCScript2.0~\cite{Mcscript2} \newline HellaSwag~\cite{zellers-etal-2019-hellaswag} \newline RecipeQA~\cite{RecipeQA} & BERT \newline BART \newline RoBERTa \newline Triplet Networks & Accuracy \newline ROUGE \newline BLEU \newline Recall@k \\
\hline
Knowledge Acq./Mining & Data sourced from WikiHow and Instruction manuals & BERT \newline LongFormer & Accuracy \newline Recall \\
\hline
\end{tabular}
}
\caption{\clrev{Summary of major ungrounded tasks in procedural text, highlighting commonly used datasets, modeling approaches, and evaluation metrics.}}
\label{tab:ungrounded-comparison}
\end{table}

\subsubsection{Summarization}
\label{sssec:summarization}

The trend focuses on extracting key instructions from either \textbf{single}~\cite{wikihowMultilingualSummary, wikihowSummary, le-luu-2024-extractive} or \textbf{multiple}~\cite{Boni2021HowSummAM} procedural documents. This can be \textbf{extractive}, where key sentences are directly compiled from the source text, or \textbf{abstractive}, where new phrases are generated to summarize the content. WikiHow~\cite{wikihowSummary} and HowSumm~\cite{Boni2021HowSummAM}, discussed in~\ref{ssec:unstr_data}, are the main resources. For instance, \citet{wikihowSummary} generate human-crafted summaries for entire WikiHow articles, while \citet{Boni2021HowSummAM} produce extractive summaries for external documents. \citet{le-luu-2024-extractive} generate extractive summaries from WikiHow without explicit labels, selecting key sentences that approximate their abstractive summaries. \citet{DeChant2022SummarizingAV} utilize the ALFRED~\cite{ALFRED20} dataset to generate abstractive summaries of robotic actions, leveraging video frames and detailed action descriptions to produce both brief, one-sentence summaries and detailed, step-by-step instructions for each task. \clrev{
Methods have evolved from seq2seq models~\cite{wikihowSummary} to fine-tuned transformers such as T5, mBART, BART, and Pegasus~\cite{wikihowMultilingualSummary, Boni2021HowSummAM, DeChant2022SummarizingAV, le-luu-2024-extractive, kwon2024ordersum}, with ROUGE as the dominant evaluation metric, supplemented by BLEU, BERTScore, and METEOR~\cite{DeChant2022SummarizingAV, le-etal-2024-abs2ext, wikihowSummary}.}

\clrev{
\textbf{Limitations and Open Questions.}
The core limitation is the \textbf{optimization-task mismatch}: models optimize sentence-level salience but are evaluated on summary-level coherence~\cite{kwon2024ordersum}. This explains why systems underperform despite strong sentence selection: independently chosen sentences fail to form coherent procedural sequences. We believe three deficiencies might drive this: (1) models arrange sentences by probability rather than semantic dependencies, causing step-ordering failures; (2) procedural details like measurements are lost when optimizing for salience~\cite{DeChant2022SummarizingAV}; (3) metrics like ROUGE-L cannot penalize ordering violations~\cite{kwon2024ordersum}, masking these failures. These evaluation limitations connect to broader metric inadequacy issues discussed in~\S\ref{ssec:cross_cutting}, with potential solutions in~\S\ref{sec:future_eval}.}

\subsubsection{Event Alignment}
\label{sssec:event_alignment}

\begin{figure}[h]
    \centering
    \includegraphics[width=0.6\textwidth]{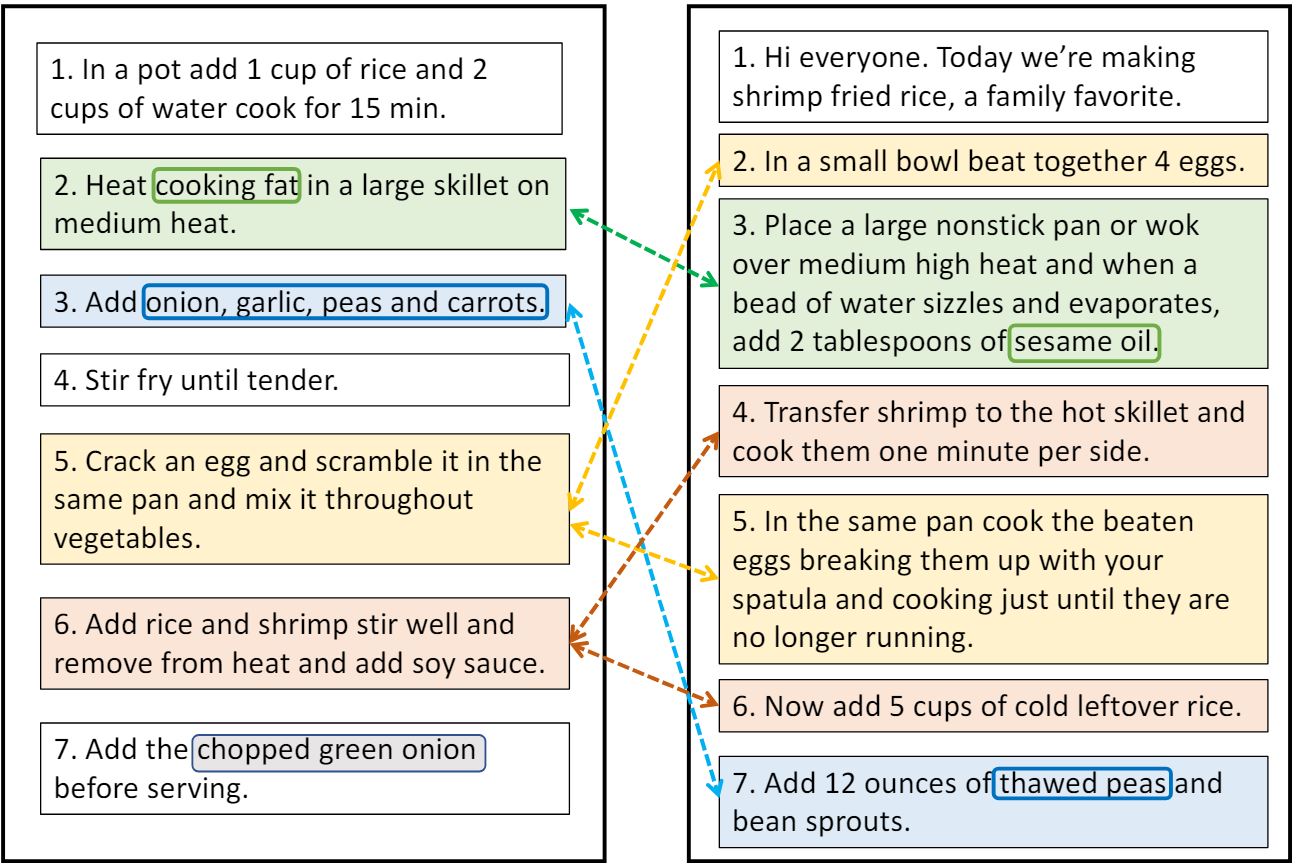}
    \caption{Recipe Alignment Example from \citet{lin-etal-2020-recipe}}
    \label{fig:ms-recipes}
\end{figure}

Event alignment typically refers to the process of identifying and matching corresponding actions or steps across different sets of unstructured instructions. An example on recipes~\cite{lin-etal-2020-recipe} is given in Fig.~\ref{fig:ms-recipes}. Event alignment is either one-to-one~\cite{nandy-etal-2023-clmsm, regneri-etal-2010-learning, wanzare-etal-2017-inducing}, a.k.a. \textit{paraphrasing}, or one-to-many~\cite{lin-etal-2020-recipe, DonatelliSBKZK21}.

The task is rich in domains thanks to the availability of raw procedural text. For example, several works~\cite{nandy-etal-2023-clmsm, lin-etal-2020-recipe, DonatelliSBKZK21} focus on the alignment of instructions within different cooking \textbf{recipes}---\citet{nandy-etal-2023-clmsm} use a dataset containing 1 million recipes from Recipe1M+~\cite{marin2019learning}, RecipeNLG~\cite{bien-etal-2020-recipenlg}, and various how-to blog websites, while \citet{lin-etal-2020-recipe} introduce a dataset comprising 150K instruction pairwise alignments across 4,262 dishes. More recently, \citet{DonatelliSBKZK21} provide the \gls{ARA} corpus comprising recipes for 10 different dishes, originally sourced from the \citet{lin-etal-2020-recipe}. Apart from recipes, a considerable amount of work~\cite{regneri-etal-2010-learning,DeScript,wanzare-etal-2017-inducing} focuses on \textbf{everyday tasks} and compile novel resources for \gls{ESD} such as DeSCRIPT, SMILE and OMICS\footnote{\url{http://openmind.hri-us.com/}}.

\clrev{
Most datasets lack hand-annotated alignments, driving reliance on unsupervised methods: HMMs~\cite{lin-etal-2020-recipe}, semi-supervised clustering~\cite{wanzare-etal-2017-inducing}, and neural encoders (BERT, LSTM)~\cite{nandy-etal-2023-clmsm, DonatelliSBKZK21}. Evaluation uses accuracy, precision, recall, and F1 against manually annotated subsets or held-out gold data~\cite{regneri-etal-2010-learning, DonatelliSBKZK21, lin-etal-2020-recipe}.}

\clrev{
\textbf{Limitations and Open Questions.}
Approaches to recipe action alignment achieve 66.97--72.4\% accuracy~\cite{DonatelliSBKZK21,nandy-etal-2023-clmsm}, maintaining a 6.6--12\% gap from human performance (79\%~\cite{DonatelliSBKZK21}). This gap concentrates in cases requiring commonsense inference~\cite{DonatelliSBKZK21}: implicit actions (24\% of disagreements) and light verb constructions (22\% of disagreements). Analysis of these failures suggests two underlying problems: (1) local context bias may prevent capturing step-wise contexts across entire recipes~\cite{nandy-etal-2023-clmsm}; (2) light verb constructions like ``let rest'' create inconsistent action boundaries, with 22\% of human disagreements stemming from whether causative verbs constitute separate actions~\cite{DonatelliSBKZK21}. These failures exemplify the implicit knowledge problem discussed in~\S\ref{ssec:cross_cutting} and~\S\ref{sec:future_implicit}, creating an immediate opening to explore the potential of recent large language models for this task.}

\subsubsection{Detecting and Correcting Implicit Instructions}
\label{sssec:implicit}

An intriguing but understudied challenge in procedural text is the presence of implicit or unclear statements. \citet{AnthonioBR20} address this by constructing a corpus based on revisions to WikiHow articles, focusing on whether sentences need clarification (e.g., adding missing pronouns or quantifiers). They also introduce a shared task~\cite{roth-anthonio-2021-unimplicit} to promote research on implicit language, and initiated a workshop series\footnote{\url{https://roth-anthonio-2021-unimplicit.github.io/}\\\url{https://roth-anthonio-2021-unimplicit2022.github.io/}}. \citet{Zeng2020Missing} tackle the challenge of \textit{repairing procedural text} by using external activity templates to fill in missing information. \citet{RecipeNER} develop a dataset of Chinese recipes from Haodou recipe website, where missing entities are supplemented using both text and images. \citet{bisk-etal-2019-benchmarking} introduce the KIDSCOOK dataset, pairing cooking instructions with hierarchical sub-steps, and proposed a task to infer hierarchical relations between instructions and sub-steps. Similarly, \citet{FengFLW22} create the R2R-ULN dataset, which involves guiding an agent through environments with underspecified navigation instructions. The works in this task fall into two main categories: (1) \textbf{detecting and correcting implicit or missing information}, and (2) \textbf{clarifying procedural instructions}. The first category focuses on identifying and completing missing or implicit details in procedural texts, using inputs that often include underspecified texts, sometimes with images or process models. The output is a refined text, a simulated representation, or a predicted completion, where missing actions, entities, or arguments are corrected, inferred, or predicted~\cite{clark2018leveraging,cheng-erk-2018-implicit,Zeng2020Missing,RecipeNER}. The second category aims to improve clarity and structure in instructional texts by refining vague or ambiguous instructions. The output is a clearer, explicit, and more actionable set of instructions or a prediction of where clarification is needed to achieve this~\cite{AnthonioBR20,roth-anthonio-2021-unimplicit,bisk-etal-2019-benchmarking,FengFLW22,zhang-etal-2023-clevr}.

\clrev{
Works on implicit information use pretrained language models and vision-language 
models~\cite{roth-anthonio-2021-unimplicit, FengFLW22}. These methods 
are typically evaluated using precision, recall, F1, and 
accuracy~\cite{cheng-erk-2018-implicit,RecipeNER,Zeng2020Missing}.}

\clrev{
\textbf{Limitations and Open Questions.}
Performance splits sharply between relatively \emph{explicit} clarification phenomena and more \emph{implicit} ones: UnImplicit systems reach 92.7\% on pronoun resolutions but hover near chance for modifier/quantifier insertions (53.6\%--54.2\% vs.\ 50\% random)~\cite{roth-anthonio-2021-unimplicit}. Similar patterns recur across procedural benchmarks where crucial content is \emph{expected but not stated}: models struggle when success requires recovering omitted or underspecified steps, arguments, or action effects rather than matching surface cues~\cite{jiang-etal-2020-recipe,ZengTNDLX20,roth-anthonio-2021-unimplicit}. We identify two factors to be particularly important: \textbf{(1) Reference under state change:} entities are repeatedly transformed or combined, so resolving ``what does this refer to now?'' becomes a state-tracking rather than standard coreference problem~\cite{jiang-etal-2020-recipe}. \textbf{(2) Sparse, indirect supervision for absence:} supervision is dominated by observed mentions and edits, while events and arguments that \emph{should} be present are only indirectly observable via revision filters or model--text misalignments, leading to skewed learning~\cite{roth-anthonio-2021-unimplicit,AnthonioBR20,ZengTNDLX20,jiang-etal-2020-recipe}.}

\clrev{These challenges show up both in clarification-from-revision settings and in omission-recovery frameworks where only a subset of activities or entities is missing from the surface text. Moreover, revision-derived supervision risks conflating \emph{data convenience} with \emph{instructional adequacy}: edits may reflect style or correctness rather than necessity, and UnImplicit's ``unchanged $\Rightarrow$ no clarification'' assumption is explicitly a heuristic~\cite{AnthonioBR20,roth-anthonio-2021-unimplicit}. A central open question is therefore how to define and evaluate ``implicitness'' relative to user expertise, task goals, and external context, not just observed edits or local textual cues~\cite{roth-anthonio-2021-unimplicit,AnthonioBR20,jiang-etal-2020-recipe}. This task most directly exposes the implicit knowledge problem central to procedural understanding, discussed in~\S\ref{ssec:cross_cutting} and~\S\ref{sec:future_implicit}.} 

\subsubsection{Entity State Tracking}
\label{sssec:entity_tracking}

In \S\ref{ssec:ent_repr}, we present a detailed analysis of how entities, procedures, and their relationships are represented in the literature. Entity (state) tracking task is defined on these representations and focuses on tracking entities (e.g., ingredients, tools) and their properties (e.g., color, location) and interactions through each step. Several datasets have been created, each focusing on different domains, e.g., hardware and software setup~\cite{TechTrack}, WikiHow~\cite{OpenPI, openPIv2} and cooking recipes~\cite{rim-etal-2023-coreference}. Entity tracking is mostly the primary task~\cite{clark2018leveraging, ProPara1, ProPara2, ZhangGQWJ21, SconeLongPL16, li-etal-2021-implicit, KimS23, BosselutLHEFC18, openPIv2}, or serves as an intermediate step for other event related tasks~\cite{CREPE, rim-etal-2023-coreference, cheng-erk-2018-implicit, kazeminejad-palmer-2023-event, nandy-etal-2023-clmsm, nandy-etal-2024-order}. For the first category, \citet{ProPara1} and \citet{clark2018leveraging} leverage scientific procedural text, i.e., Process Paragraph~(ProPara), while \citet{ZhangGQWJ21} and \citet{BosselutLHEFC18} use Recipes to model entity movements and state changes across multiple steps. These datasets use predefined set of entities, properties and actions, offering a more controlled setup. More recently, open-domain tracking datasets~\cite{ProPara2,OpenPI} are introduced where entities and properties need to be \textbf{extracted or inferred}, which provides a more challenging setup. On the other hand, several others~\cite{SconeLongPL16,KimS23} build simple synthetic datasets for a predefined set of actions and states. For the second category, entity tracking plays a supporting role for other tasks like causal reasoning, coreference resolution, and multi-task learning. Tracking how entities change state helps identify relationships between actions and their effects~\cite{ProPara2} or predict hypothetical event outcomes~\cite{CREPE} for causal reasoning. Tracking also helps resolving coreferences~\cite{rim-etal-2023-coreference} and identifying implicit arguments~\cite{cheng-erk-2018-implicit}. Furthermore, tracking supports action alignment, step prediction, and understanding event-driven state changes~\cite{kazeminejad-palmer-2023-event, nandy-etal-2023-clmsm, nandy-etal-2024-order}.

\clrev{
Approaches range from tailored architectures like Neural Process Networks~\cite{BosselutLHEFC18} to pretrained models~\cite{ZhangGQWJ21, nandy-etal-2023-clmsm}, with more recent work integrating symbolic reasoning via ConceptNet and VerbNet~\cite{kazeminejad-palmer-2023-event}. Evaluation uses accuracy, F1, and---for open-vocabulary settings---generation metrics like BLEU and BERTScore~\cite{OpenPI, openPIv2}.}

\clrev{
\textbf{Limitations and Open Questions.}
Controlled experiments expose severe limitations in current models’ entity tracking abilities. Probing classifiers achieve 86.8\% accuracy on trivial cases but collapse to \textbf{3.1\% on non-trivial tracking}~\cite{KimS23}. Even GPT-3.5 degrades from roughly 70--75\% to about 25\% accuracy as the number of state-changing operations \emph{affecting a single entity} increases to seven, despite the entire description fitting within the context window~\cite{KimS23}. These failures are consistent with vanilla Transformers lacking persistent state mechanisms for updating entity representations over sequences of operations~\cite{KimS23,fagnou-etal-2024-chain}. Notably, only code-pretrained models succeed at tracking: \citet{KimS23} report that ``only models in the GPT-3.5 series, which have been trained on both text and code, are able to perform non-trivial entity tracking,'' suggesting that text-only pretraining is insufficient for this capability to surface. Complementary work on open-vocabulary procedural state tracking finds that even strengthened baselines with explicit temporal dependency and entity awareness leave state-of-the-art models far from competent, especially when tracking many attributes over multiple steps~\cite{OpenPIC,li2023DynamicWorld,openPIv2}.}

\clrev{
These findings suggest a research direction: \emph{Do transformers require explicit persistent-state mechanisms to track entity attributes compositionally?} The degradation with increasing state-changing operations—despite the full description fitting within the context window—indicates that access to tokens alone is insufficient for reliable state updating~\cite{KimS23}. This aligns with theoretical and empirical evidence that vanilla attention can be structurally inefficient for repeated entity updates, motivating modified attention/state-propagation mechanisms~\cite{fagnou-etal-2024-chain}. More broadly, the continued difficulty of open-vocabulary procedural tracking—especially with many attributes over multiple steps—suggests evaluating memory-augmented and structured-state variants under the same stressors used by OpenPI-style benchmarks~\cite{OpenPI,openPIv2,li2023DynamicWorld}.}

\subsubsection{Instruction Parsing}
\label{sssec:parsing}

Parsing natural language instructions into structured data representations such as trees, graphs, and \gls{BPMN} is one of the most crucial steps to be able to run/execute them. There are two main approaches: 1) Supervised parsing, where the model uses labeled data to train on \textbf{\gls{PSE}}, and 2) Semi-supervised or unsupervised parsing, referred to as \textbf{\gls{PSI}}, explained in detail below.

\paragraph{\textbf{Process Structure Extraction}}
\label{sssec:extraction}

\begin{figure}[h]
    \centering
    \includegraphics[width=0.85\textwidth]{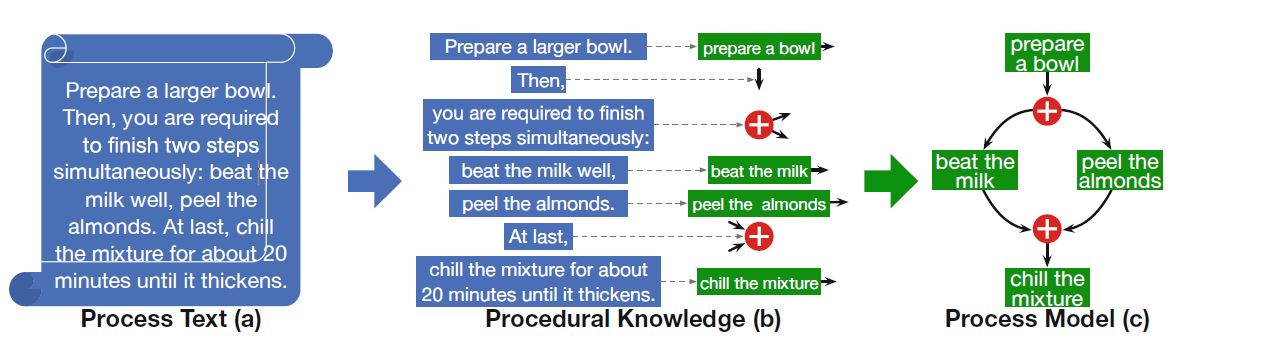}
    \caption{Illustration of The PME Problem~\cite{ProcessModelExtraction}}
    \label{fig:pme}
\end{figure}

In \gls{PSE}, unstructured procedural text is parsed into structured formats like trees, graphs, or \gls{BPMN}s. \gls{PSE} plays a key role in domains like \textbf{business process management} by extracting actions, actors, and their relationships within task sequences. Fig.~\ref{fig:pme} shows an illustration of the \gls{PSE} problem from \citet{ProcessModelExtraction}. This task is rich with diverse domains like \textbf{cooking recipes}~\cite{diwan2020named, pan2020multi, wang2022learning, ProcessModelExtraction, kiddon-etal-2015-mise, maeta-etal-2015-framework, MIAIS, bhatt-etal-2024-end, ManualToPDDL, FengZK18, Brach2025Effectiveness}, \textbf{household and everyday tasks}~\cite{zhang-etal-2021-learning, FengZK18, textMining23, kourani2024process, zhou-etal-2022-show, zhang-etal-2024-proc2pddl,ManualToPDDL}, \textbf{technical and scientific procedures}~\cite{MysoreJKHCSFMO19,textMining23, ManualToPDDL}, and \textbf{general instructions}~\cite{ito2020natural}. \citet{diwan2020named}, \citet{FengZK18} and \citet{wang2022learning} extract the structure (ingredients, cooking techniques, and utensils) of recipes from the RecipeDB dataset~\cite{10.1093/database/baaa077}, \url{cookingtutorials.com}, and the Recipe1M dataset~\cite{8099810}, respectively. On the other hand, several other works generate new datasets. \citet{MysoreJKHCSFMO19} generate 230 synthesis procedures (discrete process steps taken to synthesize the target material) with labeled graphs that express the semantics of the synthesis sentences (actions, materials involved, conditions, etc.). \citet{zhang-etal-2021-learning} collect the MSComplexTasks dataset that contains complex tasks from Wunderlist\footnote{\url{https://en.wikipedia.org/wiki/Wunderlist}} with their sub-task graph, while \citet{MIAIS} create the MIAIS dataset (see \S\ref{sec:data}). Graphs and trees are the most commonly used due to their relations with workflow representations~\cite{zhang-etal-2021-learning, pan2020multi, MysoreJKHCSFMO19, ProcessModelExtraction, maeta-etal-2015-framework, MIAIS, textMining23, zhou-etal-2022-show, wang2022learning}. Other works produce well-structured key-value pairs~\cite{diwan2020named}, \gls{PDDL}~\cite{ManualToPDDL, zhang-etal-2024-proc2pddl}, or lists of structured sequential actions~\cite{FengZK18}.

\clrev{
Approaches evolved from basic NLP techniques (NER, POS tagging)~\cite{diwan2020named} through neural architectures (Bi-LSTM, CNN)~\cite{ProcessModelExtraction, ParkM18} to transformer models and LLMs for PDDL generation~\cite{bhatt-etal-2024-end, zhang-etal-2024-proc2pddl, kourani2024process}. Evaluation uses precision, recall, F1, and, for executable outputs, PDDL solve rates~\cite{zhang-etal-2024-proc2pddl}.}

\clrev{
\textbf{Limitations and Open Questions.}
High local action-extraction scores do not guarantee reliable executable process models. Contextual extractors achieve 96.22\% F1 on technical manuals but drop to 82.59\% on home-and-garden instructions, indicating substantial performance variation across procedural domains~\cite{ManualToPDDL}. Similarly, component-level correctness fails to ensure planning success: GPT-4o reaches only 21.4\% intrinsic action accuracy and 45.3\% problem solve rate, with preconditions (31.1--40.1\%) notably harder to predict than effects (53.5--62.5\%), consistent with the analysis that preconditions require deeper implicit knowledge about entity states and are often less explicitly stated in text~\cite{zhang-etal-2024-proc2pddl}. LLM-assisted BPMN generation frameworks can produce sound/executable models yet still show semantic deviations from intended process specifications, motivating iterative refinement via feedback~\cite{ProcessModelExtraction}. These extraction-to-execution gaps highlight that current systems capture surface procedural patterns while struggling with compositional generalization and implicit state reasoning, motivating the need for execution-grounded evaluation metrics discussed in~\S\ref{sec:future_eval}.}

\paragraph{\textbf{Process Structure Induction}}
\label{sssec:induction}

Induction uses unsupervised or semi-supervised techniques to \textit{induce the latent} structure of event sequences. Here, the goal is to automatically identify and align similar event descriptions across multiple instances of a scenario/procedure, clustering them into paraphrase sets that represent the event types (nodes) within the procedure. Most common scripts are from everyday tasks, such as baking a cake, grocery shopping, or booking a flight, described as \gls{ESD}s~\cite{regneri-etal-2010-learning, DeScript, wanzare-etal-2017-inducing}, or with text narratives describing the process~\cite{textMining23, WuZHSP23, LiuGPSL18}. Additionally, some approaches incorporate structured data, such as graphs, to generalize subprocesses into more generic ones~\cite{zhang-etal-2020-analogous}. In addition to collecting processes from resources like WikiHow~\cite{zhang-etal-2020-analogous, WuZHSP23} and public manuals~\cite{textMining23, WuZHSP23}, other works~\cite{wanzare-etal-2017-inducing, regneri-etal-2010-learning} use existing datasets, such as the OMICS corpus and the MiniWoB dataset~\cite{ShiKFHL17}. Similar to the process extraction task, graph-based structures dominate the field~\cite{regneri-etal-2010-learning, wanzare-etal-2017-inducing, WuZHSP23, zhang-etal-2020-analogous, DeScript}. Several other works generate data in other formats, such as sequences of low-level actions (e.g., atomic or specific user actions, such as clicking buttons)~\cite{LiuGPSL18} or ontologies~\cite{textMining23}.

\clrev{
Traditional methods use \gls{MSA} and clustering for temporal script graphs~\cite{DeScript, wanzare-etal-2017-inducing, regneri-etal-2010-learning}, while recent work applies seq2seq models~\cite{zhang-etal-2020-analogous} and GPT-4 for zero-shot structuring~\cite{textMining23}. Evaluation relies on E-ROUGE, precision/recall, and crowdsourced annotation of subsets~\cite{zhang-etal-2020-analogous, wanzare-etal-2017-inducing}.}

\clrev{
\textbf{Limitations and Open Questions.}
Models achieve only 14.8 E-ROUGE1 versus 29.0 human performance at verb-level prediction, degrading to 3.5 vs. 11.6 when predicting complete event structures (all words)~\cite{zhang-etal-2020-analogous}. This degradation reveals a core bottleneck: models identify individual steps but cannot assemble them into coherent structures. We believe two task-specific deficiencies might drive this: (1) \textbf{weak abstraction mechanisms}: performance is highly sensitive to hyperparameter choices for abstraction depth~\cite{zhang-etal-2020-analogous}; and (2) \textbf{absence of explicit dependency modeling}: current approaches lack representations of preconditions~\cite{WuZHSP23} and step interchangeability~\cite{wu-etal-2022-understanding}. Current evaluation worsens the problem: sparse reference orderings (only 1.7 ground truths per instance~\cite{zhang-etal-2020-analogous}) fail to distinguish valid variations from genuine failures~\cite{DeScript}. The compositional reasoning limitations connect to architectural gaps discussed in~\S\ref{ssec:cross_cutting} and~\S\ref{sec:future_hierarchy}.}

\subsubsection{Procedural Generation}
\label{sssec:generation}

Procedural generation, is also referred to as script construction~\cite{lyu-etal-2021-goal}, similar to other tasks, cooking recipes~\cite{kiddon-etal-2016-globally, liu-etal-2022-counterfactual, lu2023multimodal, 9288722}, and everyday tasks~\cite{InScript, proScript, lyu-etal-2021-goal, 9288722, sun-etal-2023-incorporating, NguyenNCTP17, yuan-etal-2023-distilling} like ``baking a cake'' or ``fueling a car'' from WikiHow and other sources, are the most common domains. While most datasets are derived from existing resources~\cite{kiddon-etal-2016-globally,NguyenNCTP17, rojowiec-etal-2020-generating,lyu-etal-2021-goal,9288722, sun-etal-2023-incorporating}, several are created specifically for this task, such as proScript~\cite{proScript}, CoScript~\cite{yuan-etal-2023-distilling}, WIKIPLAN and RECIPEPLAN~\cite{lu2023multimodal}, InScript~\cite{InScript} and XIACHUFANG~\cite{liu-etal-2022-counterfactual} datasets. The output of the models vary, ranging from sequences of natural language steps~\cite{rojowiec-etal-2020-generating, kiddon-etal-2016-globally, lyu-etal-2021-goal, sun-etal-2023-incorporating, liu-etal-2022-counterfactual, yuan-etal-2023-distilling, proScript}, to multimodal procedural plans pairing text and images~\cite{lu2023multimodal, 9288722} and to graph in DOT language---graph description language detailing event sequences~\cite{proScript}. Input can also be diverse; such as descriptions of scenarios with ordered~\cite{NguyenNCTP17} or partially ordered~\cite{proScript} events, or just the procedure goal~\cite{lyu-etal-2021-goal, lu2023multimodal}, with additional details like a list of ingredients~\cite{kiddon-etal-2016-globally}, constraints~\cite{yuan-etal-2023-distilling}, or user preferences~\cite{sun-etal-2023-incorporating, liu-etal-2022-counterfactual}. Other input formats can include image sequences or pairs of ``before'' and ``after'' images~\cite{rojowiec-etal-2020-generating}.

\clrev{
Approaches evolved from seq2seq models with LSTMs~\cite{NguyenNCTP17, 9288722, rojowiec-etal-2020-generating} to pretrained LMs like T5 and GPT-3~\cite{proScript, yuan-etal-2023-distilling, lyu-etal-2021-goal}. Evaluation uses BLEU, ROUGE, BERTScore, and METEOR, though these metrics poorly capture procedural coherence~\cite{lyu-etal-2021-goal, yuan-etal-2023-distilling}.}

\clrev{\textbf{Limitations and Open Questions.}
Despite steady gains in fluency, current procedural generators still struggle for two recurring reasons. The first is \textbf{stateful consistency over long horizons}. Even when outputs are locally plausible, models often fail to maintain and update latent state as a procedure unfolds. This is visible in structured script generation on proScript, where graph edit distance remains far from human quality (model--human: 4.97 vs.\ 2.98; lower is better)~\cite{proScript}. Similar issues appear at longer lengths. On LONGPROC, average performance degrades sharply with output length (GPT-4o: 83.4 at 2K vs.\ 38.1 at 8K)~\cite{ye2025longproc}. A targeted ToM Tracking analysis at 8K reports that 19/20 inspected GPT-4o errors arise from incorrect long-range state inferences~\cite{ye2025longproc}. SCEDIT similarly finds that edits often pass direct post-edit checks but fail to transfer to downstream script generation, with an average 27\% drop when moving from post-edit success to script-level success~\cite{ScEdit2025}. The second recurring issue is \textbf{adaptation under interventions}. When procedures must change under substitutions or counterfactual conditions, models remain brittle. Counterfactual recipe generation highlights this brittleness under ingredient replacements, where models often fail to consistently incorporate replaced or added ingredients and to insert or delete ingredient-dependent actions~\cite{liu-etal-2022-counterfactual}. Together, these observations motivate approaches that make constraints and state explicit rather than leaving them implicit in next-token prediction. For example, constrained-generation pipelines such as over-generate-then-filter can substantially improve manual-evaluated script accuracy in controlled settings (e.g., 69\%$\rightarrow$95\%)~\cite{yuan-etal-2023-distilling}, suggest required paradigm shifts, discussed further in~\S\ref{sec:future_state}.}

\subsubsection{Question Answering}
\label{sssec:qa}

QA for procedural text typically requires an understanding of action sequences, causal relationships between steps, and implicit knowledge of the process described. Numerous datasets have been developed to support this task. For instance, datasets sourced from \textbf{WikiHow} include WikiHowQA~\cite{Bolotova-Baranova23}, the WikiHow dataset~\cite{zhou-etal-2019-learning-household}, PARADISE~\cite{uzunoglu-etal-2024-paradise}, and other WikiHow variants~\cite{wikiHowStepInference, yang-etal-2021-visual,UzunogluS23}. \textbf{Everyday task} datasets include MCScript~\cite{Mcscript} and its successor, MCScript2.0~\cite{Mcscript2}, along with HellaSwag~\cite{zellers-etal-2019-hellaswag}. Additional datasets address other domains, such as RecipeQA~\cite{RecipeQA} for \textbf{recipes} and \textbf{social interactions}~\cite{bauer-bansal-2021-identify}. QA in procedural text either focuses on reasoning or reading comprehension. \textbf{Reading comprehension} benchmarks~\cite{RecipeQA, Bolotova-Baranova23, zhou-etal-2019-learning-household} mostly rely on retrieving correct information directly from the text without requiring deeper inferences. In contrast, \textbf{reasoning-based} benchmarks~\cite{Mcscript,Mcscript2, bAbI, tandon-etal-2019-wiqa, zellers-etal-2019-hellaswag, bauer-bansal-2021-identify, yang-etal-2021-visual, wang-etal-2023-steps} require models to go beyond merely extracting information, such as understanding causal relationships, making inferences (e.g., determining the effect of a missing step in a cooking recipe), predicting outcomes (e.g., what would happen if a particular process in a machine were altered), or applying commonsense knowledge to answer questions. Further categorization can be made based on whether the approach requires answering from a single document or multiple documents. \textbf{Single-document} retrieval involves tasks where models must derive answers from a single text or data source~\cite{Mcscript,Mcscript2, tandon-etal-2019-wiqa,zellers-etal-2019-hellaswag, wang-etal-2023-steps, RecipeQA, zhang2023entity, zhou-etal-2019-learning-household}, while \textbf{multiple-document} retrieval~\cite{Bolotova-Baranova23, yang-etal-2021-visual, wikiHowStepInference, bauer-bansal-2021-identify} requires synthesizing information from various documents to answer questions.

\clrev{
Reading comprehension uses LSTM-based or fine-tuned transformer models (BERT, RoBERTa)~\cite{zhou-etal-2019-learning-household, RecipeQA, Bolotova-Baranova23}, while reasoning tasks employ knowledge bases like ConceptNet~\cite{bauer-bansal-2021-identify} and reinforcement learning for strategic reasoning~\cite{zhang2023entity}. Evaluation uses accuracy for multiple-choice and ROUGE/BLEU for abstractive QA~\cite{Mcscript, Bolotova-Baranova23}.}

\clrev{
\textbf{Limitations and Open Questions}
Despite 70--94\% accuracy on classification-style benchmarks~\cite{wikiHowStepInference, tandon-etal-2019-wiqa, uzunoglu-etal-2024-paradise}, only 8--11\% of predictions on TRIP are supported by fully correct reasoning chains~\cite{storks-etal-2021-tiered-reasoning}. This stark gap: high task accuracy with near-zero verifiable reasoning reveals models exploit surface correlations rather than understanding procedures. Three task-specific deficiencies emerge: (1) \textbf{keyword dependence}: dropping overlapping goal-candidate keywords induces 15--20 point accuracy drops~\cite{uzunoglu-etal-2024-paradise}; (2) \textbf{unverified successes}: 31\% of instances are correct but have entirely wrong underlying state chains~\cite{storks-etal-2021-tiered-reasoning}; (3) \textbf{counterproductive supervision}: adding end-task supervision collapses verifiability from 9--11\% to 0--1\%~\cite{storks-etal-2021-tiered-reasoning}. Furthermore, human--model gaps of up to 10--22 points persist on tasks requiring reasoning (essential-step detection, warning inference)~\cite{wang-etal-2023-steps, uzunoglu-etal-2024-paradise}. These evaluation and reasoning gaps connect to~\S\ref{ssec:cross_cutting} and~\S\ref{sec:future_eval}.}
\subsubsection{Knowledge Acquisition/Mining}
\label{sssec:knowledge}
This task focuses on constructing comprehensive semantic knowledge structures by identifying and formalizing entities, actions, relationships, and processes from unstructured sources (e.g., natural language instructions). The goal is to enable AI systems to better reason, plan, and execute procedural tasks. Unlike process extraction and parsing, which focus on identifying sequential steps or syntactic structures, knowledge acquisition emphasizes building rich contextual understanding and relationships between concepts. Again WikiHow serves as the primary source~\cite{chen-etal-2020-trying, jung2010Automatic, ChuTW17, steinert2020planning, sen2023task2kb, ManualToPDDL}, in addition to scenarios and instruction manuals~\cite{Babli2019ACK, ManualToPDDL}. Knowledge acquisition encompasses two main areas: 1) \textbf{Task Mining} extracts procedural knowledge (tasks, sub-tasks, methods) from instructional texts to support task-oriented conversational agents, recommendation systems, and knowledge bases~\cite{sen2023task2kb, ChuTW17, chen-etal-2020-trying, jung2010Automatic}, and 2) \textbf{Domain Acquisition} formalizes knowledge from natural language task descriptions into structured representations (e.g., \gls{PDDL} models) for automated planning systems. These can be either static models~\cite{steinert2020planning, ManualToPDDL} or dynamically adaptable ones that acquire new knowledge during execution~\cite{Babli2019ACK}.

\clrev{
Earlier methods use rule-based systems and syntactic parsing~\cite{ChuTW17, steinert2020planning, Babli2019ACK}, while recent work employs pretrained models (BERT, LongFormer) with SRL~\cite{chen-etal-2020-trying, sen2023task2kb}. Task Mining evaluates against human annotations; Domain Acquisition tests PDDL executability in simulated environments~\cite{steinert2020planning, jung2010Automatic}.}

\clrev{
\textbf{Limitations and Open Questions.}
Performance remains severely limited: PDDL generation produces a valid plan for only 50\% of instances (with just 0.97 solved actions on average per instance. i.e., actions whose parameters are available in the evaluated sentence)~\cite{steinert2020planning}, while action identification achieves 21.21\% recall@1 and object typing only 11.07\%~\cite{chen-etal-2020-trying}. These failures stem from fundamental data characteristics: 68.3\% of action types and 88.2\% of object types occur fewer than 10 times, creating a classic rare-data / long-tail regime (i.e., modeling long-tail distributions) where rare-but-valid actions are easily treated as noise~\cite{chen-etal-2020-trying}. More critically, in around 91.2\% of processes the action type labels differ from the predicates of associated events, and 84.2\% of processes have object type labels that do not appear as event objects~\cite{chen-etal-2020-trying}---purely extractive approaches struggle to bridge this semantic gap. The task requires compositional generalization from descriptive language to formal representations, a capability current architectures lack. These limitations exemplify the implicit knowledge and compositional reasoning gaps discussed in~\S\ref{ssec:cross_cutting} and~\S\ref{sec:future_implicit}.}

\subsection{Grounded Tasks}
\label{ssec:grounded_tasks}

    The final category of tasks we consider is \textbf{grounded tasks}, which aim to perform instructions that can be represented as a sequence of actions in a symbolic language on \textbf{simulated environments} (e.g., mobile phone, 3D simulation), external \textbf{knowledge resources} (e.g., querying a database, searching the web), or in \textbf{structured or unstructured documents} through multi-turn task-oriented dialogues. These tasks typically require understanding and utilizing contextual information and domain-specific knowledge to effectively perform the tasks within specific domains, such as web, mobile, game, robotics, and navigation. We build a taxonomy of environments that are commonly used in literature, namely as Web~(\S\ref{ssec:web}), Navigation/Household~(\S\ref{ssec:Navigation}), Robotic~(\S\ref{ssec:robot}), Game~(\S\ref{ssec:game}), and GUI~(\S\ref{ssec:Mobile}); and provide the list of papers for each environment in Appendix~\ref{ssec:grounded_env_papers}. 
    
    Meanwhile, \textbf{grounded task-oriented dialogue systems} build upon the principles of grounded tasks by enabling interactive, language-based guidance within various environments. These systems ground their responses in external sources such as documents~\cite{feng-etal-2020-doc2dial, feng2021multidoc2dial, strathearn-gkatzia-2022-task2dial} or real-time environmental data~\cite{Narayan-ChenJH19, TEACh}. For instance, they can guide users to navigate in dynamic virtual environments~\cite{Narayan-ChenJH19, TEACh}, or in troubleshooting and household management through static resources~\cite{feng-etal-2020-doc2dial, feng2021multidoc2dial, strathearn-gkatzia-2022-task2dial}. Recent advancements in \gls{NLP}, including \gls{LLM}s and \gls{RAG} frameworks, have expanded the capabilities of these systems, making them essential in fields such as customer support and healthcare~\cite{feng2021multidoc2dial, FlowDial}. We discuss them in detail in \S\ref{ssec:dialogue}. \clrev{Furthermore, Table~\ref{tab:grounded-environments} provides a unified view of the diverse grounded task environments, comparing their structural properties (state/action spaces, observability) and evaluation metrics.}
    \begin{table}[!htp]
\centering
\scalebox{0.85}{
\footnotesize
\begin{tabular}{m{1.8cm}|m{2.8cm}|m{2.5cm}|m{1.8cm}|m{2.5cm}}
\hline
\textbf{Environment} & \textbf{State Space} & \textbf{Action Space} & \textbf{Observability} & \textbf{Key Metrics} \\
\hline
Web & Webpage content \newline DOM \newline past actions & Click \newline type \newline query \newline navigate & Partial (POMDP) & SR \newline EM \newline Fuzzy Match \\
\hline
Navigation & Spatial layouts \newline object locations & Camera \newline move \newline manipulate & Partial/Full & SR \newline Path Length \newline GCS \newline LCS \\
\hline
Robotics & Robot pose \newline objects \newline joints & Navigate \newline grasp \newline manipulate & Partial & Task SR \newline Sequence Completion \\
\hline
Game & Board/world state \newline positions & Move pieces \newline manipulate & Varies & Task Success \newline Move Accuracy \\
\hline
GUI/Mobile & View hierarchy \newline UI elements & Tap \newline swipe \newline drag & Partial & Task SR \newline Accuracy \\
\hline
Dialogue & Utterances + doc/env state & Actions \newline retrieval & Partial/Full & SR \newline GCS \newline EM \newline F1 \newline BLEU \\
\hline
\end{tabular}
}
\caption{Comparison of grounded task environments. SR = Success Rate, GCS = Goal-Condition Success, EM = Exact Match, LCS = Longest Common Subsequence.}
\label{tab:grounded-environments}
\end{table}

\subsubsection{Web Environments}
\label{ssec:web}

    These environments mimic real-world web interactions, tasking agents with activities from information seeking~\cite{WebArena23, deng-etal-2024-multi} to e-commerce~\cite{WebShopN22, ShiKFHL17, LiuGPSL18}. 
    It is generally defined as partially observable Markov decision processes(POMDP)~\cite{zhu2024knowagent}, where the state is the webpage content or past actions~\cite{Mind2Web2023, WebArena23}; and actions are clicking, typing, and sending queries~\cite{Mind2Web2023, WebArena23, LiuGPSL18}.
    
    \clrev{Web agents perform step-by-step actions, interacting with DOM Trees, screenshots, or forms~\cite{Russ2021, WebArena23, ShiKFHL17}, serving as benchmarks for language models~\cite{WebArena23} and reinforcement learning~\cite{ShiKFHL17, LiuGPSL18}. Methods range from BERT-based models~\cite{Russ2021} to GPT-4~\cite{Mind2Web2023} and WebGPT~\cite{Nakano2021WebGPTBQ}, with RL approaches using binary success/failure rewards~\cite{ShiKFHL17, LiuGPSL18} or action-specific evaluation~\cite{WebShopN22}. Annotation granularity varies from command/navigation instructions~\cite{FLINMazumderR21} to actions with HTML snippets~\cite{Mind2Web2023}. Evaluation uses exact/fuzzy match and success rates~\cite{WebArena23, Mind2Web2023, ShiKFHL17}.}

    \clrev{\textbf{Limitations and Open Questions.}
    The 63-point gap between GPT-4 (14.41\%) and humans (78.24\%) on WebArena~\cite{WebArena23} reveals fundamental architectural limitations rather than scaling issues. We suggest that three interacting factors might be driving this failure. First, \textbf{context overflow}: real-world HTML contains 7,000--14,000 tokens versus $\sim$500 in simulators, forcing lossy summarization that discards task-relevant information~\cite{gur2024realworld}. Second, \textbf{error cascades}: poor HTML understanding produces incorrect plans, which degrade state tracking in subsequent steps---a destructive feedback loop without closed-loop recovery mechanisms~\cite{Kim2023LanguageMC, WebArena23}. Third, \textbf{uncertainty blindness}: in the \emph{UA-hint} setting---where the prompt explicitly tells the agent it may declare a task ``unachievable''---GPT-4 erroneously identifies 54.9\% of feasible tasks as impossible (an effect WebArena attributes primarily to the hint), causing premature termination rather than strategic exploration~\cite{WebArena23}. Finally, planning errors constitute the critical bottleneck: incorrect task decomposition causes most failures and compounds over multi-step trajectories~\cite{gur2024realworld}. This suggests progress requires architectures with HTML-specific inductive biases and explicit uncertainty quantification: the challenges we return to in~\S\ref{sec:future_grounding} and~\S\ref{sec:future_recovery}.}

\subsubsection{2D/3D Navigation/Household Environment}   
\label{ssec:Navigation}
        
    This environment involves simulated settings where autonomous agents interpret natural language instructions to execute navigation tasks~\cite{shi-etal-2024-opex}. VirtualHome~\cite{VirtualHome} is one such environment, characterized by spatial layouts agents must navigate, often using maps or virtual simulations~\cite{shi-etal-2024-opex}. Common tasks are household assistance~\cite{VirtualHome, artist} and virtual navigation~\cite{SayNav}, with some environments focused solely on navigation~\cite{REVERIE, REVE-CE, MacMahonSK06, zang-etal-2018-translating, artzi-zettlemoyer-2013-weakly, Anderson2017VisionandLanguageNI} and others on complex, multi-step tasks in household settings~\cite{ALFRED20, VirtualHome, FAVOR, zhou-etal-2022-hierarchical-control, FengFLW22, MisraBBNSA18, embodiedQA}. Environments vary in structure and focus: REVE-CE~\cite{REVE-CE} and REVERIE~\cite{REVERIE} emphasize visual navigation, while others use linguistic-driven navigation, guiding agents with natural language instructions~\cite{MapTaskVogelJ10, artzi-zettlemoyer-2013-weakly, zang-etal-2018-translating}. Meanwhile, VirtualHome~\cite{VirtualHome, artist} and FAVOR~\cite{FAVOR} involve object manipulation and rearrangement through detailed instructions and AR-driven systems. Some, like ALFRED~\cite{ALFRED20}, integrate both visual and linguistic inputs to enhance navigation and interaction capabilities~\cite{misra-etal-2015-environment, zhou-etal-2022-hierarchical-control}. The studies differ in terms of \textbf{observability}, with some, like VirtualHome~\cite{VirtualHome}, offering \textbf{complete map awareness}, while others, like ALFRED~\cite{ALFRED20}, present \textbf{partial observability} where agents must explore and adapt~\cite{embodiedQA, REVE-CE, REVERIE, MacMahonSK06, Anderson2017VisionandLanguageNI, FengFLW22, MapTaskVogelJ10}. This dynamic adaptation makes tasks more challenging as agents gather new information and encounter obstacles~\cite{REVE-CE, REVERIE, ALFRED20}. The \textbf{action space is often limited} and repeated across environments, involving tasks like camera adjustments, moving through spaces, and manipulating objects~\cite{ou2024worldapisworldworthapis, Anderson2017VisionandLanguageNI, REVERIE, MapTaskVogelJ10, REVE-CE, VirtualHome, FAVOR}. Task complexity in environments like REVERIE~\cite{REVERIE} and REVE-CE~\cite{REVE-CE} is increased by adding object identification or manipulation~\cite{FAVOR, VirtualHome, ALFRED20}.
        
    \clrev{Methods have evolved from sequence-to-sequence models with attention and reinforcement learning~\cite{Anderson2017VisionandLanguageNI, REVERIE} to LLM-based planners combining language models with hierarchical planning~\cite{Song_2023_ICCV, SayNav}; see \citet{NavSurvey} for detailed coverage. Evaluation uses success rate and path length~\cite{Anderson2017VisionandLanguageNI, ALFRED20}, with variations including navigation-plus-object-identification in REVERIE~\cite{REVERIE}, Longest Common Subsequence in VirtualHome, and path-weighted success rates in ALFRED~\cite{ALFRED20}.}

    \clrev{\textbf{Limitations and Open Questions.}
    A large gap remains between human performance on ALFRED (91\% task success on test-unseen) and automated agents~\cite{ALFRED20}. The original ALFRED baselines achieved only 2.1--4.0\% task success on test-seen and about 0.4--0.5\% on test-unseen~\cite{ALFRED20}. More recent \textbf{LLM-based} agents substantially improve performance (e.g., OPEx-S reports 44.55\% on test-seen and 42.25\% on test-unseen)~\cite{shi-etal-2024-opex}, yet still fall short of human success. Moreover, OPEx shows that oracle perception (perfect depth and instance segmentation/object masks) yields a large boost (e.g., 59.43\% success on valid-unseen), but still does not close the gap~\cite{shi-etal-2024-opex}. Sub-goal analysis reveals a striking asymmetry: agents achieve approximately 90\% success on object-agnostic sub-goals like heating or cooling items~\cite{ALFRED20}, but performance is much lower for navigation-dependent sub-goals such as going to and picking up objects (about 51\% and 32\% on seen, dropping to about 22\% and 21\% on unseen)~\cite{ALFRED20}.} \clrev{This pattern suggests that agents learn basic interactions but lack the \textbf{spatial reasoning} and \textbf{long-horizon planning} needed to reliably localize targets, maintain spatial context over time, and recover from partial observability. These navigation-heavy failures are consistent with limitations in building and using structured spatial representations, and they relate to broader state-tracking and memory challenges discussed in~\S\ref{sssec:entity_tracking}.}

\subsubsection{Robotic Environments} 
\label{ssec:robot}

    We differentiate robotic environments from 2D/3D environments, as the former focuses on enabling real robots to understand and carry out tasks based on natural language instructions in dynamic settings~\cite{Howard2014ANL, Matuszek2012LearningTP}, whereas the latter refers to simulated environments. All surveyed papers in this category involve robots or robotic systems executable on robots~\cite{nair2021learning}. Key features include integrating sensory inputs like visual, tactile\cite{Mees2021CALVINAB}, and proprioceptive data~\cite{nair2021learning}, allowing robots (e.g., Franka Emika Panda robot arm~\cite{Mees2021CALVINAB, Singh2022LearningNP, nair2021learning}, PR2~\cite{Misra2014TellMD, Bucker2022ReshapingRT}, Baxter~\cite{Scalise2018NaturalLI}, robotic lift~\cite{Tellex2011ApproachingTS, Matuszek2012LearningTP} etc...) to interact effectively with their surroundings. 
    Robots mostly operate in \textbf{partially observable} environments, interpreting complex language commands for tasks like \textbf{navigation, manipulation, and trajectory reshaping}~\cite{Bucker2022ReshapingRT, Singh2022LearningNP, Tellex2011ApproachingTS}. State spaces typically include the robot's position, object locations, and environmental features. For example, \cite{Tellex2011ApproachingTS, TellexKDWBTR11, Misra2014TellMD} define locations and paths using semantic maps, while \citet{Mees2021CALVINAB} and others focus on object positions and robot joints~\cite{bisk-etal-2016-natural, Bucker2022ReshapingRT}. CALVIN~\cite{Mees2021CALVINAB} uses multiple cameras to capture sensory inputs, focusing on RGB data, while other papers incorporate depth images for multi-step tasks~\cite{Singh2022LearningNP}.
    
    \clrev{Methods combine probabilistic and neuro-symbolic approaches~\cite{TellexKDWBTR11, Singh2022LearningNP} using tools like Generalized Grounding Graphs~\cite{TellexKDWBTR11}, imitation learning and offline reinforcement learning~\cite{nair2021learning, Mees2021CALVINAB, InstructRobot2025}, and more recently LLM-based planning with RL execution (e.g., SayCan~\cite{SayCan}). We refer the readers to \citet{Khan2020systematicRLRobot, tang2024drlRobot} for comprehensive RL-robotics coverage. Evaluation focuses on task success rates, sequence completion, and trajectory prediction accuracy.}

    \clrev{\textbf{Limitations and Open Questions.} Results across robotic manipulation benchmarks reveal three recurring limitations. \textbf{(1) Long-horizon execution collapses under multi-step instruction sequences.} In CALVIN, a representative imitation-learning baseline succeeds on 48.9\% when executing a single instruction but drops to 0.08\% when required to execute five instructions sequentially \cite{Mees2021CALVINAB}. In SayCan, overall execution success is 74\% in the mock-kitchen setting, yet the long-horizon family achieves only 47\% execution success \cite{SayCan}. These outcomes indicate that even when individual skills can be executed, chaining multiple language-specified subtasks remains unreliable and likely amplifies small errors over time. \textbf{(2) Robustness to environment variation remains limited.} In CALVIN, success decreases from 53.9\% when training and testing in the same environment to 35.6\% when training across multiple environments and testing in a held-out one \cite{Mees2021CALVINAB}. In SayCan, overall execution drops from 74\% in a mock kitchen to 60\% in a real kitchen \cite{SayCan}. Together, these results suggest that transferring language-conditioned behavior across setups is still sensitive to perception and low-level control details. \textbf{(3) Key linguistic phenomena and reference ceilings are under-specified.} SayCan explicitly notes that the base system does not handle negation \cite{SayCan}, and neuro-symbolic program learning reports drops performance when relational concepts are present \cite{Singh2022LearningNP}.}

\subsubsection{Game Environments} 
\label{ssec:game}

    Simulating real-world challenges in game environments, e.g., strategic planning in chess~\cite{Toshniwal2021ChessAA}, collaborative building in Minecraft~\cite{Narayan-ChenJH19, jayannavar-etal-2020-learning}, and procedural level generation in educational games~\cite{Hooshyar2018ADP}, is common. We further divide game environments w.r.t. the task they focus on: \textbf{Structured, Strategic, and Predictive Environments} feature deterministic environments like chess and Othello~\cite{Toshniwal2021ChessAA, Li2022EmergentWR}, where state, action, and observation spaces are defined by the positions and legal moves on the board. \textbf{Collaborative and Interactive Virtual Worlds} like Minecraft involve 3D environments where agents manipulate blocks and navigate using a limited view of the world~\cite{Narayan-ChenJH19, jayannavar-etal-2020-learning}. These environments focus on collaborative tasks. \textbf{Abstract and Procedural Knowledge Environments} such as HEXAGONS~\cite{DrawMeAFlower22} and ScriptWorld~\cite{ScriptWorld} emphasize abstract reasoning and procedural knowledge. In ScriptWorld, tasks are defined by sequential textual scenarios, while HEXAGONS focuses on tile painting within a hexagonal grid. \textbf{Adaptive and Text-based Environments} like TextWorld~\cite{TextWorld} and SmartPlay~\cite{wu2024smartplay} challenge AI systems with tasks requiring spatial reasoning and decision-making. TextWorld involves text-based navigation, while SmartPlay evaluates agents across diverse mini-games, including Rock-Paper-Scissors and Minecraft-like environments.
    \clrev{Methods vary by environment: Chess~\cite{Toshniwal2021ChessAA} and Othello~\cite{Li2022EmergentWR} use the GPT-2 architecture and Othello-GPT, respectively, for move prediction; Minecraft uses Seq2Seq-style models for the Builder Action Prediction (BAP) task~\cite{Narayan-ChenJH19}; HEXAGONS~\cite{DrawMeAFlower22} uses DeBERTa and T5; TextWorld~\cite{TextWorld} tests RL agents like LSTM-DQN; and SmartPlay~\cite{wu2024smartplay} evaluates \gls{LLM} families (e.g., GPT-style models and open-weight families such as LLaMA and Mistral), including instruction-tuned variants, across games. Evaluation includes task success and communication efficiency~\cite{Narayan-ChenJH19}.}

    \clrev{
    \textbf{Limitations and Open Questions.}
    Performance benchmarks reveal fundamental gaps in interactive reasoning and/or world state tracking: GPT-4 achieves only 26\% of human performance on complex planning (Crafter), and 61\% on 3D navigation (Minecraft). While transformers reach 97.7\% legal move accuracy in chess, exact move prediction drops to 52\%~\cite{wu2024smartplay, Toshniwal2021ChessAA}. In procedural instruction execution, most steps (>60\%) contain abstract instructions, with 50\% at mid-to-high abstraction levels that models struggle to execute correctly~\cite{DrawMeAFlower22}. Similar abstraction demands also arise in script-based environments~\cite{ScriptWorld}.}
    \clrev{
    We hypothesize that three key problems underlie these failures. \textbf{Spatial reasoning}: models generate contradictory directional commands and lose fine-grained spatial grounding in text-based descriptions~\cite{wu2024smartplay, Narayan-ChenJH19}. \textbf{Inadequate world models}: linear probes exhibit $\ge$20\% error versus single-digit error for nonlinear probes (down to $\sim$1.7\%), suggesting that representational structure may be present but not linearly accessible~\cite{Li2022EmergentWR}. The gap between 97.7\% legal move generation and 52\% exact move accuracy suggests that satisfying local rule constraints can coexist with difficulty in selecting strategically appropriate actions, implicating limits in interactive reasoning and/or world-state tracking\footnote{We use \emph{world-state tracking} to mean maintaining a consistent latent representation of the underlying environment state (e.g., a chess board) as it evolves through actions, and \emph{interactive reasoning} to mean selecting actions by anticipating multi-step consequences under that evolving state.}~\cite{Toshniwal2021ChessAA}, while HEXAGONS error rates increase substantially from concrete to high-abstraction instructions~\cite{DrawMeAFlower22}. \textbf{Attention limitations}: even in domains like chess, restricting attention to 50-token windows degrades legal move accuracy by $\sim$2 points (97.7\%→95.8\%), with approximate attention (Reformer) dropping to 88.0\%~\cite{Toshniwal2021ChessAA}.}

    \clrev{These results expose a gap between surface-level pattern learning and robust world modeling—while models may encode latent state, they often fail to reliably use it for compositional spatial and strategic reasoning.}

\subsubsection{GUI Environments} 
\label{ssec:Mobile}

    Mobile environments for grounded task execution, also known as GUI environments, enable AI agents to interact with mobile app interfaces through actions like tapping, swiping, and navigating to perform a diverse set of tasks. Similar to web environments, mobile environments focus on GUI interactions but extend to a broader range of applications, from system-level tasks to diverse fields like education~\cite{MotifBurnsAAKSP22}, business~\cite{Banerjee2023LexiSL, DanyangZhang2023_MobileEnv}, and entertainment~\cite{LiHZZB20}.
    Key datasets and platforms for mobile environments share a focus on vision-language navigation and UI interaction. MoTIF~\cite{MotifBurnsAAKSP22} provides 6,100 natural language tasks with action localization, while AndroidEnv~\cite{AndroidEnv21} offers a reinforcement learning platform for Android, with task-specific annotations. UICaption~\cite{Banerjee2023LexiSL} pairs UI images with captions, though it lacks event annotations. PIXELHELP, ANDROIDHOWTO, and RICOSCA~\cite{LiHZZB20} map natural language to UI actions. Mobile-Env's WikiHow Task Set~\cite{DanyangZhang2023_MobileEnv} offers a benchmark for LLM-based agents with detailed annotations for cross-page navigation and QA tasks.
    
    Action, state, and observation spaces in mobile environments reflect real app interactions. MoTIF~\cite{MotifBurnsAAKSP22} and Mobile-Env~\cite{DanyangZhang2023_MobileEnv} define the state space by app view hierarchy, and actions like tapping and swiping. On the other hand, AndroidEnv~\cite{AndroidEnv21} focuses on pixel-based reinforcement learning interactions, while Lexi~\cite{Banerjee2023LexiSL} expands UI understanding to both mobile and desktop. Observation spaces also vary, including UI object hierarchies~\cite{LiHZZB20}, pixels, and screenshots~\cite{Banerjee2023LexiSL}, while actions involve diverse UI manipulations like dragging and typing. Finally, platforms like AndroidEnv~\cite{AndroidEnv21} and Mobile-Env~\cite{DanyangZhang2023_MobileEnv} simulate real-world operations on actual devices, including Pixel phones for realistic evaluations~\cite{LiHZZB20}, while MoTIF~\cite{MotifBurnsAAKSP22} focuses on task feasibility but does not extend to real-world testing on physical devices, concentrating on controlled simulations. 

    \clrev{Models include GPT-3.5-turbo, GPT-4, LLaMA 2~\cite{DanyangZhang2023_MobileEnv}, and vision models like ViLBERT~\cite{Banerjee2023LexiSL}. Evaluation uses accuracy, recall, completion rates, and rewards~\cite{DanyangZhang2023_MobileEnv, AndroidEnv21}.}

    \clrev{\textbf{Limitations and Open Questions.}
    Even the strongest agents (GPT-4, AgentLM-70B) achieve only 30--43\% success on Mobile-Env's WikiHow tasks, with GPT-4 dropping to $\approx$7.5\% on QA~\cite{ZhangMobileEnvAE}. These failures stem from compounding weaknesses: \textbf{Feasibility blindness}: MoTIF's classifier misclassifies 44\% of infeasible tasks as feasible, causing agents to attempt impossible goals~\cite{MotifBurnsAAKSP22}. \textbf{Representation mismatch}: generic CLIP encoders on screen images outperform domain-specific Screen2Vec on view hierarchies (58.2 vs.\ 33.7 F1), suggesting existing GUI representations may fail to capture functional semantics~\cite{MotifBurnsAAKSP22}. \textbf{Visual grounding brittleness}: GPT-4V achieves only 3--10\% success on pixel-level action prediction, improving only with element-level detection aids~\cite{ZhangMobileEnvAE}. A key next step is to build agents that can recognize infeasibility early and recover gracefully, and to develop UI representations that support compositional, cross-app functional generalization rather than relying on pixel- or layout-level cues.} \clrev{We find the GUI domain to be intrinsically challenging: identical icons serve different functions across apps, and dissimilar widgets implement the same operation, undermining assumptions in vision-language models trained on natural images~\cite{Banerjee2023LexiSL, MotifBurnsAAKSP22}. These patterns reveal that LLM planning advances do not transfer directly to GUI settings, where representation failures and brittle grounding dominate. Progress likely depends on training signals and interaction scaffolds that make grounding robust to UI variability, as well as benchmarks that disentangle planning mistakes from perception/grounding failures so improvements can be targeted.}

\subsubsection{Grounded Task-Oriented Dialogue}
\label{ssec:dialogue}

    Task-oriented dialogue (TOD) systems facilitate specific tasks through conversational interactions. These systems use underlying resources such as documents, databases, or live environmental data to ensure accurate and contextually relevant responses. Researchers have extended TOD systems across various domains, tailoring them to function effectively within interactive and embodied environments, such as guiding users in navigation, construction, or household chores within simulated or virtual settings~\cite{gao2022dialfred, srinet-etal-2020-craftassist, Narayan-ChenJH19, TEACh, moghe-etal-2024-interpreting}. In customer support, TOD systems enhance technical assistance or customer service by utilizing documents or flowcharts~\cite{feng2021multidoc2dial, strathearn-gkatzia-2022-task2dial, FlowDial, feng-etal-2020-doc2dial, chen-etal-2021-action}. Additionally, specialized TOD systems cater to specific needs, such as managing cooking recipes~\cite{Jiang_2022_cookdial}, supporting data visualization~\cite{shao-nakashole-2020-chartdialogs}, simulating interactions in mobile GUIs~\cite{MetaGUISunCCDZY22}, and facilitating storytelling within virtual worlds~\cite{AmmanabroluR21a}.

    Data collection for these dialogue systems is generally expensive, as it typically requires two users to engage in a conversation. Researchers have adopted several approaches to address this challenge. For instance, Wizard-of-Oz methods have been employed in datasets such as META-GUI~\cite{MetaGUISunCCDZY22}, ChartDialogs~\cite{shao-nakashole-2020-chartdialogs}, and CookDial~\cite{Jiang_2022_cookdial}. Human-human interaction is used in the creation of datasets like Minicraft~\cite{Narayan-ChenJH19}, ABCD~\cite{chen-etal-2021-action}, Task2Dial~\cite{strathearn-gkatzia-2022-task2dial}, and TEACh~\cite{TEACh}. Additionally, human annotation is used in datasets such as CraftAssist Instruction Parsing~\cite{srinet-etal-2020-craftassist}, DialFRED~\cite{gao2022dialfred}, Doc2Dial~\cite{feng-etal-2020-doc2dial}, MultiDoc2Dial~\cite{feng2021multidoc2dial}, and FLONET~\cite{FlowDial}. 

    Systems can be grounded on \textbf{documents} and other static data sources, such as flowcharts and knowledge bases~\cite{moghe-etal-2024-interpreting, shao-nakashole-2020-chartdialogs, Jiang_2022_cookdial, chen-etal-2021-action, feng2021multidoc2dial, strathearn-gkatzia-2022-task2dial, feng-etal-2020-doc2dial, FlowDial, AmmanabroluR21a}, where the model has user utterances and grounding documents as input, to generate a sequence of actions or commands. Alternatively, systems can be grounded on the \textbf{environment}~\cite{gao2022dialfred, srinet-etal-2020-craftassist, MetaGUISunCCDZY22, Narayan-ChenJH19, TEACh}, where the input consists of user utterances plus real-time environmental state, and the output is dynamically responsive sequences of actions performed within that environment. This allows for interactions based on navigating physical spaces or manipulating objects within simulations.
        
    \clrev{Transformer-based models dominate, with BERT for retrieval and action tracking~\cite{feng-etal-2020-doc2dial, feng2021multidoc2dial, Jiang_2022_cookdial, chen-etal-2021-action, MetaGUISunCCDZY22, TEACh}, Seq2Seq for instruction generation and parsing~\cite{Narayan-ChenJH19, gao2022dialfred, srinet-etal-2020-craftassist, shao-nakashole-2020-chartdialogs, AmmanabroluR21a}, and RAG for document-grounded response generation~\cite{feng2021multidoc2dial, FlowDial}. Evaluation uses Exact Match and F1 for span selection~\cite{feng-etal-2020-doc2dial, feng2021multidoc2dial}, BLEU for response quality~\cite{shao-nakashole-2020-chartdialogs, feng-etal-2020-doc2dial, chen-etal-2021-action}, and Success Rate/Goal Completion for task achievement~\cite{gao2022dialfred, TEACh, FlowDial, chen-etal-2021-action}. See App.~\ref{app:tods} for dataset details.}

    \clrev{\textbf{Limitations and Open Questions.}
    Benchmarks expose systematic, not marginal gaps: ABCD shows 50.8 points between RoBERTa-Large (31.9\%) and humans (82.7\%)~\cite{chen-etal-2021-action}; TEACh collects sessions with a 74.17\% human success rate, while baseline models achieve 4.8--7.06\% success on the EDH benchmark~\cite{TEACh}. These gaps likely reflect three intertwined failure modes.}
    \clrev{\textbf{Structure blindness}: the Episodic Transformer drops from 38.24\% on single-instruction ALFRED to 5--7\% on TEACh's multi-turn dialogues; TEACh further uses unimodal ablations to test whether models are simply memorizing action sequences, suggesting flat sequence models struggle to learn hierarchical task structure~\cite{TEACh}. \textbf{Retrieval brittleness}: FLODial achieves only 66.1\% Recall@1 on unseen flowcharts, with 45.5\% of errors on non-neighboring nodes, showing retrieval-augmented systems fail to reason over graph topology~\cite{FlowDial}. \textbf{Difficulty with abstraction}: rule-based agents achieve 0\% success on several compositional TEACh tasks despite 150 hours of engineering; DialFRED shows generalization drops from 25.4--47.8\% (seen) to 18.3--33.6\% (unseen)~\cite{TEACh, gao2022dialfred}.}
    
    \clrev{The compositional nature of grounded dialogue---requiring simultaneous procedural grounding, long-horizon planning, multi-modal integration, and commonsense reasoning (e.g., TEACh averages $\sim$165 actions per session with $\sim$14 utterances, with irreversible operations like slicing and toasting)~\cite{TEACh, strathearn-gkatzia-2022-task2dial}---suggests that isolated component improvements are insufficient. Both supervised models and hand-crafted rules lack mechanisms for emergent hierarchical abstraction. They also lack robust error detection and recovery strategies that humans employ naturally. Together, these results point to the need for integrated end-to-end solutions rather than isolated component improvements. This raises a central open question: in end-to-end settings, which failures are primary versus downstream effects, and what mechanisms enable hierarchical abstraction and reliable recovery under irreversible state changes?}

\subsection{Cross-Cutting Challenges}
\label{ssec:cross_cutting}

\clrev{While the preceding sections document task-specific findings, several fundamental limitations recur across benchmarks. Table~\ref{tab:cross-cutting-summary} summarizes these six challenges; the paragraphs below synthesize evidence from multiple tasks and offer our interpretation of why these limitations persist, necessarily bounded by the solutions attempted thus far.}

\begin{table*}[t]
\centering
\small
\scalebox{0.8}{
\begin{tabular}{p{3cm}p{5.2cm}p{7cm}}
\toprule
\textbf{Challenge} & \textbf{Core Problem} & \textbf{Affected Tasks} \\
\midrule
Implicit Knowledge & Cannot reason about information that is expected but not explicitly stated & 
Entity Tracking~\cite{OpenPI}, 
Knowledge Acquisition~\cite{chen-etal-2020-trying, steinert2020planning}, 
Alignment~\cite{DonatelliSBKZK21, wanzare-etal-2017-inducing}, 
Implicit Detection~\cite{roth-anthonio-2021-unimplicit, AnthonioBR20} \\
\addlinespace
\midrule
State Maintenance & Performance degrades sharply as the number of steps increases & 
Entity Tracking~\cite{KimS23, OpenPI}, 
Robotics~\cite{Mees2021CALVINAB}, 
Navigation~\cite{ALFRED20, shi-etal-2024-opex} \\
\addlinespace
\midrule
Evaluation & High benchmark accuracy does not reflect genuine procedural understanding & 
QA~\cite{storks-etal-2021-tiered-reasoning, uzunoglu-etal-2024-paradise}, 
Process Extraction~\cite{zhang-etal-2024-proc2pddl, ProcessModelExtraction}, 
Summarization~\cite{kwon2024ordersum, DeChant2022SummarizingAV} \\
\addlinespace
\midrule
Compositional Generalization & Success on atomic operations does not transfer to composed sequences & 
Generation~\cite{ye2025longproc, proScript}, 
Induction~\cite{zhang-etal-2020-analogous}, 
Dialogue~\cite{TEACh, gao2022dialfred} \\
\addlinespace
\midrule
Grounding & Language understanding does not transfer to action execution & 
Web~\cite{WebArena23, gur2024realworld}, 
Navigation~\cite{ALFRED20, shi-etal-2024-opex}, 
Robotics~\cite{SayCan, Mees2021CALVINAB}, 
GUI~\cite{MotifBurnsAAKSP22, ZhangMobileEnvAE} \\
\addlinespace
\midrule
Error Accumulation & Lack mechanisms to detect and recover from intermediate failures & 
Web~\cite{WebArena23, Kim2023LanguageMC}, 
Robotics~\cite{Mees2021CALVINAB, SayCan}, 
Dialogue~\cite{TEACh, chen-etal-2021-action} \\
\bottomrule
\end{tabular}
}
\caption{\clrev{Cross-cutting challenges in procedural text understanding. Each challenge recurs across multiple tasks, suggesting fundamental limitations rather than task-specific difficulties.}}
\label{tab:cross-cutting-summary}
\end{table*}

\clrev{
\subsubsection{The Implicit Knowledge Problem}
\label{sssec:challenge_implicit}
Consider a repair manual that instructs users to ``reassemble in reverse order.'' A human reader effortlessly expands this into the specific sequence of steps performed earlier, now inverted. Or consider a recipe that says ``saut\'{e} until fragrant'': readers infer appropriate heat levels, timing, and what ``fragrant'' means for onions versus garlic. Procedural text is saturated with such implicit content: studies of entity tracking find that roughly 40\% of \emph{referred-to} entities are unmentioned in the text~\cite{OpenPI}; analyses of knowledge acquisition show that a large majority of action labels require inference beyond surface lexical matching~\cite{chen-etal-2020-trying}; and event alignment work reveals systematic failures on functionally equivalent actions that differ only lexically~\cite{DonatelliSBKZK21}. Tests that isolate implicit reasoning show a stark split: in the UnImplicit shared task, approaches achieve over 90\% accuracy on pronoun-based clarifications but only up to 56\% accuracy on harder categories such as modifiers, quantifiers, and modal verbs~\cite{roth-anthonio-2021-unimplicit}.}

\clrev{Why does this remain difficult? We hypothesize that the core issue is representational: approaches (from sequence-to-sequence models through encoder-based transformers to large language models) learn to process tokens that are \textit{present}, but procedural reasoning requires attending to what is \textit{absent}. Training objectives that optimize for next-token prediction on observed text do not directly reward reasoning about expected-but-missing elements. Whether future approaches can address this through explicit world models, abductive reasoning modules, or training objectives that reward inference over implicit content remains an open question.
}
\clrev{
\subsubsection{The State Maintenance Problem}
\label{sssec:challenge_state}
Consider tracking a piece of dough through a bread recipe: it begins as ``flour and water,'' becomes ``the dough'' after mixing, transforms into ``the risen dough'' after proofing, and finally ``the loaf'' after baking. Each reference points to the same physical entity in different states. This kind of state tracking is fundamental to procedural understanding, yet proves remarkably fragile across tested approaches. In controlled entity tracking experiments, accuracy on single-state cases exceeds 85\%, but collapses below 5\% when tracking entities through multiple state changes, even when the entire procedure fits within context limits~\cite{KimS23}. Robotic manipulation shows parallel degradation: 54\% success on single instructions drops to below 1\% on five-step chains~\cite{Mees2021CALVINAB}.}

\clrev{The sharp shape of this degradation suggests that tested approaches perform step-by-step pattern matching without maintaining coherent state representations. One notable finding offers a clue: only approaches pretrained on source code, where variables are explicitly declared and state updates syntactically marked, show improved entity tracking~\cite{KimS23}. This suggests that natural language procedural text may lack the structural cues needed to learn robust state maintenance from text alone, pointing toward training regimes that incorporate explicit state-update supervision or draw from programming language semantics.
}
\clrev{
\subsubsection{The Evaluation Gap}
\label{sssec:challenge_eval}
A question-answering system might correctly predict that ``the cake will be dry'' if the eggs are omitted, but did it reason through the causal chain (eggs provide moisture and binding); or merely associate ``omit ingredient'' with ``bad outcome''? Across procedural benchmarks, we find a troubling pattern: high task accuracy masks shallow reasoning. QA systems achieve 70--90\% accuracy while producing verifiably correct reasoning chains less than 11\% of the time~\cite{storks-etal-2021-tiered-reasoning}. For process extraction, Proc2PDDL shows that intrinsic action prediction accuracy remains low (18--21\%), and that executing predicted PDDL solves only 36--45\% of planning problems~\cite{zhang-etal-2024-proc2pddl}. Summarization metrics reward lexical overlap but cannot detect step-ordering violations that destroy procedural coherence~\cite{kwon2024ordersum}.}

\clrev{These patterns reveal that standard metrics reward surface similarity rather than procedural validity. More troubling, training directly on end-task objectives can \textit{worsen} interpretable reasoning: one study found that adding task-specific supervision collapsed verifiable reasoning from roughly 10\% to near zero~\cite{storks-etal-2021-tiered-reasoning}. This suggests that approaches learn shortcuts that inflate benchmark scores while bypassing the procedural understanding those benchmarks were designed to measure.
}
\clrev{
\subsubsection{The Compositional Generalization Gap}
\label{sssec:challenge_general}
Consider the difference between knowing how to chop vegetables and how to make a stir-fry. The latter requires not just executing ``chop,'' ``heat oil,'' and ``add ingredients'' individually, but sequencing them correctly, tracking what has been added, and adjusting timing based on ingredient properties. Across procedural benchmarks, approaches that succeed on atomic operations fail when those operations must be composed. Process induction achieves reasonable performance on predicting individual verbs but degrades sharply when predicting complete event structures that combine verbs with arguments and temporal relations~\cite{zhang-etal-2020-analogous}. Dialogue systems collapse from 38\% success on single instructions to 5--7\% on multi-turn interactions~\cite{TEACh}. Procedural generation degrades as output length increases, with detailed error analysis revealing that failures stem from incorrect state inferences across steps, rather than from forgetting earlier content~\cite{ye2025longproc}.}

\clrev{Is this simply a memory or length limitation? Several findings suggest otherwise. The degradation in process induction occurs when moving from verbs to complete structures at the same length~\cite{zhang-etal-2020-analogous}. Generation failures are state-inference errors, not content-forgetting errors~\cite{ye2025longproc}. The pattern suggests that tested approaches lack hierarchical abstraction: they cannot decompose complex procedures into subgoals, execute them while tracking intermediate states, and verify results before proceeding.
}
\clrev{
\subsubsection{The Grounding Gap}
\label{sssec:challenge_ground}
An agent that can fluently describe how to navigate a website may nonetheless fail catastrophically when asked to actually book a flight. This gap between language understanding and action execution, which we term the grounding gap, persists across interactive environments. On web navigation, household robotics, and embodied dialogue, the gap between automated approaches and human performance spans 60--85 percentage points~\cite{WebArena23, ALFRED20, TEACh}.} 
\clrev{What causes this gap? One hypothesis is perceptual: perhaps language understanding is adequate but environmental perception fails. However, experiments providing oracle perception, including perfect object segmentation and ground-truth scene descriptions, show the gap remains largely unchanged~\cite{shi-etal-2024-opex}. Analysis of navigation failures reveals a telling asymmetry: agents achieve roughly 90\% success on perception-light sub-goals (heating an object, cooling an object) but drop to 20--30\% on navigation-dependent ones (finding objects, picking them up)~\cite{ALFRED20}. Error analyses in robotic settings find that roughly 65\% of failures stem from high-level planning mistakes, such as selecting incorrect actions or action sequences, rather than low-level execution errors~\cite{SayCan}.}

\clrev{These findings suggest that approaches trained on text have learned associations between words and situations, but not the causal structure of how actions transform world states. They can describe what actions mean without modeling their effects, preconditions, or interactions.
}

\clrev{
\subsubsection{The Error Accumulation Problem}
\label{sssec:challenge_error}
A web agent attempting to book a flight may click the wrong dropdown menu, causing subsequent actions such as selecting dates or choosing seats to operate on the wrong form entirely. Each step then compounds the initial error, and because the agent lacks a mechanism to detect the mistake, the failure propagates unchecked. This error accumulation is particularly damaging in procedural settings, where later steps depend on earlier ones.}

\clrev{The pattern appears across grounded environments. In web navigation, incorrect actions change page state unpredictably, degrading subsequent action selection in a destructive feedback loop~\cite{WebArena23, Kim2023LanguageMC}. In robotic manipulation, failed grasps or incorrect placements cascade into unrecoverable trajectories~\cite{Mees2021CALVINAB}. Perhaps most notably, when given the option to declare tasks unachievable, approaches show poor calibration: over 50\% false-negative rates lead them to abandon feasible tasks while persisting on impossible ones~\cite{WebArena23}. They cannot distinguish ``this task cannot be done'' from ``I do not know how to proceed.''}

\clrev{This contrasts sharply with human procedure-following, in which continuous progress monitoring, error detection, and plan revision are routine. Addressing error accumulation, therefore likely requires explicit mechanisms for uncertainty estimation, state verification, and recovery, moving from open-loop action sequences toward closed-loop systems that detect and correct failures during execution.
}

\clrev{\section{Pointers to Future Research}}
\label{sec:future}
\clrev{The challenges identified in \S\ref{ssec:cross_cutting} point toward specific research directions. We frame each as a shift from current practice toward capabilities that would address fundamental limitations.}

\subsection{Recovering Implicit Knowledge}
\label{sec:future_implicit}

\EmphBox{\textbf{Learning to infer unstated information} instead of \textbf{matching surface patterns}.\\
\noindent\textit{Addresses: Implicit Knowledge (\S\ref{sssec:challenge_implicit}) $\rightarrow$ Entity Tracking, Knowledge Acquisition, Alignment, Implicit Detection}}

\medskip

\clrev{
\noindent Procedural text is rarely self-contained. Studies suggest that a substantial portion of relevant entities---up to 40\% in some analyses---may never be explicitly mentioned~\cite{OpenPI}, and most action labels cannot be recovered through simple word matching alone~\cite{chen-etal-2020-trying}. Instructions routinely assume readers know default values (``season to taste''), typical preconditions (heating a pan before searing), and domain conventions that distinguish similar-sounding actions (``fold'' vs.\ ``stir'').}

\clrev{
Moving beyond surface pattern matching requires architectures that reason about what \textit{should} be present given procedural context. Several directions show promise:}

    \clrev{\textbf{(1) Training on execution traces}. Pairing procedural text with execution logs---from cooking videos, robotic demonstrations, or simulated environments---makes implicit states explicit. This approach has shown encouraging results in robotics, where learning from demonstration provides supervision for unstated preconditions~\cite{nair2021learning}. Extending this paradigm to textual procedures could provide the missing signal for implicit state inference.}
    
    \clrev{\textbf{(2) Neuro-symbolic integration with procedural knowledge bases.} Knowledge resources like \gls{VerbNet} encode typical preconditions and effects for action classes. Early work integrating \gls{VerbNet} for effect prediction~\cite{clark2018leveraging} demonstrates that combining neural flexibility with such structured knowledge enables principled inference about unstated information. Following these encouraging results, future work could explore richer integration---for instance, using knowledge bases to generate candidate implicit states that neural models then verify against context.}
    
    \clrev{\textbf{(3) Self-supervised prediction of omissions.} Training objectives that require predicting deliberately omitted steps or masked entity states could teach models to reason about expected-but-absent content. This approach mirrors successful masked language modeling but targets procedural structure rather than surface tokens. Related work in document understanding has shown that predicting elided content improves downstream reasoning~\cite{czinczoll-etal-2024-nextlevelbert}; adapting such objectives to procedural text is a concrete next step.}

\subsection{Maintaining Coherent World States}
\label{sec:future_state}

\EmphBox{\textbf{Explicitly representing and updating entity states} instead of \textbf{treating procedures as flat token sequences}.\\
\noindent\textit{Addresses: State Maintenance (\S\ref{sssec:challenge_state}) $\rightarrow$ Entity Tracking, Robotics, Navigation}}

\medskip

\clrev{
\noindent State tracking accuracy degrades sharply as procedures lengthen, even when the entire text fits within context limits~\cite{KimS23, Mees2021CALVINAB}. This pattern suggests that access to tokens alone is insufficient; standard architectures lack computational primitives for persistent state maintenance.}

\clrev{
    \textbf{(1) External memory systems.} Memory-augmented architectures such as Infini-attention and external associative-memory \gls{LLM}~\cite{Larimar, munkhdalai2024leavecontextbehindefficient} provide explicit read-write mechanisms that could store entity-attribute bindings updated by each action. While these architectures have shown promise in algorithmic tasks~\cite{bulatov2022RMT, rodkin2024associativerecurrentmemorytransformer}, their application to procedural state tracking remains underexplored. The challenge lies in learning \textit{when} and \textit{what} to write---decisions that require understanding how actions transform entity states.}
    
    \clrev{\textbf{(2) Dynamic structured representations.} Current retrieval-augmented approaches treat structured data (scene graphs, entity databases) as static context to be retrieved, not dynamic state to be updated. This is insufficient for procedural understanding, where each step transforms the world. Future work should explore architectures where structured representations are \textit{modified} by each procedural step: after ``dice the onions,'' the entity representation for onions should update from 
    \texttt{whole} to \texttt{diced}. Graph neural networks operating over entity-relation structures, updated incrementally as procedures unfold, offer one concrete instantiation of this idea.}
    
    \clrev{\clrev{\textbf{(3) Verification through simulation and verifiable rewards.} Recent work on inference-time scaling shows that allocating more computation at test time---e.g., sampling multiple solution paths and selecting via verifiers---can substantially improve performance on reasoning-heavy tasks~\cite{snell2024scalingllmtesttimecompute, liang2024improvingllmreasoningscaling}. For procedural tasks, this additional computation could be used to verify state consistency: after each step, checking whether proposed state updates are coherent with prior states and action semantics. Reinforcement learning with verifiable rewards, where state consistency provides an automatically checkable training signal, is a promising paradigm. Such approaches can \textit{mitigate} some forms of reward hacking by grounding learning in verifiable signals, but remain vulnerable to verifier/specification errors and verifier hacking, making verifier design critical.}}

\clrev{The success of code-pretrained models on tracking tasks~\cite{KimS23} provides an important clue: exposure to explicit variable assignment and state mutation during pretraining creates useful inductive biases. This suggests that procedural corpora augmented with explicit state annotations, or training that incorporates code-like state operations, could substantially improve tracking capabilities.
}

\subsection{Evaluating Procedural Understanding}
\label{sec:future_eval}

\EmphBox{\textbf{Measuring reasoning validity} instead of \textbf{rewarding output similarity}.\\
\noindent\textit{Addresses: Evaluation Gap (\S\ref{sssec:challenge_eval}) $\rightarrow$ QA, Process Extraction, Summarization}}

\medskip

\clrev{
\noindent A recurring pattern across procedural benchmarks is the disconnect between surface metrics and genuine understanding: high accuracy on end tasks coexists with poor performance on reasoning verification; strong token-level scores fail to predict whether outputs actually execute~\cite{storks-etal-2021-tiered-reasoning, zhang-etal-2024-proc2pddl}. This evaluation gap allows---and perhaps encourages---approaches that exploit surface shortcuts rather than developing genuine procedural competence.}

\clrev{\textbf{(1) Intermediate state verification.} Rather than evaluating only final outputs, metrics should require models to produce intermediate states verifiable against ground truth or simulation. For QA, this means evaluating the reasoning chain, not just the answer. For generation, this means checking that each step produces valid state transitions. The TRIP benchmark's tiered evaluation~\cite{storks-etal-2021-tiered-reasoning} provides a template: separating task accuracy from reasoning validity reveals capabilities that aggregate metrics obscure.}

\clrev{\textbf{(2) Robustness to meaning-preserving perturbations.} If a system truly understands a procedure, its performance should be stable under surface variations that preserve procedural meaning---paraphrasing steps, reordering independent actions, substituting equivalent tools. Current benchmarks rarely test this robustness. Developing perturbation-based evaluation would help distinguish genuine understanding from surface pattern matching.}

\clrev{\textbf{(3) Execution-based evaluation for ungrounded tasks.} While grounded environments (web, robotics, navigation) inherently test execution, this paradigm could extend to traditionally text-only tasks. Process extraction outputs could be executed in PDDL planners; generated procedures could be validated in lightweight simulators; summaries could be tested for whether they preserve executable structure. The key insight is that procedural validity is often \textit{objectively verifiable}---unlike open-ended generation where quality is subjective. Building cheap-to-run procedural simulators that verify validity without requiring full embodied deployment could democratize execution-based evaluation and provide automatic training signal.
}

\subsection{Learning Hierarchical Abstractions}
\label{sec:future_hierarchy}

\EmphBox{\textbf{Decomposing procedures into goal hierarchies} instead of \textbf{generating flat action sequences}.\\
\noindent\textit{Addresses: Compositional Generalization (\S\ref{sssec:challenge_general}) $\rightarrow$ Generation, Induction, Dialogue}}

\medskip

\clrev{
\noindent Across procedural benchmarks, success on individual steps fails to transfer to step sequences---a pattern suggesting that tested approaches cannot organize procedures hierarchically~\cite{zhang-etal-2020-analogous, ye2025longproc}. Humans manage complex procedures through nested goal structures: a goal (``make dinner'') decomposes into subgoals (``prepare vegetables,'' ``cook protein,'' ``make sauce''), each with its own sub-procedure and completion criteria. Current approaches generate flat action sequences without this hierarchical scaffolding.}

\clrev{\textbf{(1) Hierarchical generation with explicit goal decomposition.} Rather than generating procedures end-to-end, approaches could first produce a goal hierarchy, then expand each subgoal into concrete steps. This mirrors how humans plan complex tasks and provides natural checkpoints for verification. Hierarchical reinforcement learning provides relevant technique: options frameworks~\cite{Lin2024Hierarchical,nayyar2025Autonomous} separate high-level goal selection from low-level action execution. Adapting these ideas to procedural text generation---where ``options'' correspond to subprocedures like ``make the sauce'' or ``prepare the vegetables''---could enable compositional generalization that flat sequence models lack.}

\clrev{\textbf{(2) Plan-then-execute with modular verification.} Separating high-level planning from low-level generation allows each component to be trained and evaluated independently. Planning modules can be assessed on goal decomposition quality; execution modules on step-level correctness. This separation also enables hybrid approaches where planning uses symbolic methods (ensuring logical consistency) while execution uses neural generation (providing linguistic fluency). The key advantage is \textit{modularity}: failures can be localized to specific components, enabling targeted improvement.}

\clrev{\textbf{(3) Curriculum learning with compositional structure.} Training on short procedures before long ones, with explicit supervision on subgoal completion, could help models learn to compose rather than memorize. The critical design choice is curricula that increase \textit{compositional depth} (more nested subgoals) rather than just sequence length. Work on over-generate-then-filter~\cite{yuan-etal-2023-distilling} demonstrates that explicit verification at intermediate stages dramatically improves output quality---suggesting that learning to verify subgoal completion may be as important as learning to generate steps.
}

\subsection{Grounding Language in Action}
\label{sec:future_grounding}

\EmphBox{\textbf{Learning from interactive execution} instead of \textbf{text-only pretraining}.\\
\noindent\textit{Addresses: Grounding Gap (\S\ref{sssec:challenge_ground}) $\rightarrow$ Web, Navigation, Robotics, GUI}}

\medskip

\clrev{
\noindent Strong performance on text-only benchmarks does not transfer to action execution in interactive environments---substantial gaps persist across web, navigation, and robotic settings~\cite{WebArena23, ALFRED20, TEACh}. Error analyses suggest failures are distributed across perception, navigation/action execution, and long-horizon planning/state modeling; improving perception alone helps substantially but does not solve the task~\cite{SayCan, shi-etal-2024-opex}. This suggests that text-based training teaches associations between words and situations, but not the causal dynamics of action and effect.}

\clrev{\textbf{(1) Pretraining on action-state transition data.} Execution traces---logs that pair actions with resulting state changes---could provide the causal structure that text alone cannot. Sources include robotic demonstration datasets~\cite{nair2021learning}, gameplay recordings, and web interaction logs. The challenge is scale: while text corpora contain billions of tokens, action-state data is orders of magnitude scarcer. Two paths forward are promising: simulation environments that generate unlimited interaction data~\cite{makoviychuk2021isaac,9636667}, and learning from video where state changes are visible but actions must be inferred~\cite{Ko2023Learning,ye2024latent}. Both approaches have shown success in robotics and could extend to procedural language understanding.}

\clrev{\textbf{(2) Active learning through environment interaction.} Most procedural learning assumes passive observation of demonstration data. An underexplored alternative is active learning: systems propose actions, observe consequences, and update their understanding of action-state dynamics. This paradigm---central to reinforcement learning---enables learning directly from the environment's causal structure rather than from human annotations. Developing sample-efficient active learning methods for procedural grounding, where exploration is guided by uncertainty about action effects, is a key open direction.}

\clrev{\textbf{(3) Multi-modal grounding beyond vision.} Current grounding efforts focus heavily on visual input, but procedural execution involves richer sensory feedback: proprioception (knowing where your hands are), haptics (feeling resistance when cutting), and auditory cues (hearing when water boils). Robotic learning increasingly incorporates these modalities~\cite{Mees2021CALVINAB}; extending multi-modal grounding to language understanding could enable more robust procedural execution. The finding that providing perfect visual information does not close the grounding gap~\cite{shi-etal-2024-opex} suggests that vision alone is insufficient---richer sensory grounding may be necessary for robust action execution.}

\subsection{Enabling Error Detection and Recovery}
\label{sec:future_recovery}

\EmphBox{\textbf{Detecting and correcting mistakes} instead of \textbf{executing open-loop action sequences}.\\
\noindent\textit{Addresses: Error Accumulation (\S\ref{sssec:challenge_error}) $\rightarrow$ Web, Robotics, Dialogue}}

\medskip

\clrev{
\noindent When procedures go wrong, errors compound: each mistake corrupts state for subsequent steps, and without detection mechanisms, systems continue executing on increasingly invalid assumptions~\cite{Kim2023LanguageMC, WebArena23, Mees2021CALVINAB}. Addressing this requires tackling two distinct but related problems.}

\clrev{\textbf{The calibration problem.} Systems cannot reliably distinguish what they know from what they do not know. Studies find high false-negative rates on task achievability judgments~\cite{WebArena23} and systematic misclassification of infeasible tasks~\cite{MotifBurnsAAKSP22}. This is fundamentally about self-knowledge: knowing when to proceed confidently versus when to seek clarification, try alternatives, or abandon a path. Improving calibration requires training signals that reward accurate uncertainty estimation, not just task success.}

\clrev{\textbf{The recovery problem.} Even with perfect uncertainty estimation, systems need mechanisms to recover from detected errors. Procedural domains vary in recoverability---clicking the wrong button on a website may be reversible; slicing an ingredient cannot be undone. Effective recovery requires understanding which actions are reversible, how to backtrack when possible, and how to find alternative paths when backtracking fails.}

\clrev{\textbf{(1) Explicit progress monitoring.} Systems should continuously compare expected states (predicted from the procedure) with observed states (from environment feedback). Discrepancies signal potential errors requiring investigation. This closed-loop architecture contrasts with current open-loop approaches that execute action sequences without verification. Implementing such monitoring connects directly to the state representation problem (\S\ref{sec:future_state}): accurate expected-state prediction requires the coherent world models discussed there.}

\clrev{\textbf{(2) Learning recovery policies from failure data.} Beyond detecting errors, systems need strategies for responding. Recovery policies could be learned from human demonstrations of error correction, from simulation with injected failures, or from reinforcement learning with rewards for successful recovery. The robotics literature on fault-tolerant control provides relevant techniques~\cite{kalithasan2024learningrecoverplanexecution}; adapting these to procedural language tasks---where ``faults'' are incorrect actions and ``recovery'' means replanning---is a concrete research direction.}

\clrev{\textbf{(3) Separating uncertainty types.} Current approaches conflate different uncertainty sources. Architectures that explicitly represent ``I predict this action will fail'' (epistemic uncertainty), ``I don't know how to proceed'' (capability uncertainty), and ``this task cannot be done'' (task constraint knowledge) could enable more appropriate responses: seeking information when uncertain about the world, trying alternatives when uncertain about approach, and communicating impossibility when task constraints preclude success.}

\subsection{Integrating Neural and Symbolic Methods}
\label{sec:future_neurosym}

\EmphBox{\textbf{Combining learned flexibility with formal guarantees} instead of \textbf{choosing one paradigm}.\\
\noindent\textit{Addresses: Multiple challenges, particularly Implicit Knowledge (\S\ref{sssec:challenge_implicit}) and Evaluation Gap (\S\ref{sssec:challenge_eval})}}

\medskip

\clrev{
\noindent Neural approaches handle linguistic variation and data-driven generalization, while symbolic approaches enable interpretable reasoning and verification against formal constraints. Procedural understanding requires both: the flexibility to interpret diverse natural language instructions, and the rigor to ensure generated procedures are valid and executable. Studies showing that neural approaches capture action effects more reliably than preconditions~\cite{zhang-etal-2024-proc2pddl} suggest they learn some procedural structure but miss formal constraints that symbolic methods could enforce.}

\clrev{\textbf{(1) Neural proposal with symbolic verification.} Neural models could generate candidate procedures or state updates that symbolic reasoners then verify for consistency. Invalid proposals trigger regeneration or refinement. This division of labor plays to each paradigm's strengths: neural networks handle the ambiguity of natural language input; symbolic systems enforce logical constraints on outputs.}

\clrev{\textbf{(2) Grounding in formal action schemas.} PDDL and similar formalisms explicitly represent action preconditions, effects, and parameters. Neural approaches grounded in such schemas could learn to generate actions that respect formal constraints, with the schema providing automatic verification. Recent work on mapping open-domain procedural text to PDDL representations~\cite{zhang-etal-2024-proc2pddl} provides a concrete bridge between natural language and formal planning models. The key advantage is that schema-grounded approaches can be trained with automatic supervision: procedures that parse to valid PDDL and solve planning problems provide reward signal without human annotation.}

\clrev{\textbf{(3) Differentiable symbolic modules.} Rather than treating neural and symbolic components as separate systems in a pipeline, differentiable programming enables embedding \textit{symbolic structure} within neural architectures---often via differentiable relaxations, solver-in-the-loop training, or neuralized operators---so parts of the system can be trained end-to-end. For procedural understanding, this could mean modules that explicitly represent entities, apply structured state-update operations, and reason about preconditions, while learning parameters through gradient-based optimization.
}

\clfinrev{\textbf{(4) Tool Use and Function Calling as applied grounding}. 
The recent emergence of ``tool use'' and ``function calling'' capabilities 
in \gls{LLM}s~\cite{schick2023toolformerlanguagemodelsteach,qin2024toolllm, Gorilla2024} represents a 
highly practical instantiation of neuro-symbolic integration. In this 
paradigm, the \gls{LLM} serves as the neural reasoning engine that 
translates unstructured procedural intent into a formally-specified symbolic 
representation, typically a structured JSON schema. External tools and APIs 
then act as symbolic execution engines, directly instantiating the ``neural 
proposal with symbolic verification'' direction~\cite{yao2023react}. 
Concretely, the model proposes a structured action, the API executes it or 
raises a schema validation error, and the neural model uses this 
deterministic feedback to iteratively correct its plan. Formalizing API schemas as modern equivalents to planning languages 
like \gls{PDDL} offers a highly scalable pathway for grounding 
procedural language in executable actions~\cite{qin2024toolllm}, 
directly addressing the cross-domain fragmentation identified in~\ref{ssec:symbolic}. Open challenges remain, however, including 
hallucinated function calls, under-specified schemas, and the absence of 
precondition/effect validation mechanisms present in classical planning 
formalisms.}

\subsection{Improving Data Quality and Diversity}
\label{sec:future_data}

\EmphBox{\textbf{Curating procedurally-rich supervision} instead of \textbf{scaling text quantity alone}.\\
\noindent\textit{Addresses: Multiple challenges; enables progress across all tasks}}

\medskip

\clrev{
\noindent Current research concentrates heavily on WikiHow and recipe corpora, while other procedural domains remain underexplored. The finding that code-pretrained models outperform text-pretrained ones on state tracking~\cite{KimS23} indicates that data composition---not just scale---shapes procedural capabilities. Several concrete directions could diversify and enrich procedural training data:}

\clrev{\textbf{(1) Domain expansion to technical and multilingual sources.} Troubleshooting documentation from technology companies, repair manuals (iFixit), and industrial process descriptions offer procedural text with characteristics absent from cooking: more implicit domain knowledge, specialized terminology, longer dependency chains, and different failure modes. These resources remain largely untapped, partly due to inconsistent formatting that complicates extraction. Building standardized pipelines---perhaps using LLMs for initial parsing followed by human verification---could unlock this diversity.}

\clrev{Similarly, non-English procedural resources are underexplored: recipe sites in Turkish, Chinese, or Hindi; WikiHow's multilingual editions; local how-to forums. These offer both cross-lingual training data and evaluation benchmarks for testing whether procedural capabilities transfer across languages.}

\clrev{\textbf{(2) Execution trace alignment for implicit state annotation.} Current corpora annotate surface text but rarely capture the implicit states that procedures assume. Execution traces---from cooking videos, robotic demonstrations, or game playthroughs---make these states observable. For example, a cooking video reveals that ``saut\'{e} until soft'' involves visible texture and color changes that text leaves implicit; a robotic demonstration shows the gripper configurations and force patterns that ``pick up the cup'' requires.}

\clrev{Aligning such traces with procedural text is labor-intensive but feasible. Existing resources provide starting points: EPIC-KITCHENS~\cite{Damen2021PAMI} pairs cooking videos with action annotations; robotic demonstration datasets~\cite{nair2021learning} include state observations; game replay databases record action-state sequences. The research challenge is developing alignment methods that map continuous sensory traces to discrete textual descriptions---a form of grounded language learning that could provide supervision unavailable from text alone.}

\clrev{\textbf{(3) Hierarchical annotation for long-horizon procedures.} Current datasets typically contain 5--12 steps, yet real-world procedures often span dozens or hundreds of steps with nested subprocedures. WikiHow's method-step-substep structure provides some hierarchy, but most datasets flatten this into linear sequences, losing the goal structure that enables human comprehension of complex procedures.}

\clrev{Creating datasets that preserve hierarchical structure---explicitly marking which steps serve which subgoals, which subgoals combine into which higher goals---would enable training and evaluation of hierarchical approaches (\S\ref{sec:future_hierarchy}). The business process modeling literature, with BPMN representations of complex workflows~\cite{ProcessModelExtraction}, offers annotation schemas for such hierarchy. Adapting these schemas to natural language procedures, perhaps through crowdsourced annotation of existing corpora, could provide the structured supervision that flat sequence data cannot.}

\clrev{\textbf{(4) Validated synthetic data through execution filtering.} Large language models can generate procedural text at scale, but such text often contains subtle errors: invalid action sequences, impossible state transitions, physically implausible steps. CoScript~\cite{yuan-etal-2023-distilling} demonstrates an over-generate-then-filter paradigm using constraint-faithfulness filtering; extending this to execution-based filtering in simulators is a promising next step.}

\clrev{\textbf{(5) Paraphrasing for linguistic diversity.} Synthetic datasets offer control over procedural complexity but typically lack linguistic diversity, using templated language that fails to capture natural variation. A promising direction, successfully applied in other NLP domains, is two-stage generation: first create synthetic procedures for their structural properties (correct action sequences, valid state transitions), then crowdsource paraphrases to add linguistic naturalism~\cite{KimS23}. This combination yields datasets that are both structurally controlled---enabling systematic evaluation of procedural capabilities---and linguistically natural---enabling transfer to real-world text.
}

\clrev{\section{Conclusion}}
\label{sec:conclusion}
\clrev{This survey provides a systematic analysis of procedural language understanding across representation schemes, resources, and downstream tasks. Examining 181 papers, we identify fundamental patterns, persistent challenges, and opportunities for future work.}

\clrev{Our analysis reveals a central tension between current representation paradigms: event-centric approaches often capture inter-step dependencies but under-specify entity state evolution, while entity-centric representations track state changes but often treat actions as opaque triggers. Unified representations that jointly model both aspects remain rare. We observe clear patterns in how different representations serve different tasks, yet also find notable gaps where representation--task combinations remain underexplored.}

\clrev{Benchmark performance often disconnects from real-world applicability. Models achieving high scores on automatic metrics can still fail when their outputs are executed, revealing fundamental evaluation limitations. Six core challenges persist across tasks: implicit knowledge (crucial information is unstated), compounding failures in state maintenance, evaluation frameworks that reward surface similarity over functional correctness, compositional reasoning failures, the gap between language understanding and grounded action, and the inability to detect and recover from errors.}

\clrev{The field has evolved from isolated event extraction and entity tracking toward integrated procedural understanding. As large language models demonstrate instruction-following capabilities in short-horizon settings, research increasingly addresses complex, multi-step procedures requiring simultaneous event understanding, entity tracking, and environmental grounding. Continued progress necessitates advances in evaluation methodology, data curation, and architectural innovations---including neuro-symbolic integration, hierarchical planning, and error recovery mechanisms.}

\appendix
\vspace{10pt}

\appendixsection{Database Specific Keyword Settings}
\label{app:db-query}

The query includes thirteen sub-queries focusing on frequent terms in procedural text research. Initially, we used broad keywords like ``procedural text'' and ``instructional text,'' then refined the search to cover common corpora, instruction forms, and tasks. The full list is given in Table~\ref{keyword_table}.

   \begin{table}[h]
    \centering
    \scalebox{0.65}{
      \begin{tabular}{|*{2}{c|}}
      \hline
      Google Scholar, IEEE Xplore & DBLP \\
      \hline
       Exact Phrase Match (PM) &  Exact Match per Word (EMW)\\
      \hline
      \multicolumn{2}{|c|}{``procedural text''} \\
      \multicolumn{2}{|c|}{``instructional text''} \\
      \multicolumn{2}{|c|}{``natural language instruction(s)''} \\
      \multicolumn{2}{|c|}{``entity tracking''} \\
      \multicolumn{2}{|c|}{``entity state tracking''}  \\
      \multicolumn{2}{|c|}{``wikihow''} \\
      \multicolumn{2}{|c|}{``script knowledge''} \\
      \multicolumn{2}{|c|}{``BPMN generation''} \\
      \multicolumn{2}{|c|}{``action sequence natural language''}\\
      \multicolumn{2}{|c|}{``action sequence annotation''} \\
      \multicolumn{2}{|c|}{``instruction parsing''} \\
      \multicolumn{2}{|c|}{``instruction manual'' AND ``parsing''} \\
      \multicolumn{2}{|c|}{``recipe parsing''} \\
      \hline
      \end{tabular}
      }
  \caption{Keywords}
  \label{keyword_table}
  \end{table}

\appendixsection{Grounded Tasks Oriented Dialogues Datasets}
\label{app:tods}
 Table~\ref{tab:tod-datasets-overview} showcases a variety of datasets used in grounded task-oriented dialogue research, illustrating diverse tasks ranging from single subtasks, such as Knowledge Identification (which can be mentioned with different names, including User Utterance Grounding, Grounding Span Prediction, Flow-chart Retrieval) to full pipeline tasks like Cascade Dialogue Success, and employing different metrics.
    \begin{table}[htbp]
            \centering
                \scalebox{0.6}{
                \begin{tabular}{@{}lccccccc@{}}
                    \toprule
                    \textbf{Dataset} & \textbf{Docs} & \textbf{Dialogues} & \multicolumn{4}{c}{\textbf{Evaluation}} \\ 
                    \cmidrule(lr){4-7}
                    & & & \textbf{Task} & \textbf{Metric} & \textbf{Baseline} & \textbf{SoTA} \\ 
                    \midrule
                    \multirow{2}{*}{Doc2Dial} & \multirow{2}{*}{480} & \multirow{2}{*}{4470} & Knowledge Identification& EM & 55.4 &68.4~\cite{daheim-etal-2021-cascaded} \\ 
                    & & & & F1 & 66.6 & 88.7~\cite{daheim-etal-2022-controllable} \\ 
                    \midrule
                    \multirow{2}{*}{MultiDoc2Dial} & \multirow{2}{*}{488} & \multirow{2}{*}{4796} & Knowledge Identification& EM & 26.6 &51.0~\cite{zhao-etal-2023-causal} \\ 
                    & & & & F1 & 43.7 & 64.5~\cite{zhao-etal-2023-causal} \\ 
                    \midrule
                    \multirow{1}{*}{Task2Dial} & \multirow{1}{*}{353} & \multirow{1}{*}{478} & - & - & - & - \\
                    \midrule
                    \multirow{2}{*}{CookDial} & \multirow{2}{*}{260} & \multirow{2}{*}{260} & User Question Understanding & Accuracy & 94.5 & - \\ 
                    & & & & F1 & 91.0 & -\\ 
                    \midrule
                    \multirow{1}{*}{ABCD} & \multirow{1}{*}{55} & \multirow{1}{*}{10042} & Cascade Dialgoue Success & Cascading Eval &  31.9  & 60.7~\cite{ramakrishnan2023multi} \\
                    \midrule
                    \multirow{2}{*}{FLODial} & \multirow{2}{*}{12} & \multirow{2}{*}{5,476} & Knowledge Identification & Success Rate & 33.7 & 67.1~\cite{hu2024uncertainty} \\ 
                    & & & & R@1 & 91.0 & -\\ 
                    \midrule
                    \multirow{5}{*}{TEACh} & \multirow{5}{*}{12} & \multirow{5}{*}{3,047} & Execution from Dialogue History & SR & 7.06 & 18.60~\cite{jain2024odin} \\ 
                    & & & & GC & 9.57 & 18.60~\cite{jain2024odin}\\
                    & & & Trajectory from Dialogue & SR & 0.51 & - \\ 
                    & & & & GC & 20.3 & -\\
                    & & & Two-Agent Task Completion & SR & 24.4 & - \\
                    \bottomrule
                \end{tabular}
                }
                \caption{Commonly Used Grounded Task-Oriented Dialogue Datasets}
                \label{tab:tod-datasets-overview}
        \end{table}
\clearpage
\appendixsection{List of Grounded and Ungrounded Task Papers}
\label{app:tasks-papers}

List of all papers for grounded and ungrounded tasks are given in Table~\ref{tab:tasks_papers}.
\begin{table}[http]
    \centering
    \scalebox{0.55}{
    \begin{tabular}{llp{10cm}}
        \toprule
        \textbf{Category} & \textbf{Task/Environment} & \textbf{Papers} \\
        \midrule
        \multirow{8}{*}{\rotatebox{90}{Ungrounded}} & Summarization & ~\cite{wikihowMultilingualSummary, wikihowSummary, le-luu-2024-extractive, Boni2021HowSummAM, DeChant2022SummarizingAV, ALFRED20, scialom-etal-2019-answers, le-luu-2024-extractive} \\
        & Event Alignment & ~\cite{lin-etal-2020-recipe, nandy-etal-2023-clmsm, regneri-etal-2010-learning, wanzare-etal-2017-inducing, DonatelliSBKZK21, marin2019learning, bien-etal-2020-recipenlg, DeScript, Mavroudi2023LearningTG, 8237345, ALFRED20, scialom-etal-2019-answers, ProPara1, ZhXuCoAAAI18, kwon2024ordersum} \\
        & Detecting and Correcting Implicit Instructions & ~\cite{AnthonioBR20, roth-anthonio-2021-unimplicit, Zeng2020Missing, RecipeNER, bisk-etal-2019-benchmarking, FengFLW22, clark2018leveraging, cheng-erk-2018-implicit, Zeng2020Missing, zhang-etal-2023-clevr} \\
        & Entity State Tracking & ~\cite{OpenPI, ProPara2, TechTrack, zhang-etal-2024-openpi2, zhang-etal-2024-openpi2, rim-etal-2023-coreference, ProPara1, clark2018leveraging, ZhangGQWJ21, SconeLongPL16, li-etal-2021-implicit, KimS23, BosselutLHEFC18, CREPE, cheng-erk-2018-implicit, kazeminejad-palmer-2023-event, nandy-etal-2023-clmsm, nandy-etal-2024-order, ProPara1, Chen2019AlchemyAQ} \\
        & Instruction Parsing & ~\cite{harashima-hiramatsu-2020-cookpad, pan2020multi, bhatt-etal-2024-end, MIAIS, 7964092, nabizadeh-etal-2020-myfixit, ManualToPDDL, zhang-etal-2021-learning, FengZK18, maeta-etal-2015-framework, ParkM18, MysoreJKHCSFMO19, ProcessModelExtraction, diwan2020named, ito2020natural, ri-etal-2022-finding, zhou-etal-2022-show, textMining23, wang2022learning, kourani2024process, kumaran2024procedural, regneri-etal-2010-learning, DeScript, wanzare-etal-2017-inducing, LiuGPSL18, zhang-etal-2020-analogous, WuZHSP23, fan-hunter-2023-understanding, bellan2023pet, lal2024cat, zhang-etal-2024-openpi2, zhang-etal-2024-proc2pddl, MIECH19howto100m, kiddon-etal-2015-mise, regneri-etal-2010-learning, DeScript, wanzare-etal-2017-inducing, LiuGPSL18, ShiKFHL17, 10.1093/database/baaa077, 8099810, Brach2025Effectiveness} \\
        & Process Generation & ~\cite{lyu-etal-2021-goal, zhang-etal-2020-analogous, sun-etal-2023-incorporating, kiddon-etal-2016-globally, rojowiec-etal-2020-generating, liu-etal-2022-counterfactual, lu2023multimodal, 9288722, InScript, proScript, NguyenNCTP17, yuan-etal-2023-distilling, DeScript, proScript, ScEdit2025} \\
        & Question Answering & ~\cite{Bolotova-Baranova23, zhou-etal-2019-learning-household, uzunoglu-etal-2024-paradise, wikiHowStepInference, yang-etal-2021-visual, Mcscript, Mcscript2, zellers-etal-2019-hellaswag, storks-etal-2021-tiered-reasoning, RecipeQA, bauer-bansal-2021-identify, bAbI, tandon-etal-2019-wiqa, wang-etal-2023-steps, zhang2023entity} \\
        & Knowledge Acquisition/Mining & ~\cite{chen-etal-2020-trying, jung2010Automatic, ChuTW17, steinert2020planning, sen2023task2kb, ManualToPDDL, Babli2019ACK} \\
        \midrule
        \multirow{6}{*}{\rotatebox{90}{Grounded}} & Web & ~\cite{gur2024realworld, FLINMazumderR21, Russ2021, Kim2023LanguageMC, Mind2Web2023, LiuGPSL18, WebArena23, Nakano2021WebGPTBQ, WebShopN22, ShiKFHL17, deng-etal-2024-multi, zhu2024knowagent} \\
        & Simulations/Navigation & ~\cite{artist, SayNav, embodiedQA, misra-etal-2015-environment, FAVOR, zhou-etal-2022-hierarchical-control, MapTaskVogelJ10, MisraBBNSA18, REVE-CE, REVERIE, Misra2014TellMD, zang-etal-2018-translating, FengFLW22, VirtualHome, Anderson2017VisionandLanguageNI, MacMahonSK06, artzi-zettlemoyer-2013-weakly, shi-etal-2024-opex} \\
        & Robotic & ~\cite{Howard2014ANL, Tellex2011ApproachingTS, Mees2021CALVINAB, TellexKDWBTR11, nair2021learning, Singh2022LearningNP, bisk-etal-2016-natural, Scalise2018NaturalLI, Bucker2022ReshapingRT, SayCan, Matuszek2012LearningTP, InstructRobot2025} \\
        & Game  & ~\cite{Toshniwal2021ChessAA, Narayan-ChenJH19, DrawMeAFlower22, Li2022EmergentWR, Hu2019HierarchicalDM, jayannavar-etal-2020-learning, ScriptWorld, wu2024smartplay, TextWorld, Hooshyar2018ADP} \\
        & GUI/Mobile & ~\cite{MotifBurnsAAKSP22, AndroidEnv21, Banerjee2023LexiSL, LiHZZB20, DanyangZhang2023_MobileEnv} \\
        & Dialogue  & ~\cite{chen-etal-2021-action, shao-nakashole-2020-chartdialogs, Narayan-ChenJH19, Jiang_2022_cookdial, srinet-etal-2020-craftassist, gao2022dialfred, feng-etal-2020-doc2dial, FlowDial, moghe-etal-2024-interpreting, MetaGUISunCCDZY22, AmmanabroluR21a, feng2021multidoc2dial, strathearn-gkatzia-2022-task2dial, TEACh} \\
        \bottomrule
    \end{tabular}
    }
    \caption{Ungrounded and Grounded Tasks Papers}
    \label{tab:tasks_papers}
\end{table}
\clearpage
\appendixsection{Grounded Environment Details}
\label{ssec:grounded_env_papers}

 We have identified the following divergences across environments: \textit{data instance} (typical components of the dataset), \textit{dataset size}, \textit{the action space}, \textit{tasks performed} in the \textit{environment}. The end tasks vary depending on the environment. For instance, simulating user actions on websites, such as filling forms or navigating menus~\cite{WebArena23, Mind2Web2023, WebShopN22} are common in Web environments, while gaming scenarios often involve character navigation and object interaction within virtual worlds~\cite{wu2024smartplay, Narayan-ChenJH19, wu2024smartplay, DrawMeAFlower22}. A summary is given in Table~\ref{tab:grounded_repr}.
\begin{table}[h]
\centering
    \scalebox{0.60}{
    \begin{tabular}{p{3cm}p{3cm}p{3cm}p{3cm}p{3cm}p{2cm}} 
       
        \toprule
         & \textbf{Environment} & \textbf{Data instance} & \textbf{Task} & \textbf{Num of Actions} & \textbf{Size} \\ 
        
        \midrule       
        PixelHelp~\cite{LiHZZB20} & Google Phone & Screen, instruction, grounded UI actions & Account configuration, Gmail, Chrome, Photos tasks & 2-8 steps & 187 \\

        WinHelp~\cite{branavan-etal-2009-reinforcement} & Windows 2000 VM & Screen, instruction, grounded UI actions & Windows troubleshooting & 10,3 & 128 \\  
                
        AndroidHowTo~\cite{LiHZZB20} & Google Phone & Screen, instruction, grounded UI actions & Account configuration, Gmail, Chrome, Photos tasks & 2-8 steps & 187 \\
        
        AndroidEnv~\cite{AndroidEnv21} & Mobile & Emulated Android devices, touchscreen gestures, real-time interaction&  UI navigation, game playing, and basic utilities & Varies & 100 (tasks) \\
        
        \midrule
        WebArena~\cite{WebArena23} & Web & Screenshot/HTML Dom Tree/Accessibility Tree, Multitab, (high-level) instruction, grounded UI actions & web-based tasks for e-commerce, social forums, collaborative software and content management & Varies  &812 \\ 

        Mind2Web~\cite{Mind2Web2023} & Web & Real-world websites, instructions, and action sequences & variety of tasks across different domains on any website & Average = 7.3 & 2,350 (from 137 websites across 31 domains)\\

        World Of Bits~\cite{ShiKFHL17} & Web &  Screen pixels, DOM, keyboard and mouse actions & web-based tasks via natural language queries & Varies & 100 (web tasks) \\

        WOB++ ~\cite{LiuGPSL18} & Web & DOM elements, workflow steps (Click, Type)&  Email processing, form filling, etc., in noisy environments & Varies & 80 tasks \\
        
        \midrule
        
        CrossBlock~\cite{LiuGPSL18} & Grid  & Grid state, instruction, action & Clear the grid by drawing lines & 5,86 & 50 \\ 
        
        Hexagons (Draw me a flower)~\cite{DrawMeAFlower22} & Grid & Grid state, instruction, action & Fill the grids with colors & 6,73 & 620 \\   

        Chess ~\cite{Toshniwal2021ChessAA} & Grid & Chessboard state, instructions & Predicting legal chess moves & Varies & 2.5M scenarios \\

        \midrule

        SCONE~\cite{SconeLongPL16} & Simulation & Instructions, Logical form & Infer the final environment state following the actions & 5 & 4K \\    
        
        VirtualHome~\cite{VirtualHome} & 3D Simulation & Instructions, Scratch Program, Video & Household & 11,6 & 2821 \\
        
        LANI, CHAI~\cite{MisraBBNSA18} & 3D World w/Landmarks & Instructions, world states, actions & Navigation/ Househould & 4,7/7,7 & 6000/1596  \\

        \midrule 

        CALVIN~\cite{Mees2021CALVINAB} & Robotics/Simulation & language instructions, multimodal sensors & Long-horizon tasks (e.g., open drawer, push block) & $\sim$5-10 & 20K language directives \\

        \citet{Bucker2022ReshapingRT} & Robotics/Simulation & Predefined trajectories, object positions, natural language commands & Trajectory reshaping based on natural language commands & $\sim$ 5-10 steps & $\sim$ 10,000 trajectory labels \\

        LOReL~\cite{nair2021learning} & Robotics/Simulation & Vision-based interaction, natural language commands, multimodal sensors & Language-conditioned manipulation (e.g., open drawer, move stapler) & up to 150 & 53,000 scenarios\\

        \midrule 

         Meta-GUI~\cite{MetaGUISunCCDZY22} & Dialogue/Web & Screenshots, HTML Dom Tree/Accessibility Tree, touch and text inputs & Multi-modal conversational interactions (e.g., booking a hotel, checking the weather) & $\sim$ 4-5 & 1125 dialogues \\
        \bottomrule
    
    \end{tabular}
    }
\caption{Grounded instructions. Screen: The structured UI state/tree not pixels. }
\label{tab:grounded_repr}
\end{table}

\begin{figure}[h!]
            \centering
            \resizebox{0.6\textwidth}{!}{ 
                \fbox{ 
                    \begin{minipage}{0.9\textwidth} 
                        \begin{subfigure}{0.45\textwidth}
                            \centering
                            \begin{minipage}{\textwidth}
                                \textbf{Ingredients:} \\
                                1- 1 ½ cups all-purpose flour \\
                                2- 1 ½ teaspoons baking powder \\
                                3- ½ teaspoon salt \\
                                4- ½ cup unsalted butter, softened \\
                                5- 1 cup granulated sugar \\
                                6- 2 large eggs \\
                                7- 1 teaspoon vanilla extract \\
                                8- ½ cup milk \\
                                \textbf{Instructions:} \\
                                1. Prepare dry ingredients: Mix flour, baking powder, and salt. \\
                                2. Prepare wet ingredients: Cream butter and sugar, add eggs, and vanilla. \\
                                3. Combine ingredients: Gradually add dry ingredients to wet ingredients while mixing. \\
                                4. Add Milk. \\
                                5. Pour batter into a greased 8-inch cake pan. \\
                                6. Bake for 25-30 minutes at 350°F (175°C). \\
                                7. Let cool for 10 minutes \\
                                8. Transfer to a wire rack.
                            \end{minipage}
                            \caption{Recipe Text}
                            \label{fig:recipe}
                        \end{subfigure}
                        \hfill
                        \vrule width 1pt 
                        \hfill
                        \begin{subfigure}{0.5\textwidth}
                            \centering
                            \parbox{\textwidth}{%
                                \noindent
                                \textbf{Prepare} (1- 1\(\frac{1}{2}\) cups all-purpose flour, 1\(\frac{1}{2}\) teaspoons baking powder, \(\frac{1}{2}\) teaspoon salt) 
                                \\$\rightarrow$\\
                                \textbf{Prepare} (\(\frac{1}{2}\) cup unsalted butter, softened, 1 cup granulated sugar, 2 large eggs, 1 teaspoon vanilla extract) 
                                \\$\rightarrow$\\
                                \textbf{Combine} (ingredients) 
                                \\$\rightarrow$ \\
                                \textbf{Add} (\(\frac{1}{2}\) cup milk) 
                                \\$\rightarrow$\\
                                \textbf{Pour} (batter, into greased 8-inch cake pan) 
                                \\$\rightarrow$ \\
                                \textbf{Bake} (25-30 minutes at 350°F) 
                                \\$\rightarrow$ \\
                                \textbf{Cool} (10 minutes) 
                                \\$\rightarrow$\\
                                \textbf{Transfer} (to wire rack)
                            }
                            \caption{Recipe Representation in Wiki HG~\cite{FengZK18}}
                            \label{fig:vanilla_cake_process}
                        \end{subfigure}
                        \vspace{0.5cm}
                        \hrule 
                        \vspace{0.5cm}
                        \begin{subfigure}{0.45\textwidth}
                            \centering
                            \includegraphics[width=0.40\textwidth,height=7cm]{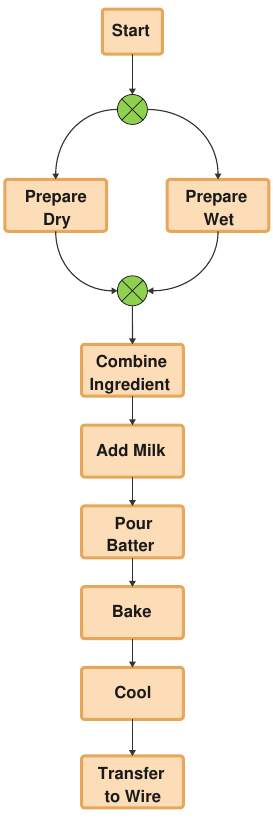}
                            \caption{Recipe Representation as Process Graph~\cite{ProcessModelExtraction}}
                            \label{fig:pm}
                        \end{subfigure}
                        \hfill
                        \vrule width 1pt 
                        \hfill
                        \begin{subfigure}{0.5\textwidth}
                            \centering
                            \includegraphics[width=0.75\textwidth,height=7cm]{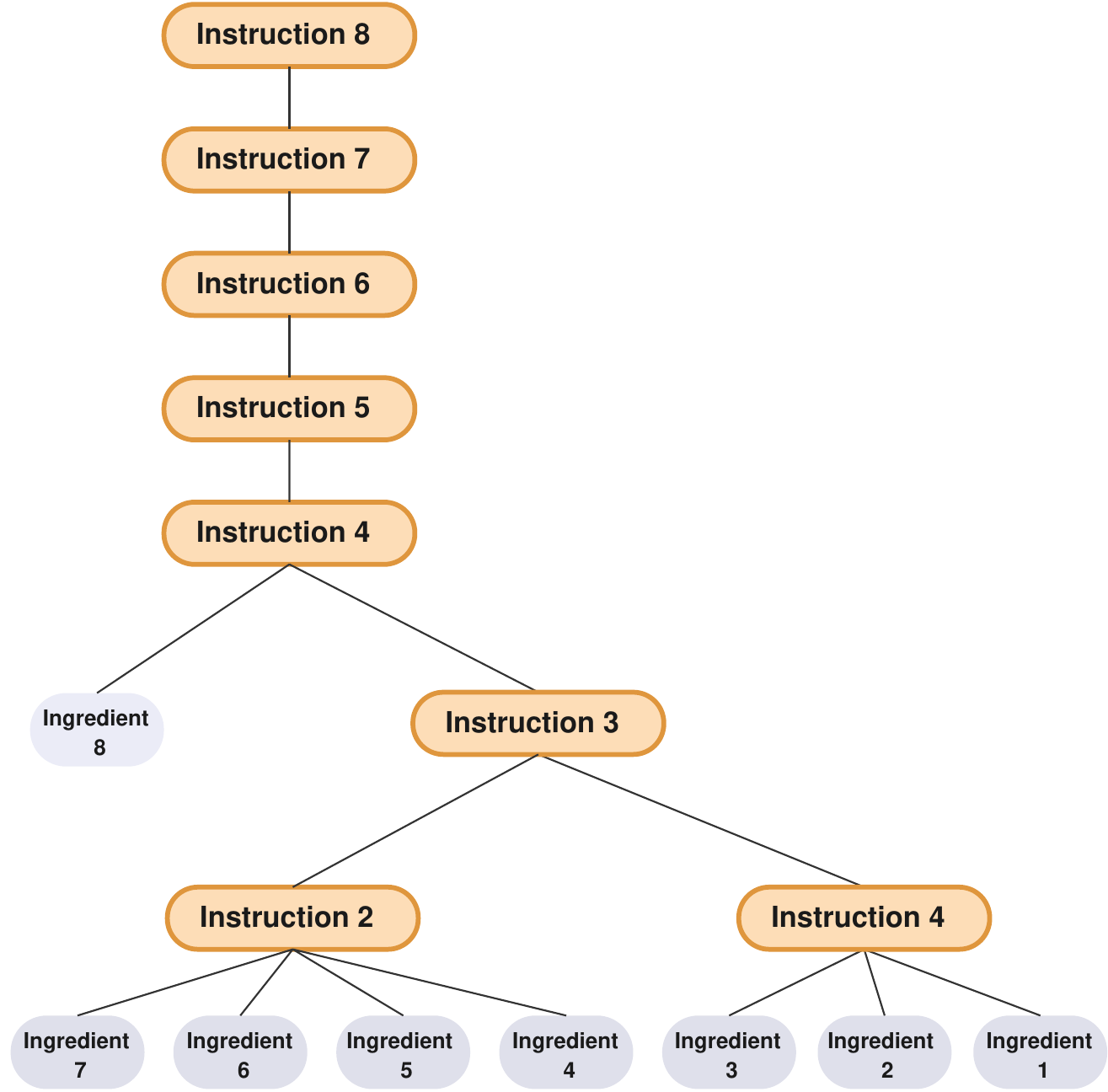}
                            \caption{Recipe Representation in SIMMR~\cite{jermsurawong-habash-2015-predicting}}
                            \label{fig:simmr}
                        \end{subfigure}
                    \end{minipage}
                }
            }
            \caption{Vanilla cake recipe representations: (a) Recipe Text, (b) Text Representation from Wiki HG~\cite{FengZK18}, (c) Process Graph~\cite{ProcessModelExtraction}, and (d) SIMMR structure~\cite{jermsurawong-habash-2015-predicting}.}
            \label{fig:prcoess-strucutr-representaions}
        \end{figure}

\begin{figure}[h]
        \centering
        \includegraphics[width=0.5\textwidth]{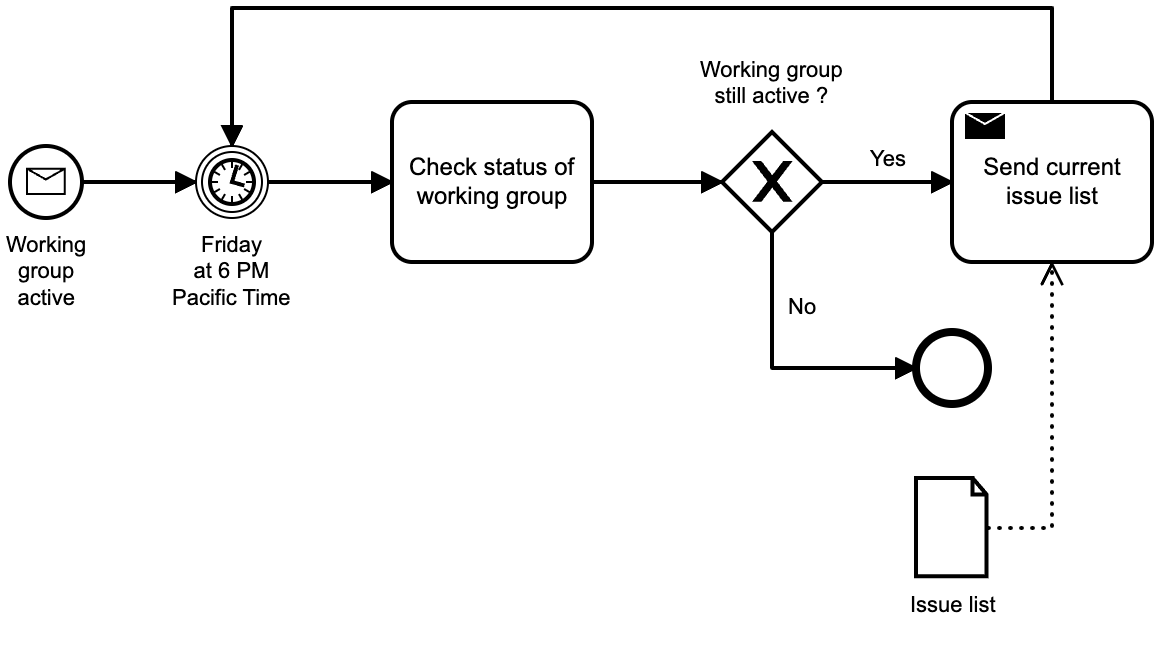}
        \caption{Example of a Business Process Model and Notation for ``Sending an issue list'' process~\cite{wikipedia_bpmn_2024}}
        \label{fig:BPMN-fig}
    \end{figure}

\clearpage
\appendixsection{Example of Event Centric Representations}
\label{app:event-centric-represent}
    Fig.~\ref{fig:prcoess-strucutr-representaions} illustrates the diversity of structures used to represent the process of baking a cake through three different formalisms. While all three approaches capture the same goal, the way in which they structure the actions and entities involved is distinct. Fig.~\ref{fig:BPMN-fig} shows an example of BPMN for a ``Sending an issue list'' process.

\begin{acknowledgments}
This work has been supported by the Scientific and Technological Research Council of Türkiye~(TÜBİTAK) as part of the project ``Automatic Learning of Procedural Language from Natural Language Instructions for Intelligent Assistance'' with the number 121C132. We also gratefully acknowledge KUIS AI Lab for providing computational support. Special thanks to Müge Kural for her valuable assistance during the paper identification stage.
\end{acknowledgments}

\bibliographystyle{compling}
\bibliography{COLI}

\end{document}